%% file: eccv2020submission.tex
\pdfoutput=1

\documentclass[runningheads]{llncs}
\usepackage{graphicx}
\usepackage{comment}
\usepackage{amsmath,amssymb} 
\usepackage{color}
\usepackage{UserCommands}
\usepackage{caption}
\usepackage{subcaption}
\usepackage{array}
\usepackage{wrapfig}
\usepackage{multirow}
\usepackage{tabularx}
\usepackage{amssymb}
\usepackage{soul}
\usepackage{cite}

\newcolumntype{C}[1]{>{\centering\arraybackslash}p{#1}}

\begin{document}
\pagestyle{headings}
\mainmatter
\def\ECCVSubNumber{3411}  

\title{{JSENet: Joint Semantic Segmentation and Edge Detection Network for 3D Point Clouds}
} 

\titlerunning{JSENet}
%
\author{Zeyu HU\inst{1}\orcidID{0000-0003-3585-7381} \and
Mingmin Zhen\inst{1}\orcidID{0000-0002-8180-1023} \and
Xuyang BAI\inst{1}\orcidID{0000-0002-7414-0319} \and
Hongbo Fu\inst{2}\orcidID{0000-0002-0284-726X} \and 
Chiew-lan Tai\inst{1}\orcidID{0000-0002-1486-1974}
}
\authorrunning{Z. HU et al.}
%
\institute{Hong Kong University of Science and Technology \email{\{zhuam,mzhen,xbaiad,taicl\}@cse.ust.hk} \and
City University of Hong Kong \\
\email{hongbofu@cityu.edu.hk}}
\maketitle

\begin{abstract}
Semantic segmentation and semantic edge detection can be seen as two dual problems with close relationships in computer vision. Despite the fast evolution of learning-based 3D semantic segmentation methods, little attention has been drawn to the learning of 3D semantic edge detectors, even less to a joint learning method for the two tasks. In this paper, we tackle the 3D semantic edge detection task for the first time and present a new two-stream fully-convolutional network that jointly performs the two tasks. In particular, we design a joint refinement module that explicitly wires region information and edge information to improve the performances of both tasks. Further, we propose a novel loss function that encourages the network to produce semantic segmentation results with better boundaries. Extensive evaluations on S3DIS and ScanNet datasets show that our method achieves on par or better performance than the state-of-the-art methods for semantic segmentation and outperforms the baseline methods for semantic edge detection. Code release: \url{https://github.com/hzykent/JSENet}

\end{abstract}

\section{Introduction}

\input{introduction.tex}
\section{Related Work}

\input{related_work.tex}

\input{method.tex}

\input{method_1.tex}

\input{method_2.tex}

\section{Experiments}

\input{experiment.tex}

\section{Conclusions}

\input{conclusion.tex}

%
%
\bibliographystyle{splncs}

\input{eccv2020submission.bbl}

\clearpage

\input{supplementary.tex}

\end{document}

%% file: introduction.tex
\begin{figure}[htp]

\centering
\includegraphics[width=.3\textwidth]{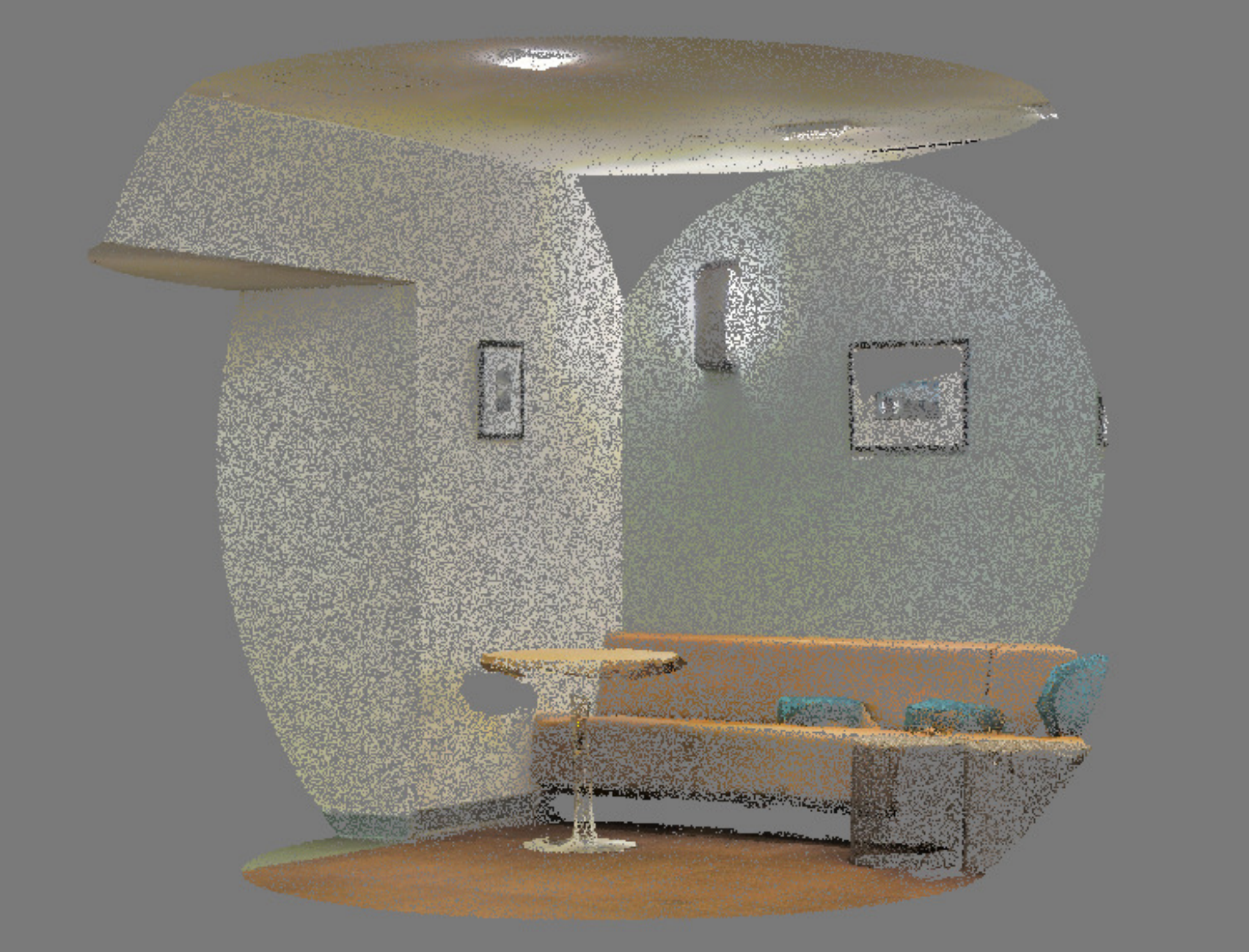}\hfill
\includegraphics[width=.3\textwidth]{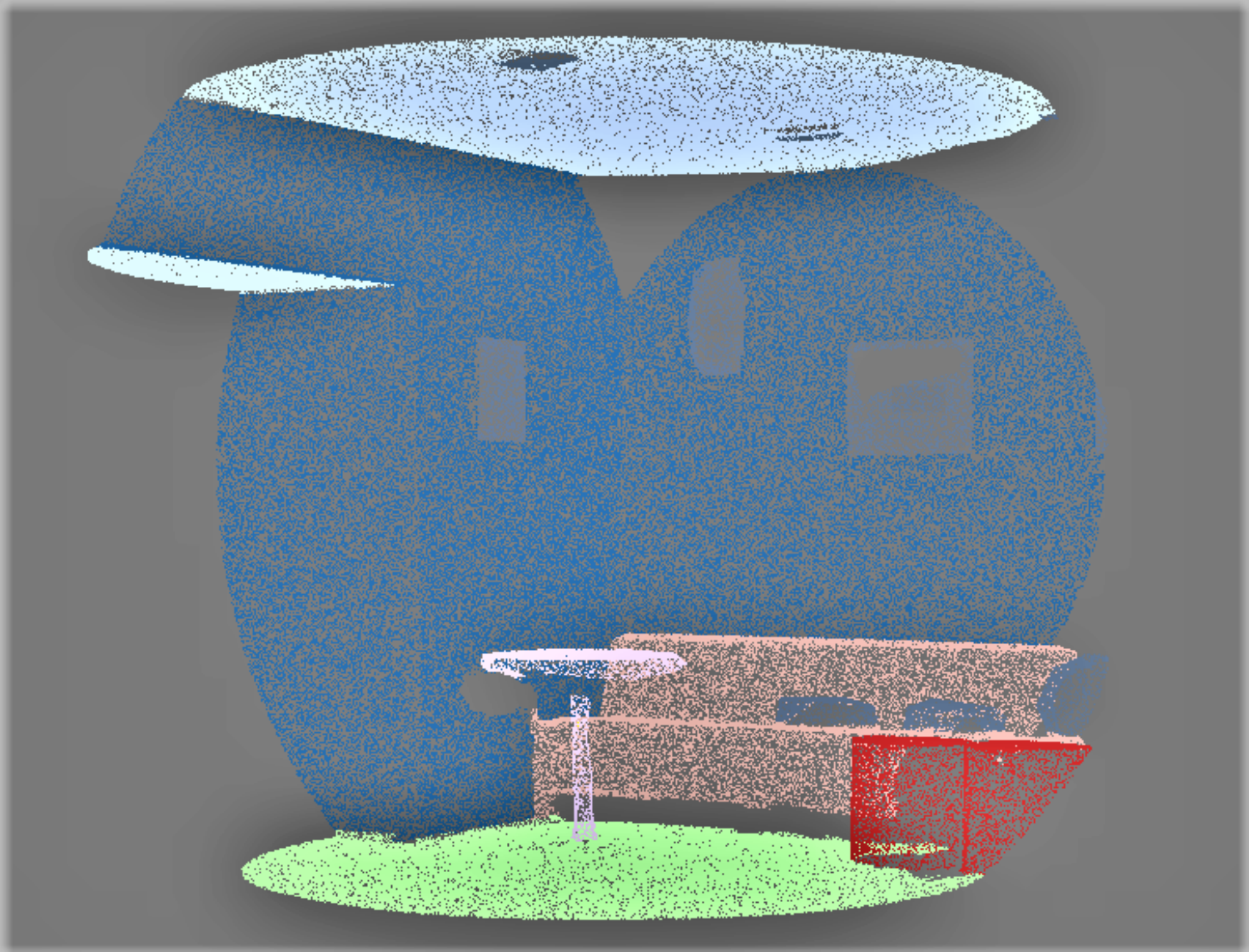}\hfill
\includegraphics[width=.3\textwidth]{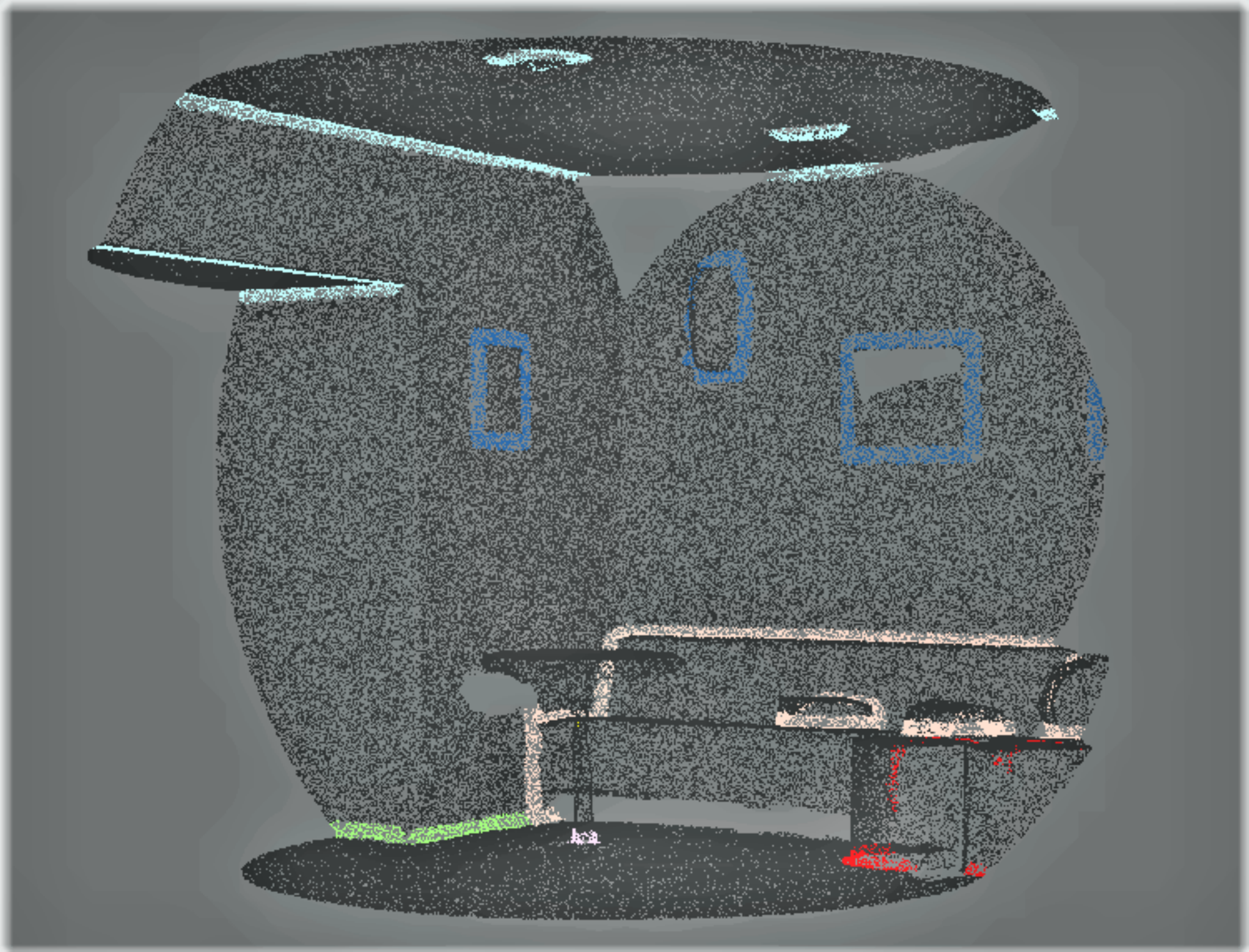}

\caption{(Left) Point cloud {of a real-world scene from S3DIS \cite{armeni_cvpr16}}; (Middle) Semantic segmentation point (\SemSegPoint) mask; (Right) Semantic edge point (\SemEdgePoint) map. {For visualization, we paint an edge point to the color of one of its class labels.}
}
\label{fig:example}

\end{figure}

\pdfoutput=1
Semantic segmentation {(\SemSeg)} and semantic edge detection {(\SemEdgeD)} are two fundamental problems for scene understanding. The former aims to parse a scene and assign a class label to each pixel in images or each point in 3D point clouds. The latter focuses on detecting edge pixels or edge points and classifying each of them to one or more classes. Interestingly, the {\SemSeg} and the {\SemEdgeD} tasks can be seen as two dual problems with even interchangeable outputs in an ideal case (see Fig. \ref{fig:example}). While {\SemSeg} has been extensively studied in both 2D and 3D \cite{long2015fully,takikawa2019gated,li2018pointcnn,jaritz2019multi,qi2017pointnet,qi2017pointnet++,dai2017scannet}, {\SemEdgeD} has only been explored in 2D  \cite{yu2017casenet,liu2018semantic,yu2018simultaneous,acuna2019devil}, to our best knowledge. 

There are strong motivations to address both problems in a joint learning framework. On one hand, previous {\SemSeg} models tend to struggle in edge areas since these areas constitute only a small part of the whole scene and thus have little effect on the supervision during training  \cite{zhao2017pyramid,lin2017refinenet}. Performing the complementary {\SemEdgeD} task simultaneously may help the network sharpen the boundaries of the predicted {\SemSeg} results \cite{wu2019stacked}. On the other hand, existing {\SemEdgeD} models are easily affected by non-semantic edges in the scene \cite{yu2017casenet,acuna2019devil}, while a trained {\SemSeg} model is less sensitive to those edges \cite{yu2018learning}. Information from the {\SemSeg} model {thus} may help {\SemEdgeD} models suppress network activations on the non-semantic edges. Despite the close relationships of the two tasks, there is no existing work tackling them jointly as far as we know.

{Although not focusing on the {\SemSeg} and the {\SemEdgeD} tasks, in 2D, several existing works have already made fruitful attempts at proposing joint learning methods for complementary segmentation and edge detection tasks 
\cite{takikawa2019gated,cheng2017fusionnet,bertasius2016semantic,peng2017large,su2019selectivity,lin2017refinenet,wu2019stacked}. These works often treat the two tasks na\"{i}vely by sharing parts of their networks and limit the interactions between them to the feature domain. Predicted segmentation masks and edge maps are used only for loss calculation and do not contribute to each other.
Strong links between the outputs of the two tasks are not fully exploited.}

In this work, we introduce the task of {\SemEdgeD} into the 3D field 
and propose \emph{JSENet}, a new 3D fully-convolutional network (FCN) for joint {\SemSeg} and {\SemEdgeD}. The proposed network consists of two streams and a joint refinement module on top of them. Specifically, we use a classical encoder-decoder FCN for {\SemSeg} in one stream, which outputs semantic segmentation point (\SemSegPoint) masks, and add an {\SemEdgeD} stream outputting semantic edge point (\SemEdgePoint) maps in parallel. The key to our architecture is the lightweight joint refinement module. It takes the output {\SemSegPoint} masks and {\SemEdgePoint} maps as inputs and jointly refine{s} them by explicitly exploiting the duality between them. Moreover, we further propose a novel loss function to encourage the predicted {\SemSegPoint} masks to {better} align with the ground truth semantic edges. 

To summarize, our contributions are threefold: 
    \begin{enumerate}
    \item We introduce the task of {\SemEdgeD} into the 3D {field}  
    and design an FCN-based architecture with enhanced feature extraction and hierarchical supervision to generate precise semantic edges.

    \item We propose a novel framework for joint learning of {\SemSeg} and {\SemEdgeD}, named JSENet, with an effective lightweight joint refinement module {that}
    brings improvements by explicitly exploiting the duality between the two tasks. 
    
    \item We propose a dual semantic edge loss{, which} 
    {encourages the network to produce SS results with finer boundaries.} 
\end{enumerate}

{To build our FCN network in 3D, we resort to KPConv \cite{thomas2019kpconv}, a recently proposed kernel-based convolution method for point clouds, for its state-of-the-art performance and ease of implementation.
We conduct extensive experiments to demonstrate the {effectiveness of our method.} Since there is no existing 3D {\SemEdgeD} dataset, we construct a new 3D {\SemEdgeD} benchmark using S3DIS \cite{armeni_cvpr16} and ScanNet \cite{dai2017scannet} datasets. We achieve state-of-the-art performance for the {\SemSeg} task with IoU scores of $67.7\%$ on S3DIS Area-5 and $69.9\%$ on ScanNet test set. For the {\SemEdgeD} task, 
{our method} outperforms the baseline methods by a large margin.}

%% file: related_work.tex
\subsection{3D Semantic Segmentation}

Based on different data representations, 3D {\SemSeg} methods can be roughly divided into three categories: multiview image-based, voxel-based, and point-based. Our method falls into the point-based category.

Although the \emph{multiview image-based} methods
{easily}
benefit from 
the success of 2D CNN \cite{boulch2017unstructured,lawin2017deep}, for {\SemSeg}, they suffer from occluded surfaces {and} density variations{,} and rely heavily on viewpoint selection. 
{Based on}
powerful 3D CNN 
\cite{roynard2018classification,ben20183dmfv,le2018pointgrid,meng2018vvnet,riegler2017octnet,graham20183d,Choy_2019}, \emph{voxel-based} methods achieve the best performance on several 3D {\SemSeg} datasets, but {they need intensive computation power}.

{{Compared}
with the previously mentioned methods, \emph{point-based} methods suffer less from information loss and {thus} achieve high point-level accuracy with less computation power consumption \cite{qi2017pointnet,qi2017pointnet++}.}
They can be generally classified into four categories:  neighboring feature pooling \cite{li2018so,Huang_2018,zhao2019pointweb,zhang2019shellnet}, graph construction \cite{wang2019dynamic,wang2019graph,jiang2019hierarchical,liu2019dynamic}, attention-based aggregation \cite{xie2018attentional}{,} and kernel-based convolution \cite{Su_2018,hua2018pointwise,wu2019pointconv,Lei_2019,Komarichev_2019,Lan_2019,mao2019interpolated,thomas2019kpconv}. {Among all the point-based methods, the recently proposed kernel-based method KPConv \cite{thomas2019kpconv}
achieves the best performance for efficient 3D convolution. Thus, we adopt KPConv
to build our backbone and refinement network.}

\subsection{2D Semantic Edge Detection} \label{2D SED}

Learning-based {\SemEdgeD} dates back to the {early} work of Prasad et al. \cite{prasad2006learning}.
Later, Hariharan et al. \cite{hariharan2011semantic} introduced the first Semantic Boundaries Dataset. After the dawn of deep learning, {the HFL method}
\cite{Bertasius_2015} 
builds a two-stage prediction process by using two deep CNNs for edge localization and classification, respectively. 
More recently, CASENet \cite{yu2017casenet} extended the CNN-based class-agnostic edge detector HED \cite{xie2015holistically} to {a} class-aware semantic edge detector
{by combining} 
low\hb{-} and high-level features with a multi-label loss function for supervision. 
Later, several follow-up works \cite{liu2018semantic,yu2018simultaneous,acuna2019devil,hu2019panoptic} improved CASENet by adding diverse deep supervision and reducing annotation noises.

{In {2D images},
semantic edges of different classes are weakly related since they are {essentially} occlusion boundaries of projected objects. Based on this observation, 2D {\SemEdgeD} methods treat {\SemEdgeD} of different classes as independent binary classification problems and utilize structures that limit the interaction between different classes like group convolution {modules}. 
Unlike the ones in 2D, semantic edges in 3D are physical boundaries of objects and thus {are highly related to each other}.
In this work, we study the problem of 3D {\SemEdgeD} for the first time.
We {adopt from} 
2D methods the idea of extracting enhanced features 
and construct a network that {does not limit}
the interaction between different classes.}

\subsection{Joint Learning of Segmentation and Edge Detection}

For 2D images, several works have explored the idea of combining networks for complementary segmentation and edge detection tasks to improve the learning efficiency, prediction accuracy, and generalization ability \cite{cheng2017fusionnet,bertasius2016semantic,peng2017large,su2019selectivity,lin2017refinenet,wu2019stacked,takikawa2019gated}. 

To be more specific, for salient object detection, researchers have exploited the duality between the binary segmentation and \emph{class-agnostic} edge detection tasks \cite{su2019selectivity,wu2019stacked}. 
As for {\SemSeg}, such \emph{class-agnostic} edges are used to build semantic segmentation masks with finer boundaries \cite{takikawa2019gated,peng2017large,lin2017refinenet,bertasius2016semantic,cheng2017fusionnet}. In contrast, we tackle the problem of joint learning for {\SemSeg} and \emph{class-aware} {\SemEdgeD}. Furthermore, {unlike previous works, which limit the interactions between segmentation and edge detection to the sharing of features and network structures, our method}
exploits the close relationship between {\SemSegPoint} masks and {\SemEdgePoint} maps.

%% file: method.tex
\pdfoutput=1
\section{{JSENet}} \label{JSEnet}

\begin{figure*}[ht]
    \centering
    \includegraphics[width=\linewidth]{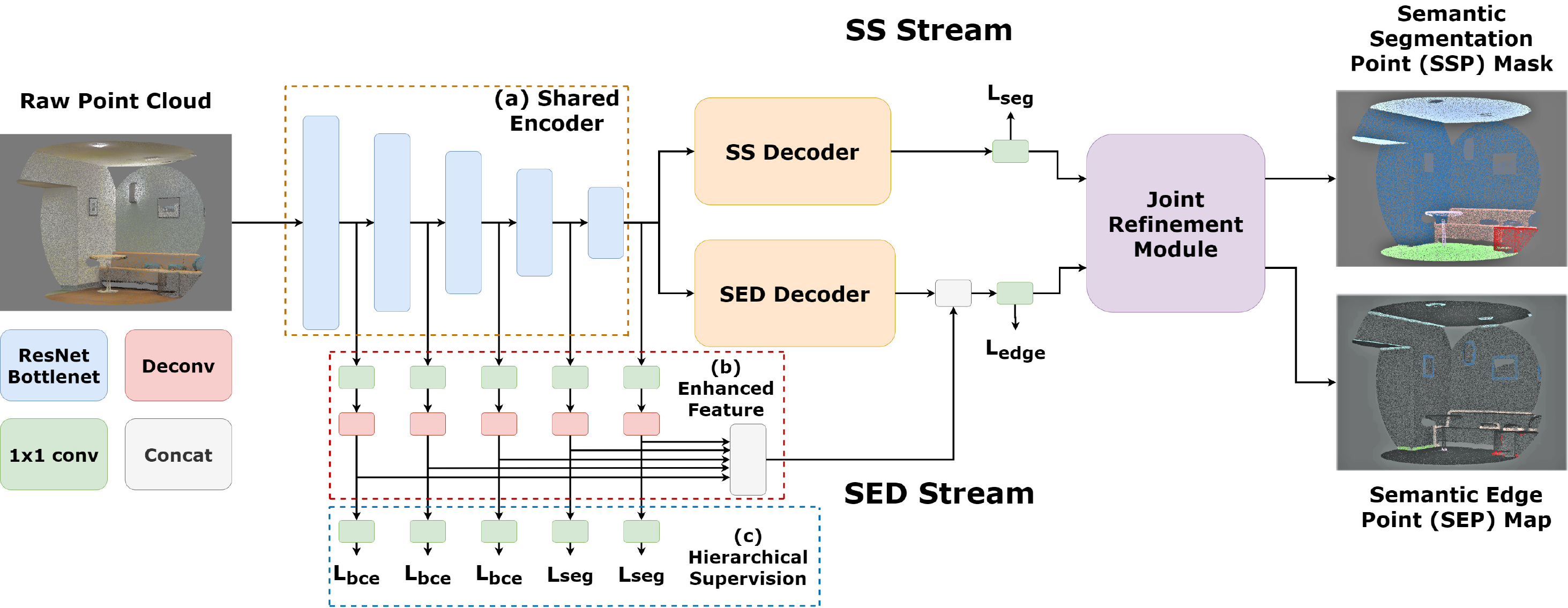}
    \caption{
    JSENet architecture. Our architecture consists of two main streams. The {{\SemSeg}} stream can be any fully-convolutional network for {\SemSeg}. The {{\SemEdgeD}} stream extracts enhanced features through a skip-layer structure and is supervised by multiple loss functions. A joint refinement module later combines {the} information from the two streams and outputs refined {\SemSegPoint} masks and {\SemEdgePoint} maps.
    }
    \label{fig:network}
\end{figure*}

In this section, we present our JSENet architecture for the joint learning of {\SemSeg} and {class-aware} {\SemEdgeD}.
As depicted in Fig. \ref{fig:network}, our architecture consists of two streams of networks with a shared feature encoder and followed by a joint refinement module. The first stream of the network{, the {\SemSeg} stream},  
is a {standard encoder-decoder FCN for {\SemSeg}}
using the same structure as presented in KPConv \cite{thomas2019kpconv}.
{The second stream,
the {\SemEdgeD} stream, is} {another} 
FCN with enhanced feature extraction and hierarchical supervision. We then fuse the outputs from {the two streams}
using our carefully designed {joint} refinement module to produce refined {\SemSegPoint} masks and refined {\SemEdgePoint} maps. 
Next, we will describe each of the modules 
in detail and then explain the supervisions used for joint learning.

\subsection{{Semantic} Segmentation Stream} \label{Segmentation stream}

We denote the 
{{\SemSeg} stream} as $\mathcal{S_\theta(\mathcal{P})}$ with parameters $\theta$, {taking a}
point cloud $\mathcal{P} \in \mathbb{R}^{N \times 6}(x,y,z,r,g,b)$ with $N$ points as input and outputting {an} {\SemSegPoint} mask.
More {specifically},
for a segmentation prediction of $K$ semantic classes, it outputs a categorical distribution $s \in \mathbb{R}^{N \times K}${:}
$s(p|\mathcal{P},\theta)$ 
{representing} the probability of point $p$ 
{belonging} to each of the K classes. 
We supervise this stream with the standard multi-class cross-entropy loss ({$L_{seg}$ in Fig. \ref{fig:network}}). The {{\SemSeg}} stream can be any feedforward 3D fully-convolutional {\SemSeg} network such as InterpCNNs \cite{mao2019interpolated} and SSCNs \cite{graham20183d}. In this work, we adopt KPConv \cite{thomas2019kpconv} and use it as our backbone network for its {efficient convolution operation} and ability to build complex network structures. {There are two types of KPConv: rigid and deformable. We use the rigid version in this work for its better convergence performance.} In order to demonstrate the improvements introduced by joint learning, we use the same structure as described in KPConv.

\subsection{{Semantic} Edge Detection Stream
} \label{Semantic edge detection stream}
We denote the {{\SemEdgeD}} stream as $\mathcal{E_\phi(\mathcal{P})}$ with parameters $\phi$, taking a point cloud {$\mathcal{P}$} with $N$ points as input and outputting {\SemEdgePoint} maps. For 
$K$ defined semantic categories, {this} stream output{s} $K$ {{\SemEdgePoint}} maps $\{e_1,...,e_K\}$, each having the same size as $\mathcal{P}$ {with one channel}. 
We denote $e_k(p|\mathcal{P},\phi)$ as the network output, which indicates the computed edge probability 
{of}
the $k$-th semantic category at point $p$. Note that one point
{may} belong to multiple categories.

\begin{figure*}[ht]
    \centering
    \includegraphics[width=\linewidth]{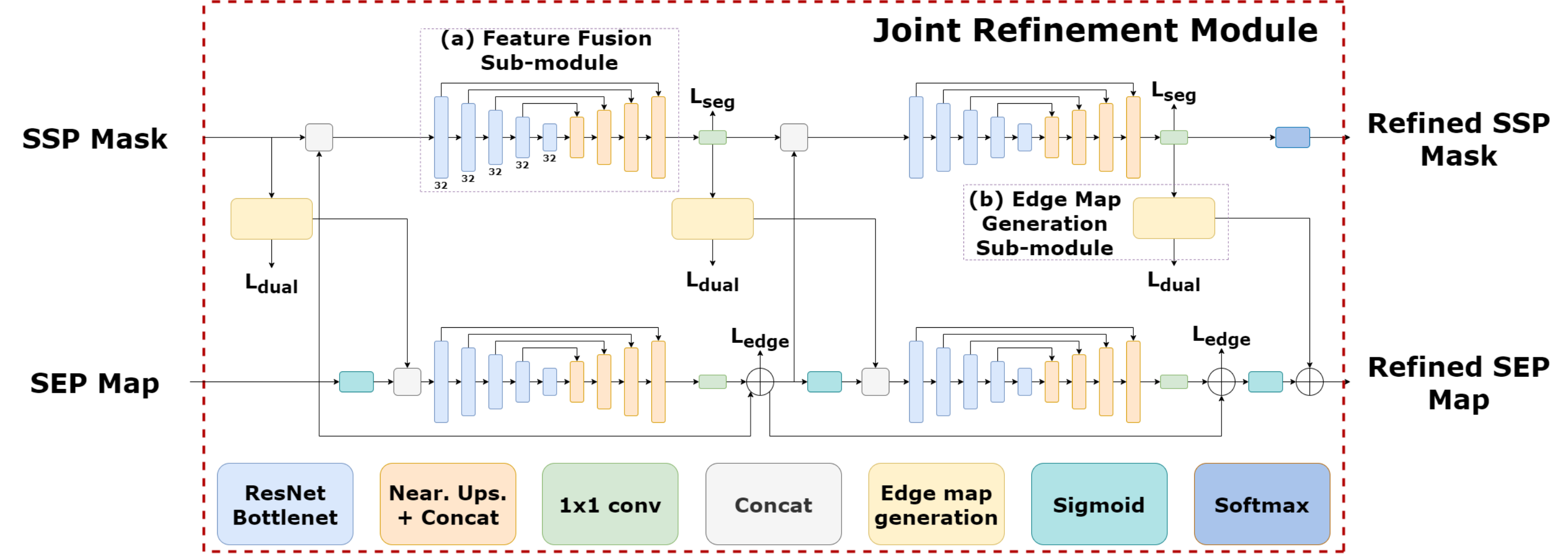}
    \caption{
    {Illustration of the joint refinement module, which} 
    consists of two branches.
    }
    \label{fig:joint}
\end{figure*}

The {{\SemEdgeD}} stream shares the same feature encoder with the {{\SemSeg}} stream to force the network to prefer representations with {a} better generalization ability for both tasks. Our backbone is an FCN-based structure. However, one {major} 
drawback of {an} FCN-based structure is the loss in spatial information of the output after propagating through several alternated convolutional and pooling layers. {This drawback would} 
harm the localization performance, which is essential for our {\SemEdgeD} task. Besides, according to the findings of CASENet \cite{yu2017casenet}, the {low}{-}level
features of CNN-based encoder are not suitable for semantic classification (due to the limited receptive fields) but {are able to} help augment top classifications by suppressing non-edge activations and providing detailed edge localization information. Based on these observations, we propose to extract enhanced features with hierarchical supervisions 
to alleviate the problem of spatial information loss and offer richer semantic cues for final classification.

{
\smallskip \noindent \textbf{Enhanced feature extraction.}
In detail, we extract the feature maps generated by the layers of the shared encoder (Fig. \ref{fig:network}(a)). We then reduce the numbers of their feature channels and deconvolve them to the size of the input point cloud. The features of different layers participate and augment the final {\SemEdgeD} through a skip-layer architecture, as shown in Fig. \ref{fig:network}(b).

\smallskip \noindent \textbf{Hierarchical supervision.}
As shown in Fig. \ref{fig:network}(c), from the extracted feature maps of the first three layers, we generate binary edge point maps indicating the probability of points belonging to the semantic edges of any classes. From the last two layers, we generate two {\SemSegPoint} masks. 
All binary edge point maps are supervised by the weighted binary cross-entropy loss ($L_{bce}$) using ground-truth (GT) binary edges obtained from {the} GT {\SemSegPoint} masks. The two {\SemSegPoint} masks are supervised by the standard multi-class cross-entropy loss ($L_{seg}$). The output {\SemEdgePoint} maps of the {{\SemEdgeD}} stream are supervised by a weighted multi-label loss ($L_{edge}$) following the idea from CASENet \cite{yu2017casenet}. We will give details about different loss functions in Section \ref{Joint multi-task learning}.
}

%% file: method_1.tex
\pdfoutput=1
\subsection{Joint Refinement Module} \label{Joint refinement module}

We denote the joint refinement module as $\mathcal{R_\gamma}(s, e_1,...,e_K)$ with parameters $\gamma$, taking 
as input the {\SemSegPoint} mask $s$ coming from the {\SemSeg} stream and the {\SemEdgePoint} maps $\{e_1,...,e_K\}$ generated by the {\SemEdgeD} stream. As shown in Fig. \ref{fig:joint}, we construct a two-branch structure with simple feature fusion sub-modules (Fig. \ref{fig:joint}(a)) and novel edge map generation sub-modules (Fig. \ref{fig:joint}(b)). The upper branch is responsible for segmentation refinement, and the lower one is for edge refinement. We feed different joint features (described in detail below) to the feature fusion sub-modules of the two branches and generate refined {\SemSegPoint} masks and {\SemEdgePoint} maps.

\smallskip \noindent \textbf{Feature fusion {sub-module}.} To fuse the region and edge features, we construct two simple U-Net \cite{Ronneberger_2015} like feature fusion {sub-modules} for each refinement branch. In detail, each feature fusion {sub-module} consists of five encoding layers with channel sizes of 32 for all layers. For segmentation refinement, the feature fusion {sub-module} takes the concatenated {\SemSegPoint} mask $s \in \mathbb{R}^{N \times K}$ and {\SemEdgePoint} maps $\{e_1,...,e_K\}$ $(e_i \in \mathbb{R}^{N})$ as input and outputs a refined {\SemSegPoint} mask directly. As for edge refinement, we find that adjusting the activation values of the {\SemEdgePoint} maps is more effective than asking the neural network to output refined {\SemEdgePoint} maps na\"{i}vely. Thus, we put the unrefined {\SemEdgePoint} maps through a sigmoid operation and concatenate them with the edge activation point maps $\{a_{1},...,a_{K}\}$ $(a_i \in \mathbb{R}^{N})$ generated by the edge map generation {sub-module}. The concatenated features are then fed to the feature fusion {sub-module} to generate auxiliary point maps, which are added to the unrefined {\SemEdgePoint} maps to adjust the edge activation values.

\begin{figure*}[ht]
    \centering
    \includegraphics[width=\linewidth]{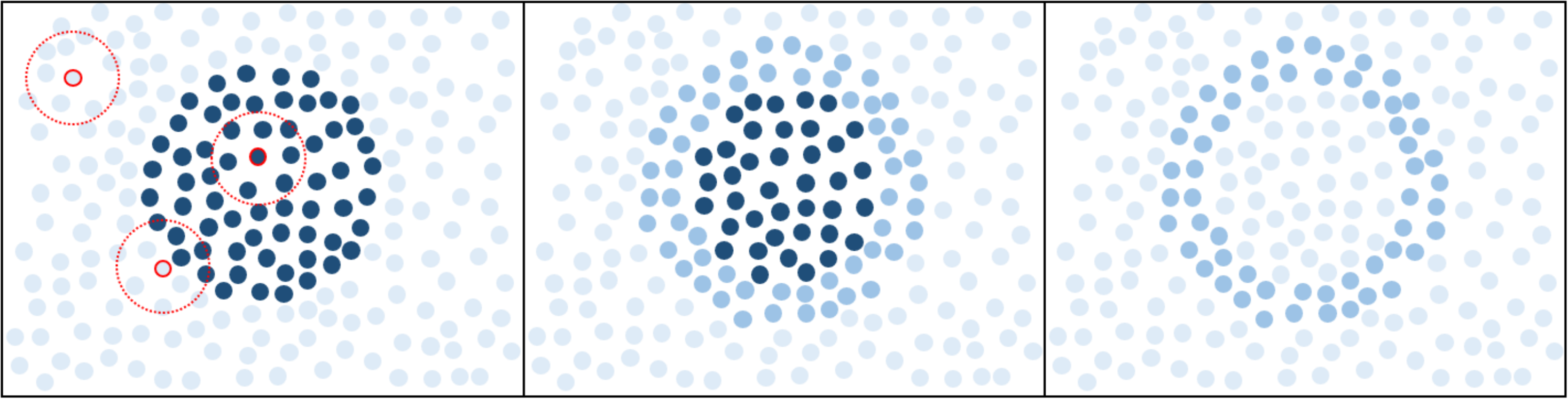}
    \caption{Illustration of the edge map generation process on 2D points. (Left) An activation point mask, {with dark colors representing} high activation values. Three red points represent three different situations: far from the activated region, near the boundary, and within the activated region. (Middle) The activation point mask after the mean filtering; (Right) The generated edge activation point map.}
    \label{fig:emg}
\end{figure*}

\smallskip \noindent \textbf{Edge map generation {sub-module}.} In order to exploit the duality between two tasks, we design an edge map generation {sub-module} to convert a{n} {\SemSegPoint} mask to edge activation point maps. More formally, we denote the edge map generation {sub-module} as $\mathcal{G}(s)$, {which} takes a categorical distribution $s \in \mathbb{R}^{N \times K}$ as input and outputs edge activation point maps $\{a_{1},...,a_{K}\}$ $(a_i \in \mathbb{R}^{N})$, where $a_i(p)$ is a potential that represents whether a particular point $p$ belongs to the semantic boundary of the $i$-th class. It is computed by taking a spatial derivative on the segmentation output as follows:  
    \begin{equation}
        a_i = col_i(\lvert M * Softmax(s) - Softmax(s) \rvert),
        \label{eqn0}
    \end{equation} 
where {$col_i$ denotes the $i$-th column and} $M$ denotes the Mean filter, which takes neighboring points within a small radius. {As illustrated in Fig. \ref{fig:emg}, the $i$-th column of the output tensor of {a} softmax operation represents an activation point mask for class\_$i${,} where a higher value indicates a higher probability of belonging to class\_$i$. On this mask, points near to the boundaries of class\_$i$ will have neighbors with activation values of significant differences, and points far from the boundaries will have neighbors with similar activation values. Thus, after the mean filtering and subtraction, points nearer to the predicted boundaries will have larger activation values.}
The converted edge activation point maps are fed to the edge refinement branch and utilized for loss calculation as well.

\smallskip \noindent \textbf{Supervisions.}
For supervision, the output {\SemSegPoint} masks are supervised by the multi-class cross-entropy loss ($L_{seg}$), and the output {\SemEdgePoint} maps are supervised by the weighted multi-label loss ($L_{edge}$). Additionally, we supervise the generated edge activation point maps with {our proposed} dual semantic edge loss ($L_{dual}$) 
to encourage the predicted {\SemSeg} results to align with the GT semantic edges correctly (see Section \ref{Joint multi-task learning}). After refinement, the final output {\SemSegPoint} mask is normalized by a softmax operation. The final output {\SemEdgePoint} maps are normalized by a sigmoid operation and added element-wise with the final edge activation point maps.

%% file: method_2.tex
\pdfoutput=1
\subsection{Joint Multi-task Learning} \label{Joint multi-task learning}

The key to joint learning of the {\SemSeg} and the {\SemEdgeD} tasks is {to design} 
a proper supervision signal. The total loss is formulated as:
    \begin{equation}
        L_{total} = \lambda_0 L_{seg} + \lambda_1 L_{edge} + \lambda_2 L_{bce} + \lambda_3 L_{dual}.
        \label{eqn1}
    \end{equation}
During training, $\lambda_0$ is set to the number of semantic classes to balance the influences of the two tasks. The other weights are set to 1.
We describe in detail all the loss functions used for supervision in our framework below. 

\smallskip \noindent \textbf{Multi-class cross-entropy loss.}
We use {a} standard multi-class cross-entropy loss, which denoted as $L_{seg}$, on {a} predicted {\SemSegPoint} mask $s$:
    \begin{equation}
        L_{seg}(\hat{s}, s) = -\sum_k\sum_p\hat{s}_k(p)\log(s_k(p)),
        \label{eqn2}
    \end{equation}
where $\hat{s} \in \mathbb{R}^{N \times K}$ denotes GT semantic labels in {a} one-hot form.

\smallskip \noindent \textbf{Weighted multi-label loss.}
Following the idea proposed in CASENet \cite{yu2017casenet}, we address the {\SemEdgeD} problem {for a point cloud}
by a multi-label learning framework and implement a point-wise weighted multi-label loss $L_{edge}$. Suppose an input point cloud $\mathcal{P}$ has $K$ {\SemEdgePoint} maps $\{e_1,...,e_K\}$ $(e_i \in \mathbb{R}^{N})$ predicted by the network and $K$ label point maps $\{\hat{e}_1,...,\hat{e}_K\}$ $(\hat{e}_i \in \mathbb{R}^{N})$, where $\hat{e}_k$ is a binary point map indicating the ground truth of the $k$-th class semantic edges. The point-wise weighted multi-label loss $L_{edge}$ is formulated as:
    \begin{multline}
         L_{edge}(\{\hat{e}_1,...,\hat{e}_K\},  \{e_1,...,e_K\}) = \sum_k\sum_p\{-\beta_k \hat{e}_k(p)\log(e_k(p)) -\\ 
        (1 - \beta_k)(1 - \hat{e}_k(p))\log(1 - e_k(p))\},
        \label{eqn3}
    \end{multline}
where $\beta_k$ is the percentage of non-edge points in the point cloud of the $k$-th class to account for {the} skewness of sample numbers.

\smallskip \noindent \textbf{Weighted binary cross-entropy loss.}
To supervise the generated binary edge point maps in the {\SemEdgeD} stream, we implement a point-wise weighted binary cross-entropy loss $L_{bce}$. Let $b$ denote
the predicted binary edge point map, 
and $\hat{b}$ the GT binary edge point map converted from the GT {\SemEdgePoint} maps. The point-wise weighted cross-entropy loss is defined as:
    \begin{equation}
        L_{bce}(\hat{b}, b) = \sum_p\{-\beta \hat{b}(p)\log(b(p)) - (1 - \beta)(1 - \hat{b}(p))\log(1 - b(p))\},
        \label{eqn4}
    \end{equation}
where $\beta$ is the percentage of non-edge points among all classes.

\smallskip \noindent \textbf{Dual semantic edge loss.}
As mentioned above, inspired by the duality between {\SemSeg} and {\SemEdgeD}, we design an edge map generation {sub-module} to convert the predicted {\SemSegPoint} mask $s \in \mathbb{R}^{N \times K}$ to edge activation point maps $\{a_{1},...,a_{K}\}$ $(a_i \in \mathbb{R}^{N})$(\textit{c.f.}, Equation~\ref{eqn0}). In {a}
similar way, we can compute GT edge activation point maps $\{\hat{a}_{1},...,\hat{a}_{K}\}$ from the GT semantic labels $\hat{s}$:
    \begin{equation}
        \hat{a}_i = col_i(\lvert M * One\_hot(\hat{s}) - One\_hot(\hat{s}) \rvert).
        \label{eqn5}
    \end{equation} 
Note that the softmax operation for predicted {\SemSegPoint} mask $s$ is changed to the one-hot encoding operation for GT semantic labels $\hat{s}$. Taking the converted GT edge activation point maps, we can {define the loss function as follows:}
    \begin{equation}
        \begin{split}
            L_{dual}(\{\hat{a}_1,...,\hat{a}_K\},  \{a_1,...,a_K\}) & = \sum_k\sum_p \beta (\lvert \hat{a}_k(p) - a_k(p) \rvert),
        \end{split}
        \label{eqn6}        
    \end{equation}
{where $\beta$ is the same weight as above. Intuitively, the network will get penalized when there are mismatches on edge points.}
{It is worth noting that the loss function will not 
be dominated by the non-edge points {since the calculated loss values on these points are zeros or very small numbers}.} The above dual loss is naturally differentiable and exploits the duality between {\SemSeg} and {\SemEdgeD}.

%% file: experiment.tex
\pdfoutput=1
To demonstrate the effectiveness of our proposed method, we now present various experiments conducted on the S3DIS \cite{armeni_cvpr16} and ScanNet\cite{dai2017scannet} datasets, for which GT {\SemSegPoint} masks are available and we can generate GT {\SemEdgePoint} maps from them easily.
{We first introduce the dataset preparation and evaluation metrics in Section \ref{Datasets}, and then present the implementation details for reproduction in Section \ref{Implementation}. We report the results of our ablation studies in Section \ref{Ablation study}, and the results on the S3DIS and ScanNet datasets in Sections \ref{S3DIS}.}

\subsection{Datasets and Metrics} \label{Datasets}

We use S3DIS \cite{armeni_cvpr16} and ScanNet \cite{dai2017scannet} datasets for our experiments. {There are two reasons for choosing these datasets: 1) they are both of high quality and 2) semantic edges are better defined on indoor data: compared to existing 3D outdoor datasets, in indoor scenes, more detailed semantic labels are defined and objects are more densely connected}.
The S3DIS dataset consists of 3D point clouds for six large-scale indoor areas captured from three different buildings. It has around 273 million points annotated with 13 semantic classes. The ScanNet dataset includes 1513 training scenes and 100 test scenes in a mesh format, 
all annotated with 20 semantic classes, for online benchmarking.

To generate the GT {\SemEdgePoint} maps, for S3DIS, we directly check the neighbors within a $2cm$ radius of each point in the dataset. For a specific point, if it has neighbors
with different semantic labels, we label it as an edge point of all the semantic classes that appear in its neighborhood. As for the ScanNet dataset, 
following the work of KPConv \cite{thomas2019kpconv}, we first rasterize the training meshes by uniformly sampling points on the faces of meshes 
and then downsample the generated point clouds with $1cm$ grids. {We use these point clouds for training and generate {the} GT {\SemEdgePoint} maps using the same way as described for the S3DIS dataset.}
During testing, we project the semantic edge labels to the vertices 
of the original meshes and test directly on meshes.

To evaluate the performance of {{\SemSeg} and {\SemEdgeD}}, 
we {adopt} 
the standard mean intersection over union (mIoU) for {\SemSeg} and use the mean maximum F-measure (MF) at the optimal dataset scale (ODS) for {\SemEdgeD} following the works in 2D \cite{yu2017casenet,liu2018semantic,yu2018simultaneous,acuna2019devil}. 
We generate thicker edges for point clouds than for images since a point cloud is much sparser than an image. 
{Since we have thicker edges, the localization tolerance used in the 2D case
is not introduced to our evaluation.}

\subsection{Implementation Details} \label{Implementation}
In this section, we discuss the implementation details for our experiments. JSENet is
coded in Python and TensorFlow. All the experiments are conducted on {a PC with 8 Intel(R) i7-7700 CPUs and} a single GeForce GTX 1080Ti GPU. 

\smallskip \noindent \textbf{Training.}
Since the 3D scenes in {both}
datasets are {of huge size}, 
we randomly sample spheres with $2m$ radius in the training set and augment them with Gaussian noise, random scaling, and random rotation. Following the settings in KPConv, the input point clouds are downsampled with a grid size of $4cm$.
In all our experiments, unless explicitly stated otherwise, we use a Momentum gradient descent optimizer with a momentum of 0.98 and an initial learning rate of $0.01$. The learning rate is scheduled to decrease exponentially. In particular, it {is} 
divided by $10$ {for} every $100$ epochs.
Although the framework is end-to-end trainable, in order to clearly demonstrate the efficacy of the proposed joint refinement module, we first train our network without the joint refinement module for 350 epochs and then optimize the joint refinement module alone with {the} other parts fixed for 150 epochs.

\begin{table}
\small
\begin{center}
\caption{Ablation experiments of network structures on S3DIS Area-5. \textbf{SEDS}: semantic edge detection stream; \textbf{EFE}: enhanced feature extraction; \textbf{HS}: hierarchical supervision; \textbf{SSS}: semantic segmentation stream; \textbf{JRM}: joint refinement module. {The results in some cells (with `-') are not available, since the corresponding models perform either SS or SED.}
}
\label{table:ablation-structures}

\begin{tabular}{ | C{0.5cm} | C{1cm} | C{1cm} | C{1cm} | C{1cm} | C{1cm} | C{1.5cm} | C{2.5cm} |}
    \hline
    0 & SEDS & EFE & HS & SSS & JRM & mIoU (\%) & mMF (ODS)(\%)\\
    \hline
    1 & \checkmark & \checkmark & \checkmark & \checkmark & \checkmark & 67.7 & 31.0\\
    \hline
    2 & \checkmark & \checkmark & \checkmark & \checkmark &  & 66.2 & 30.5\\
    \hline
    3 &  &  &  & \checkmark &  & 64.7 & -\\
    \hline
    4 & \checkmark & \checkmark & \checkmark &  &  & - & 30.2\\
    \hline
    5 & \checkmark & \checkmark &  &  &  & - & 29.9\\
    \hline
    6 & \checkmark &  &  &  &  & - & 29.4\\
    \hline
    \end{tabular}
\end{center}
\end{table}

\smallskip \noindent \textbf{Testing.}
Similar to the training process, during testing, we sample spheres with $2m$ radius from the testing set regularly 
and ensure each point to be sampled {for} multiple times. The predicted probabilities for each point are averaged through a voting scheme \cite{thomas2019kpconv}.
All predicted values are projected to the original point clouds {(S3DIS)} or meshes {(ScanNet)} for evaluation.

\subsection{Ablation Study} \label{Ablation study}

In this section, we compare the performances of JSENet under different settings on the S3DIS dataset since it is originally presented in a point cloud format and all semantic labels are available. Following the common setting {\cite{qi2017pointnet,tchapmi2017segcloud,thomas2019kpconv,li2018pointcnn,jiang2019hierarchical,tatarchenko2018tangent,Choy_2019}}, we use Area-5 as {a} test scene and train our network on the other scenes. All experiments are conducted keeping all hyperparameters the same.

\smallskip \noindent \textbf{Network structures.}
In Table \ref{table:ablation-structures}, we evaluate the effectiveness of each component of our method.
For the {\SemSeg} task, as shown in the table (Row 3),
the performance of training our {\SemSeg} stream alone
is $64.7\%$ in terms of mIoU. Our {\SemSeg} stream shares the same architecture {with} KPConv and the reported score of KPConv is $65.4\%$ in their paper. By na\"{i}vely
combining the {\SemSeg} stream and the {\SemEdgeD} stream, we can improve the {\SemSeg} task by $1.5\%$ {(Row 2)}. From the joint refinement module, we further gain about {$1.5\%$} {(Row 1)} improvement in performance. We achieve about $3\%$ improvement comparing to training our {\SemSeg} stream alone and still more than $2\%$ improvement comparing to the result reported in KPConv.  

For the {\SemEdgeD} task, it can be seen from the table that the performance of training the {\SemEdgeD} stream alone without the enhanced feature extraction and hierarchical supervision is $29.4\%$ (Row 6) in terms of mMF (ODS). We gain about {$0.5\%$} and $0.3\%$ improvements in performance from the enhanced feature extraction (Row 5) and hierarchical supervision (Row 4), respectively. 
Na\"{i}vely combining the {\SemSeg} stream and the {\SemEdgeD} stream brings a {further} improvement of $0.3\%$ {(Row 2)} in terms of mMF.
By adding the joint refinement module, we can further improve the {\SemEdgeD} task by $0.5\%$ {(Row 1)}.

\begin{table}[h]
\caption{(a) Comparison of different supervision choices for \SemEdgeD. (b) Effect{s} of the dual semantic edge loss in terms of boundary quality (F-score).}
    \begin{subtable}[t]{0.6\textwidth}
        \caption{}
        \centering
        \resizebox{\textwidth}{!}{
            \begin{tabular}{  | l | c |}
            \hline
            Method & mMF (ODS) (\%)\\
            \hline
            $L_{bce}$ for all five layers & 30.1\\
            \hline
            $L_{seg}$ for all five layers & 30.1\\
            \hline
            No hierarchical supervision & 29.9\\
            \hline
            $L_{bce}$ for first three, $L_{seg}$ for last two & 30.2\\
            \hline
            \end{tabular}}
       \label{table:ablation-HS}
    \end{subtable}
    \hfill
    \begin{subtable}[t]{0.38\textwidth}
        \caption{}
        \centering
        \resizebox{\textwidth}{!}{
        \begin{tabular}{  | l | c |}
            \hline
            Method & F-score (\%)\\
            \hline
            JSENet w/o dual loss & 22.7\\
            \hline
            JSENet & 23.1\\
            \hline
        \end{tabular}}
        \label{table:ablation-DL}
     \end{subtable}
     \label{tab:temps}
\end{table}


\smallskip \noindent \textbf{Choice of loss functions for hierarchical supervision.}
To justify our choices of {the} loss functions in {the} {\SemEdgeD} stream for hierarchical supervision, we test the performances of {\SemEdgeD} using different settings of supervision. As shown in Table \ref{table:ablation-HS}, if all hierarchical supervisions are removed, the performance of our {\SemEdgeD} model will decrease by $0.3\%$. Among all the choices listed in the table, the one used in our network achieves the best result.


\smallskip \noindent \textbf{Efficacy of dual semantic edge loss.}
We further showcase the effect{s} of the dual semantic edge loss in terms of F-score for edge alignment of the predicted {\SemSegPoint} masks in Table \ref{table:ablation-DL}. We train our network {without} 
the dual semantic edge losses {for}
the edge map generation sub-modules as the baseline. It is shown that the dual semantic edge losses bring an improvement of {$0.4\%$} in terms of F-score.

\begin{figure}
\captionsetup[subfigure]{labelformat=empty}
\begin{subfigure}{.193\textwidth}
  \centering
  \includegraphics[width=\textwidth, height=0.76\textwidth]{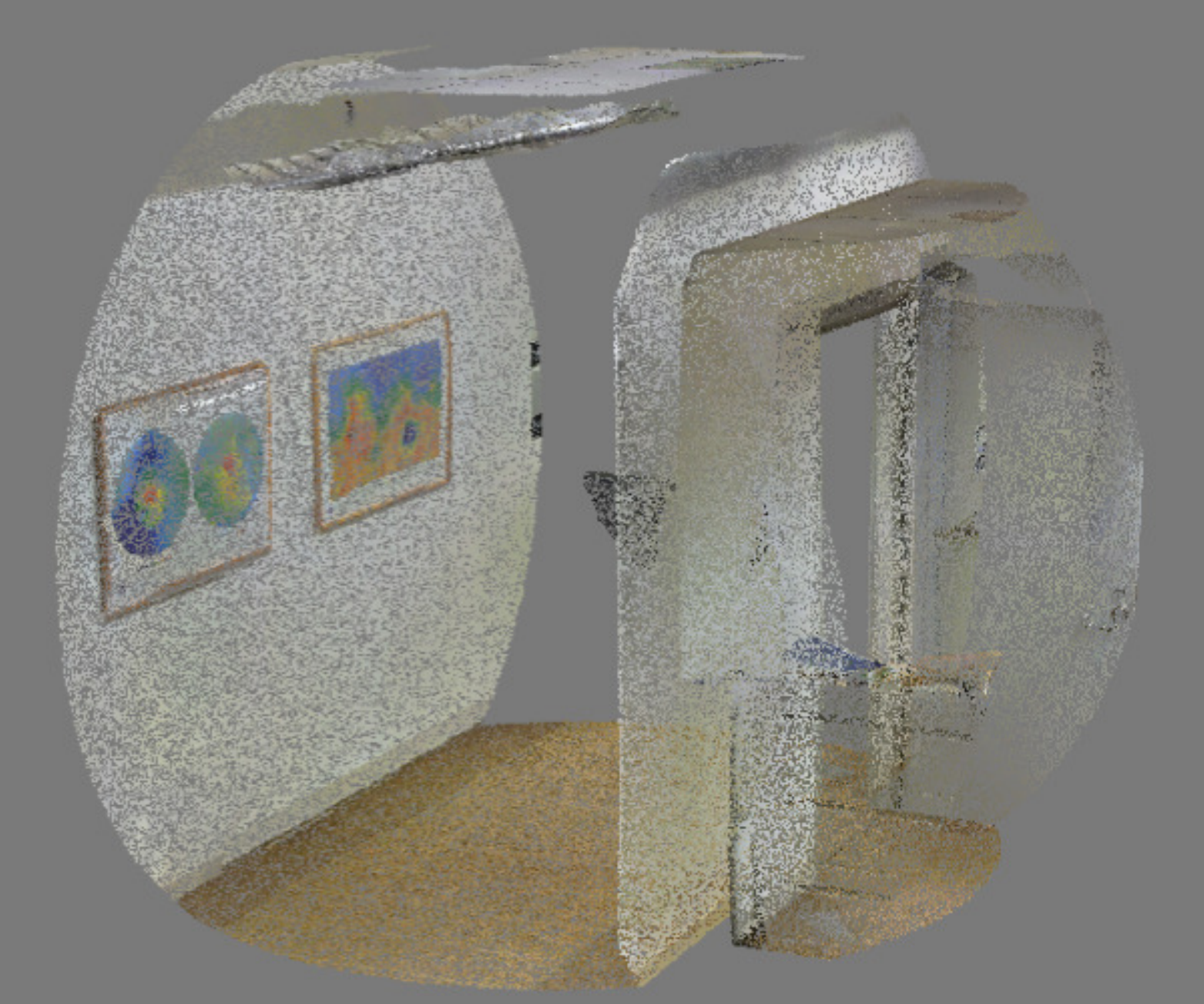}
\end{subfigure} \hfil
\begin{subfigure}{.193\textwidth}
  \centering
  \includegraphics[width=\textwidth, height=0.76\textwidth]{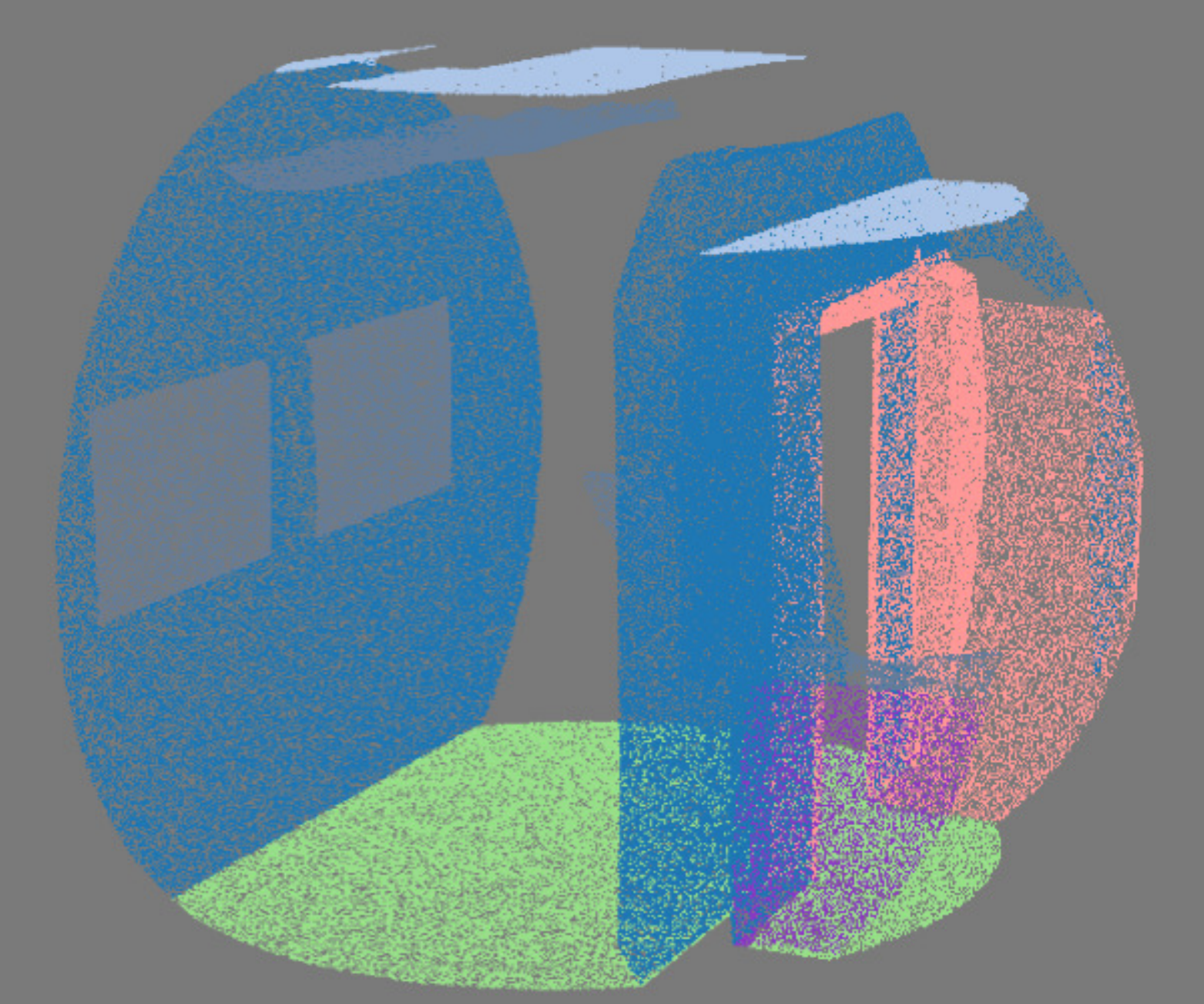}
\end{subfigure} \hfil
\begin{subfigure}{.193\textwidth}
  \centering
  \includegraphics[width=\textwidth, height=0.76\textwidth]{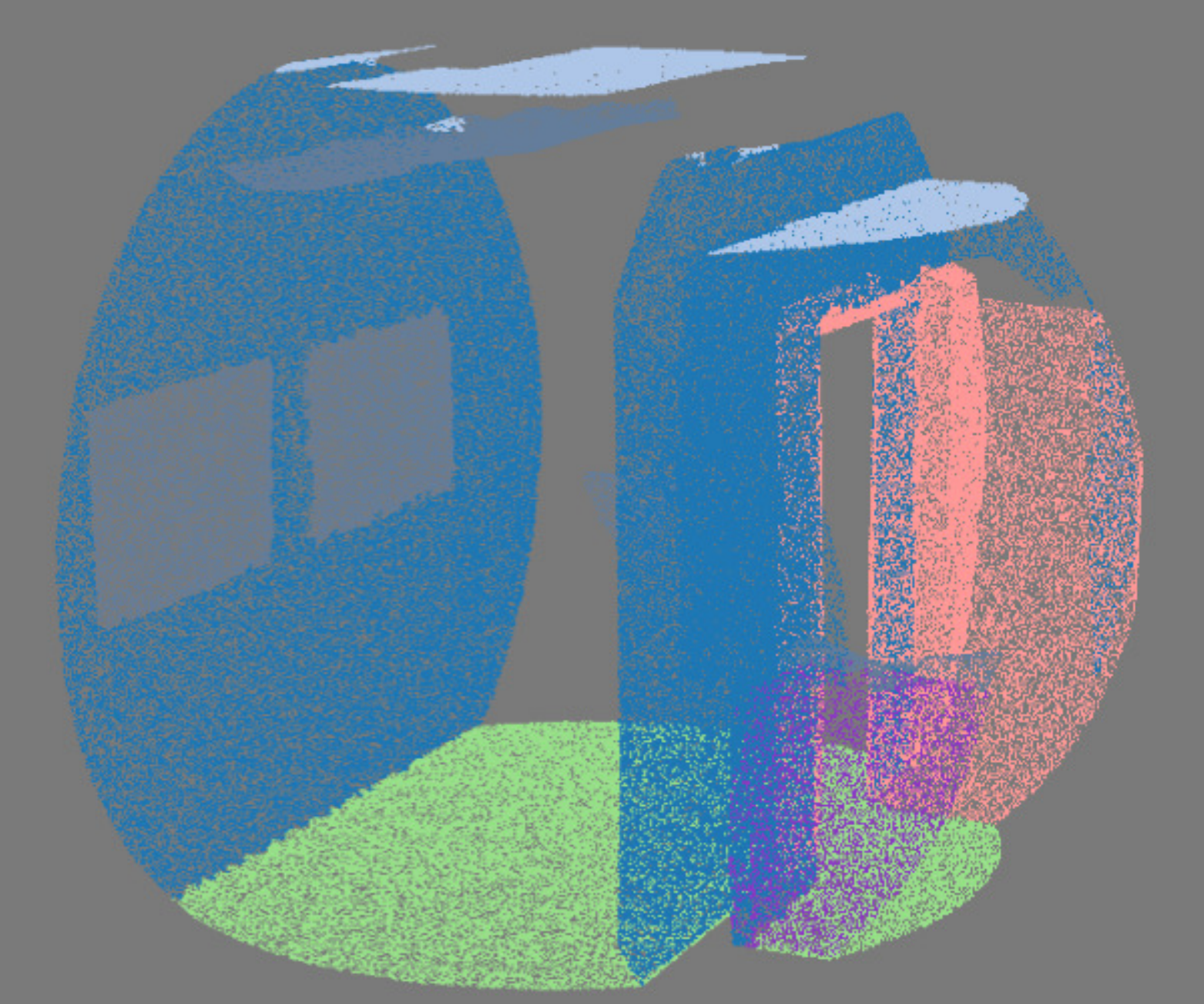}
\end{subfigure} \hfil
\begin{subfigure}{.193\textwidth}
  \centering
  \includegraphics[width=\textwidth, height=0.76\textwidth]{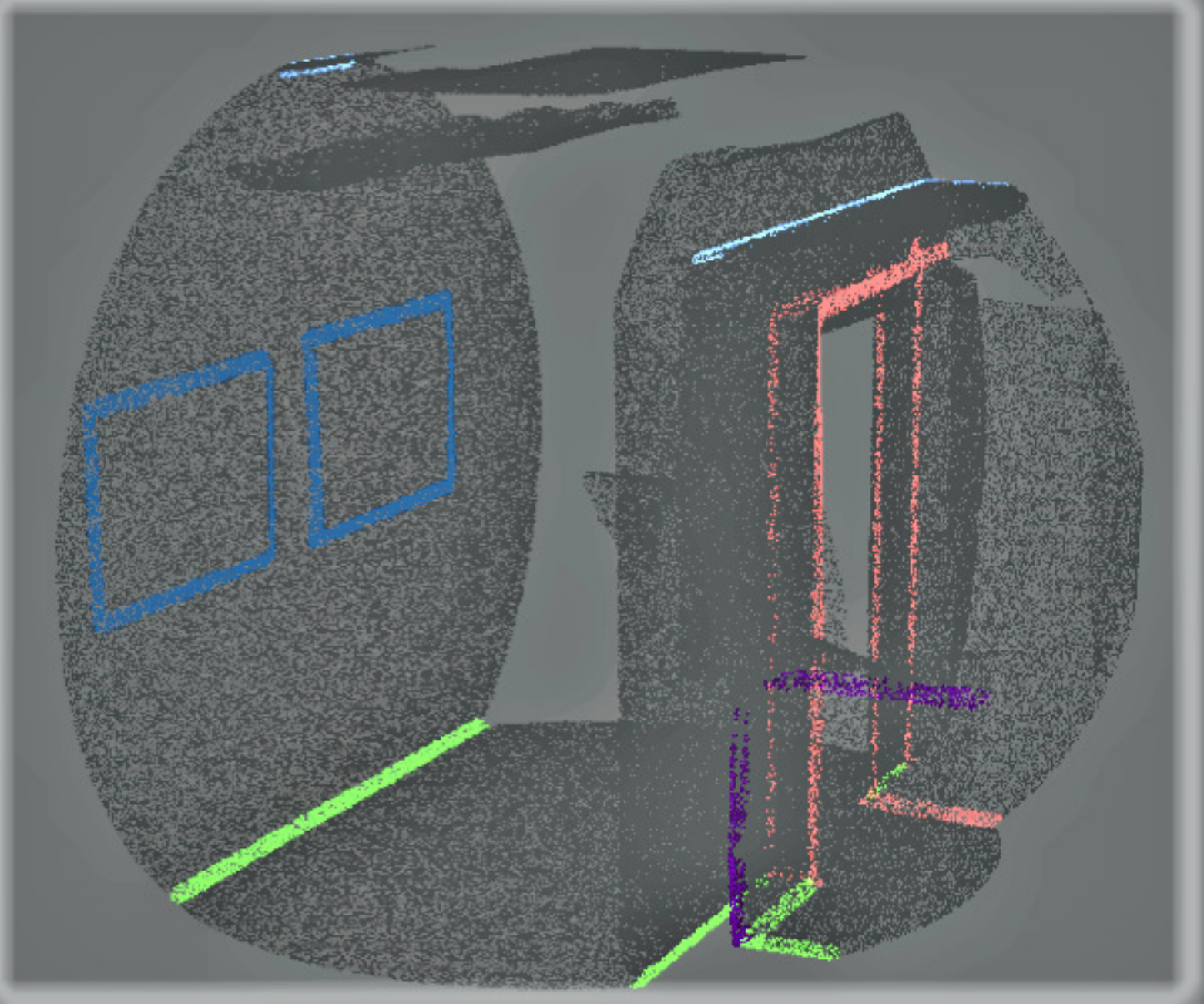}
\end{subfigure} \hfil
\begin{subfigure}{.193\textwidth}
  \centering
  \includegraphics[width=\textwidth, height=0.76\textwidth]{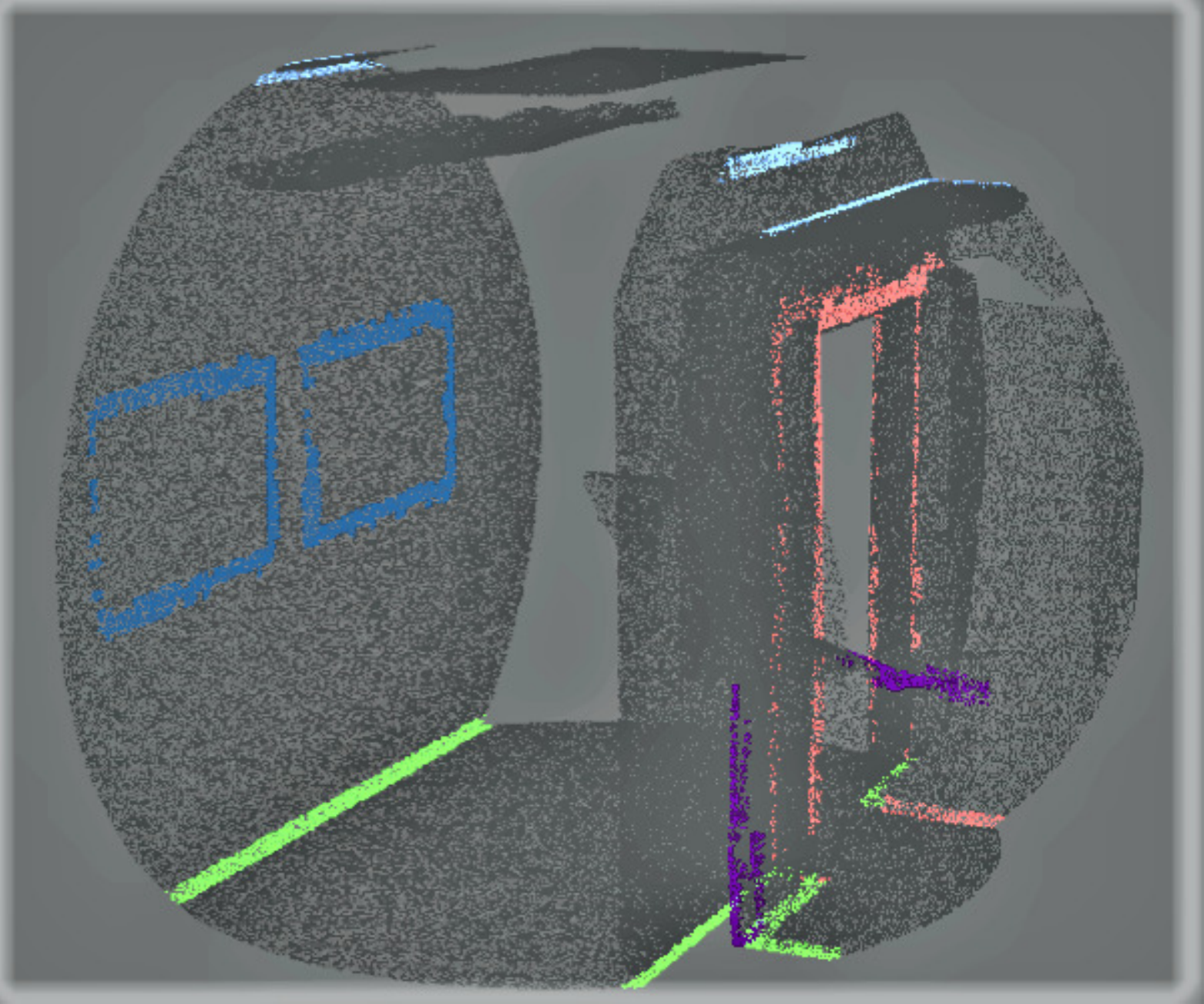}
\end{subfigure}

\begin{subfigure}{.193\textwidth}
  \centering
  \includegraphics[width=\textwidth, height=0.76\textwidth]{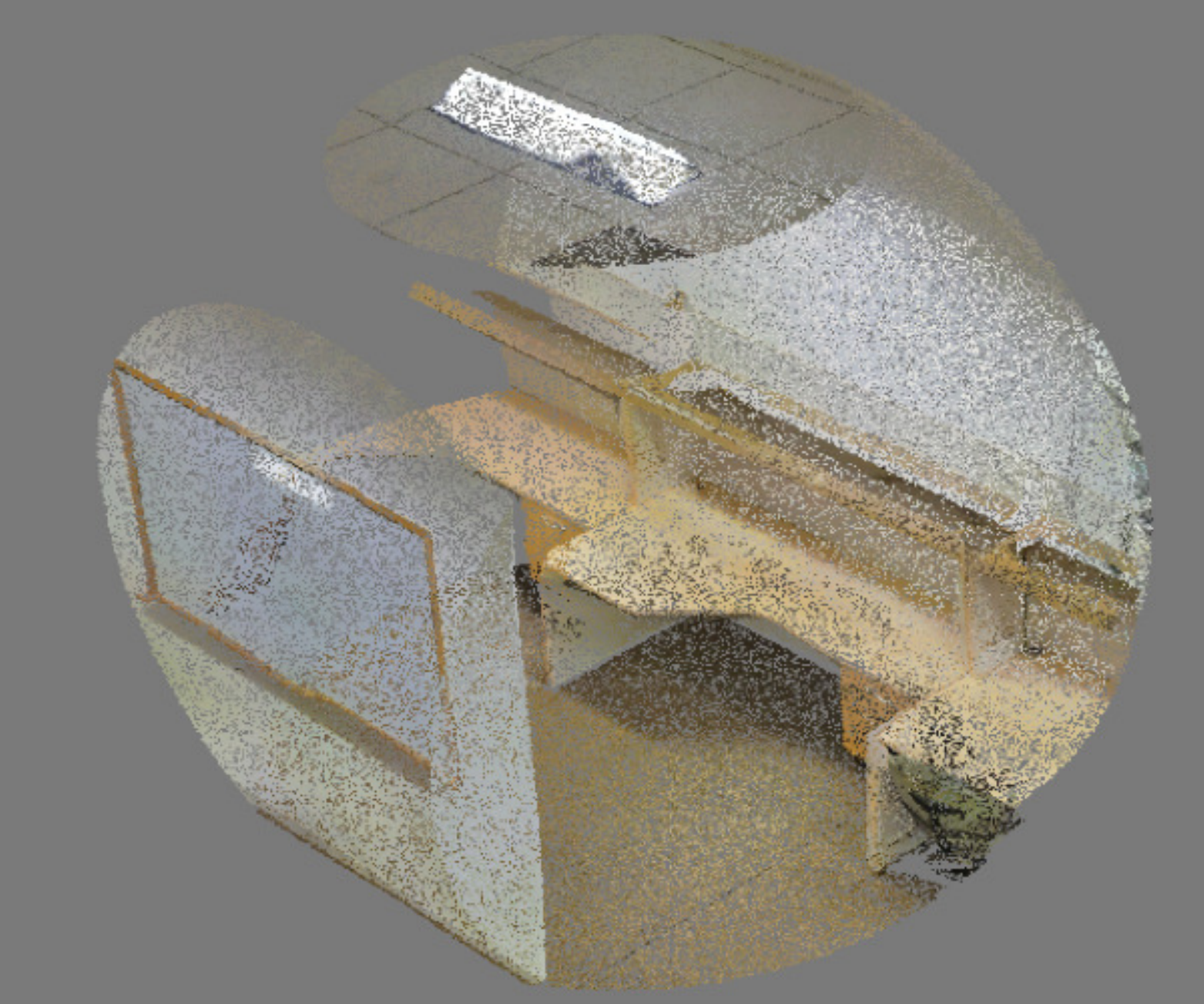}
\end{subfigure} \hfil
\begin{subfigure}{.193\textwidth}
  \centering
  \includegraphics[width=\textwidth, height=0.76\textwidth]{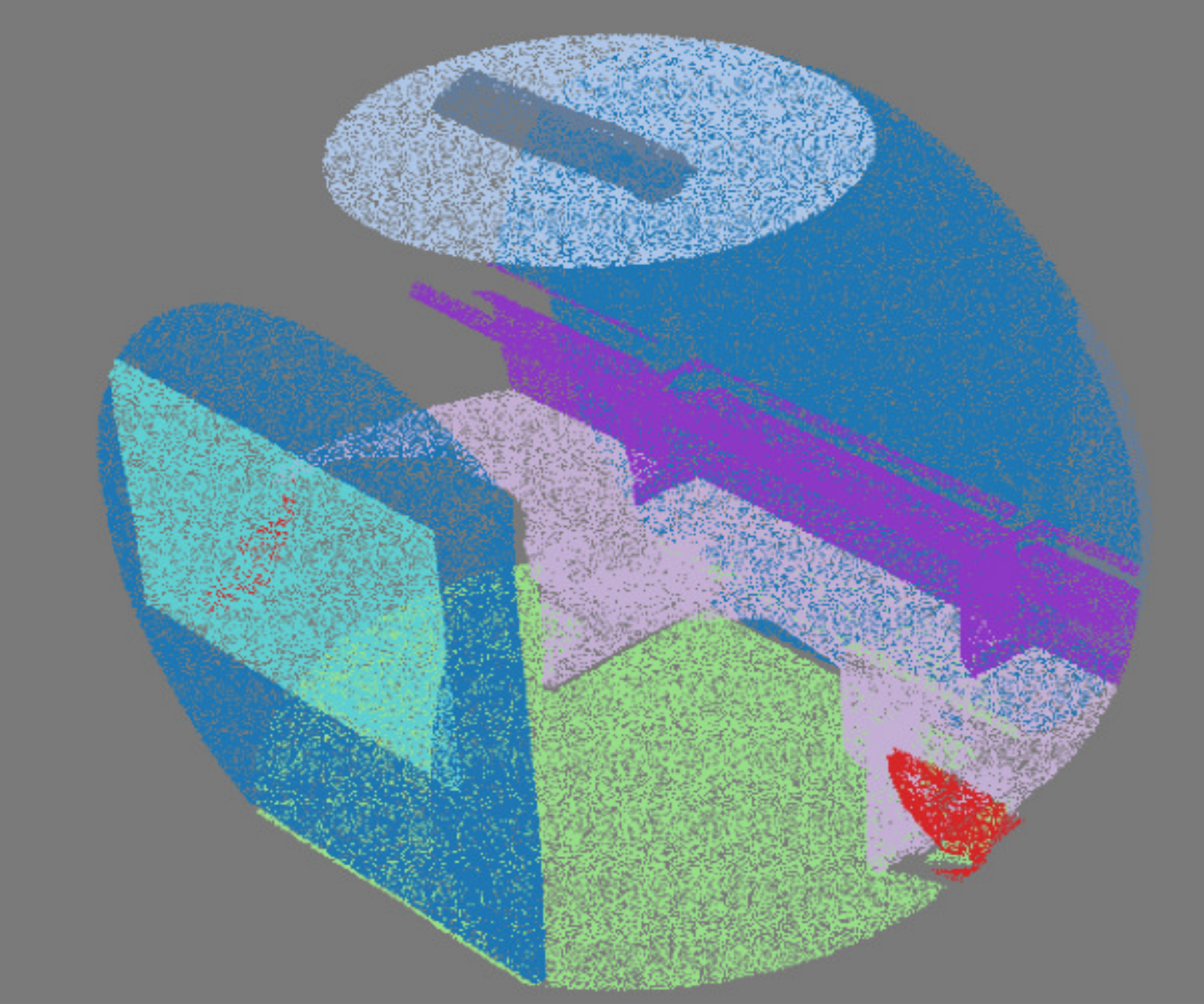}
\end{subfigure} \hfil
\begin{subfigure}{.193\textwidth}
  \centering
  \includegraphics[width=\textwidth, height=0.76\textwidth]{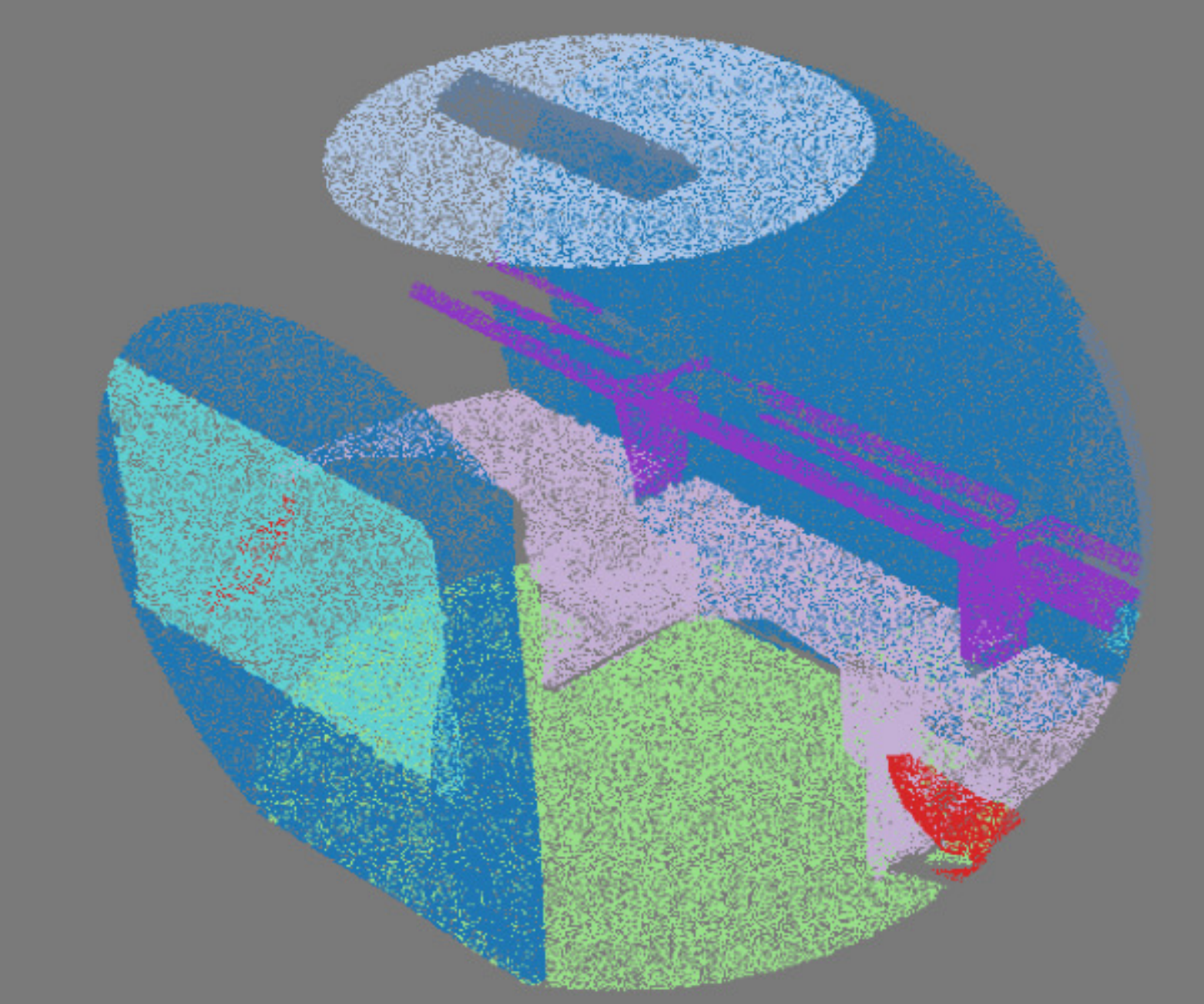}
\end{subfigure} \hfil
\begin{subfigure}{.193\textwidth}
  \centering
  \includegraphics[width=\textwidth, height=0.76\textwidth]{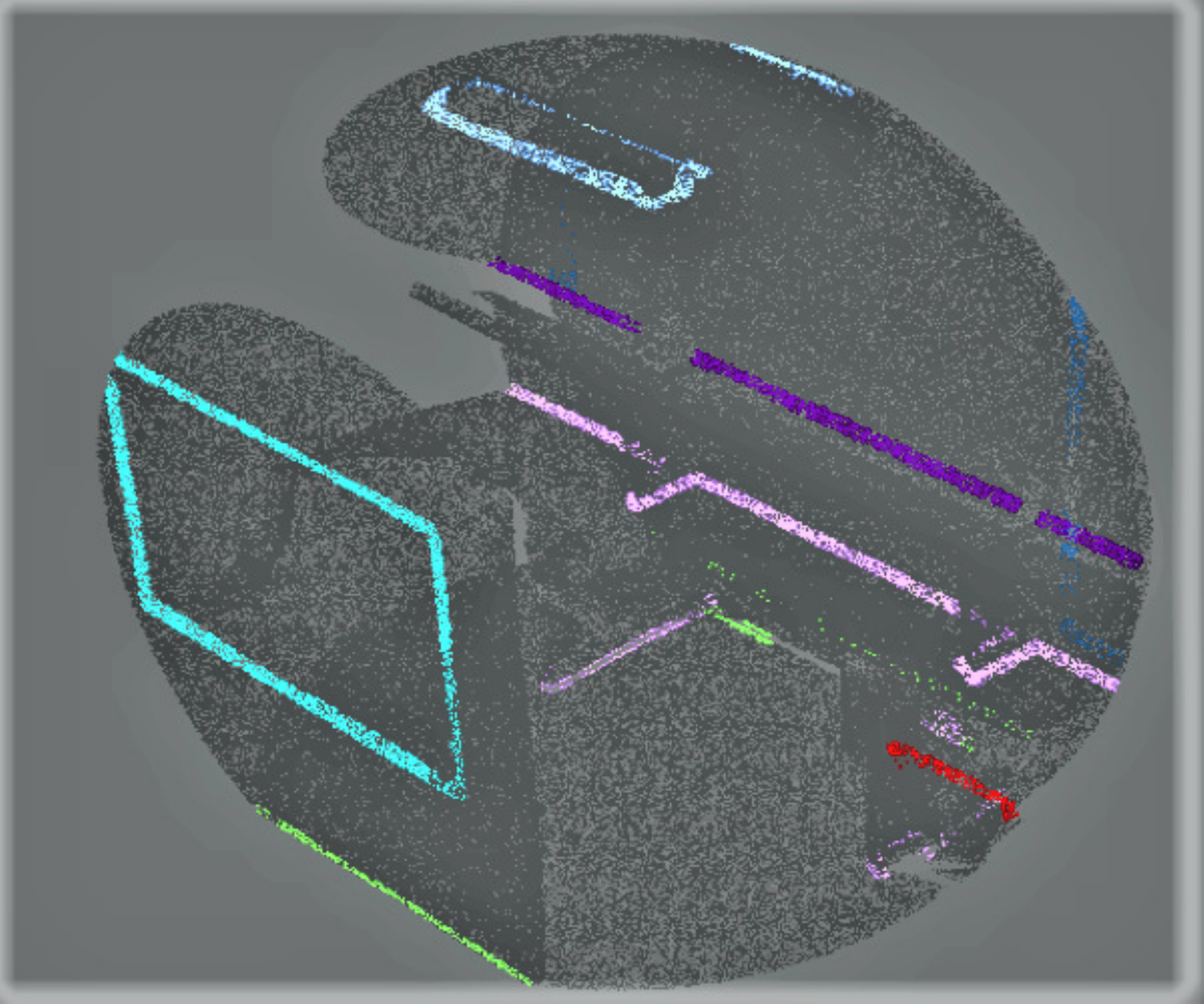}
\end{subfigure} \hfil
\begin{subfigure}{.193\textwidth}
  \centering
  \includegraphics[width=\textwidth, height=0.76\textwidth]{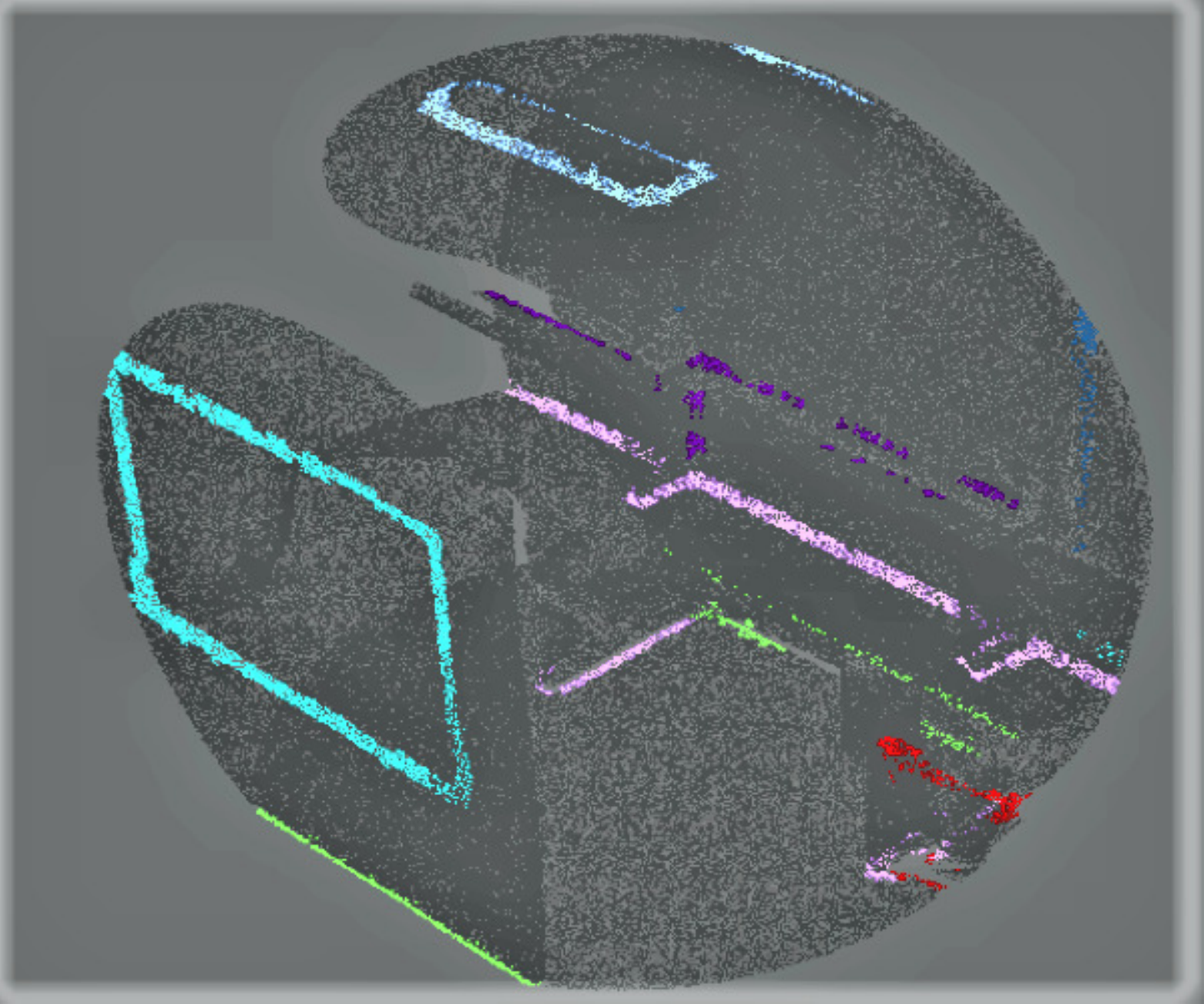}
\end{subfigure}

\begin{subfigure}{.193\textwidth}
  \centering
  \includegraphics[width=\textwidth, height=0.76\textwidth]{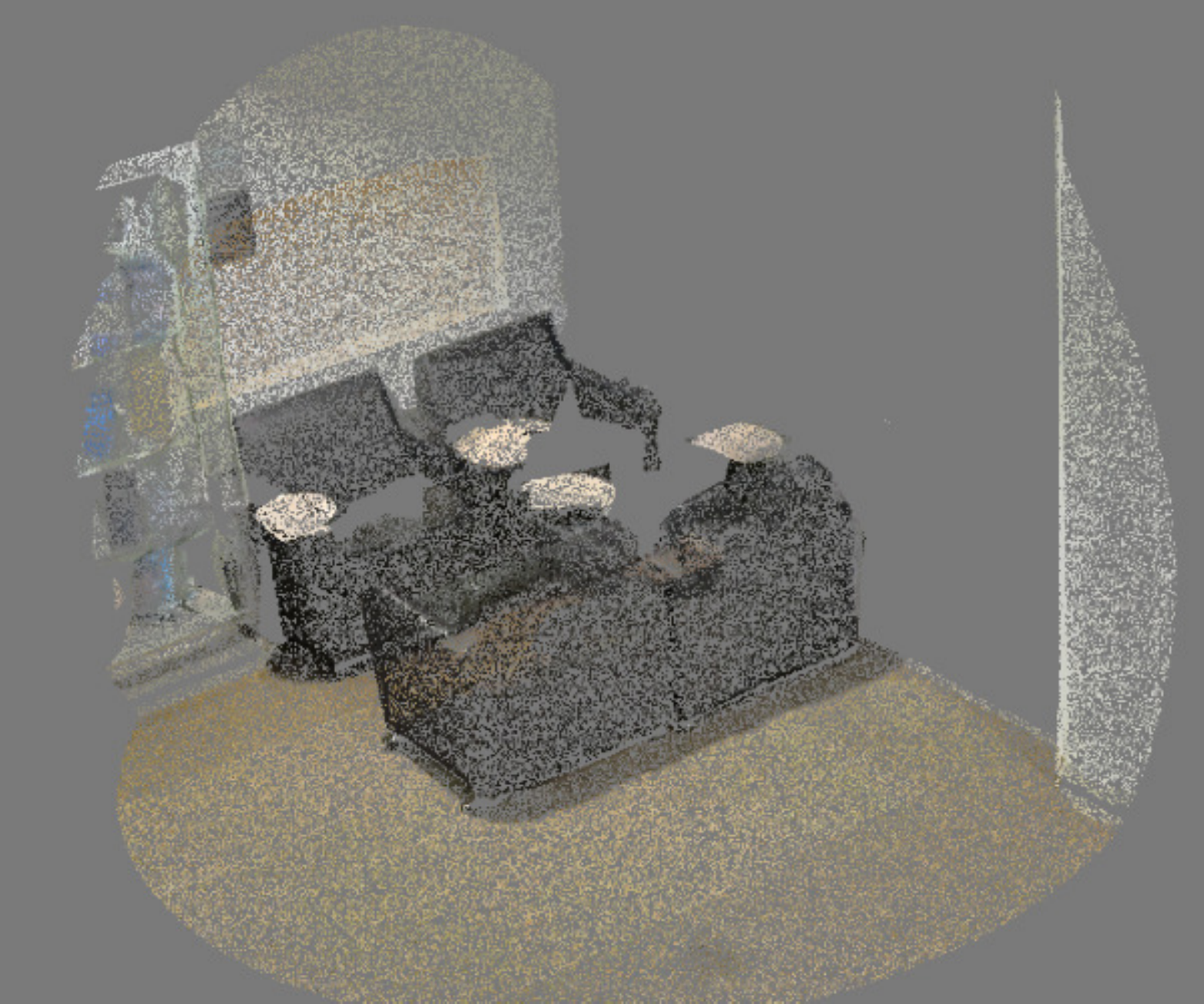}
\end{subfigure} \hfil
\begin{subfigure}{.193\textwidth}
  \centering
  \includegraphics[width=\textwidth, height=0.76\textwidth]{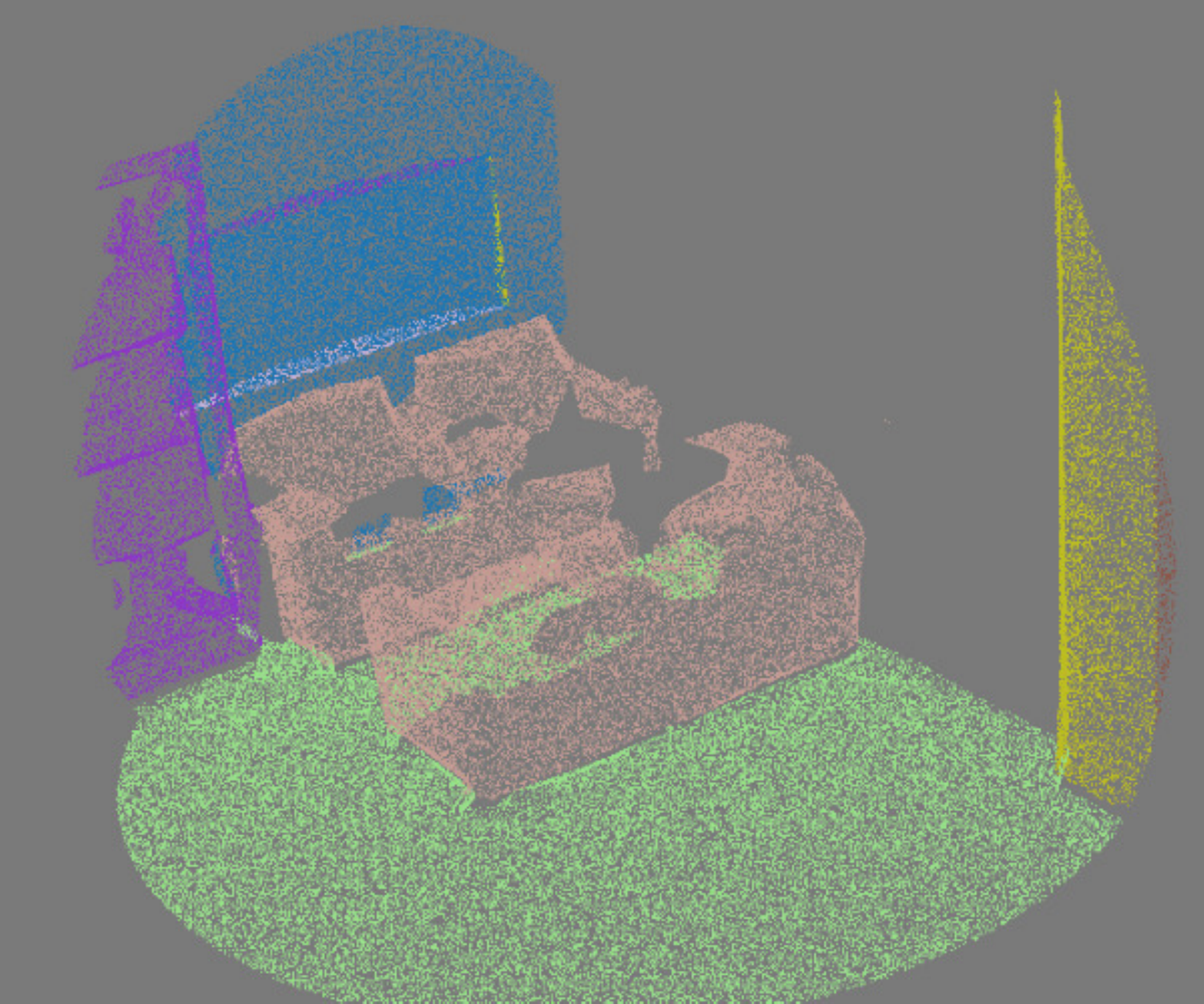}
\end{subfigure} \hfil
\begin{subfigure}{.193\textwidth}
  \centering
  \includegraphics[width=\textwidth, height=0.76\textwidth]{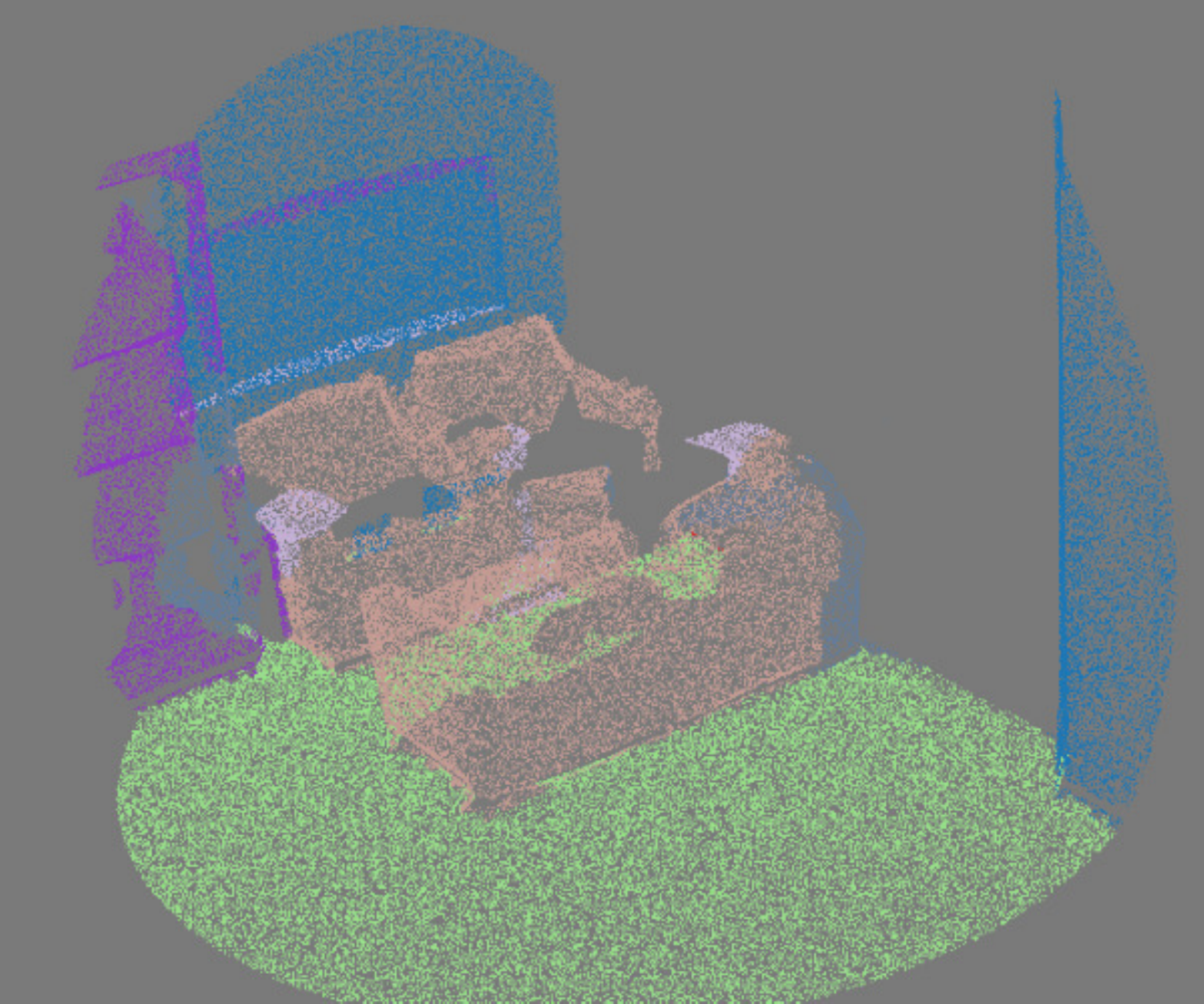}
\end{subfigure} \hfil
\begin{subfigure}{.193\textwidth}
  \centering
  \includegraphics[width=\textwidth, height=0.76\textwidth]{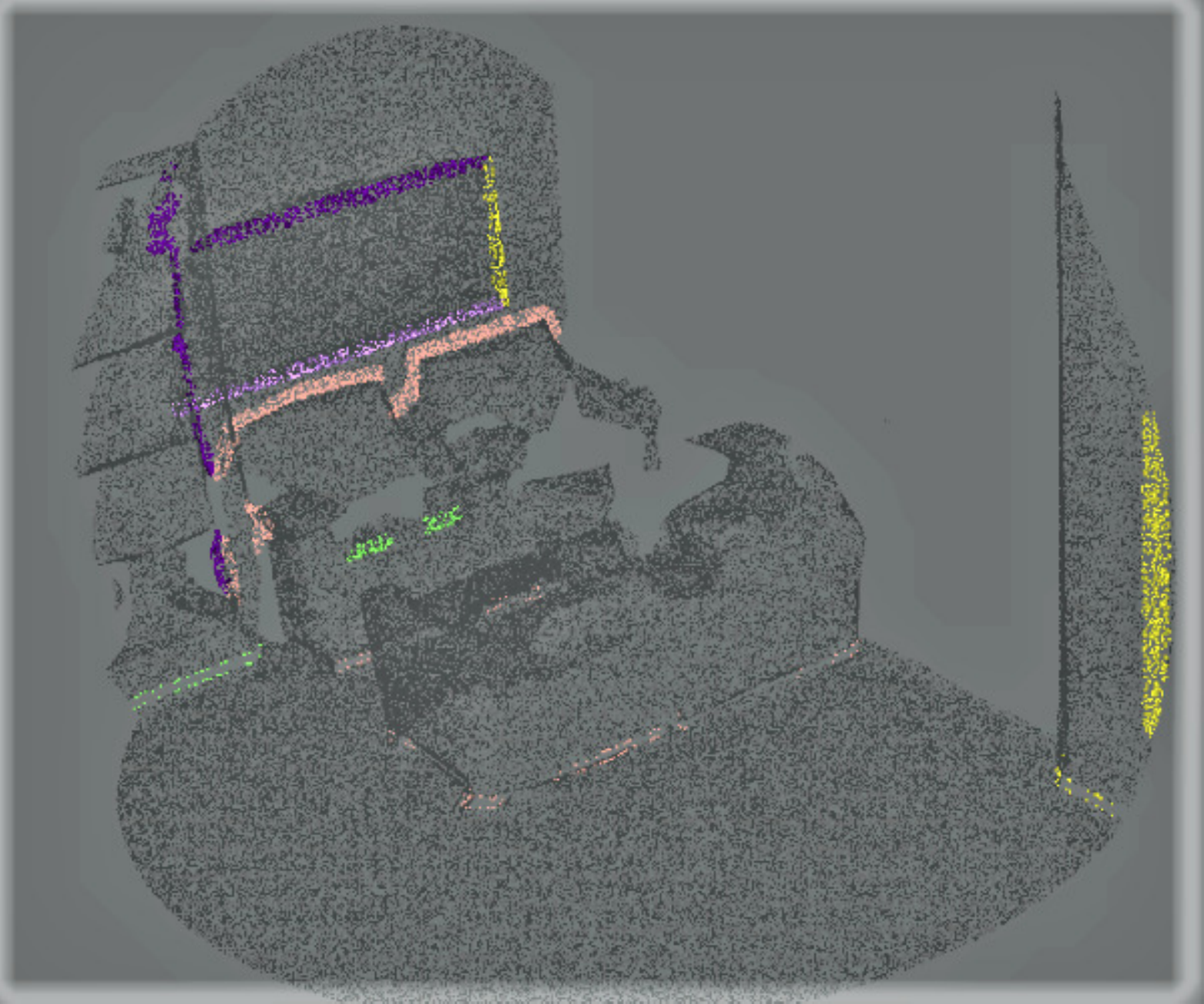}
\end{subfigure} \hfil
\begin{subfigure}{.193\textwidth}
  \centering
  \includegraphics[width=\textwidth, height=0.76\textwidth]{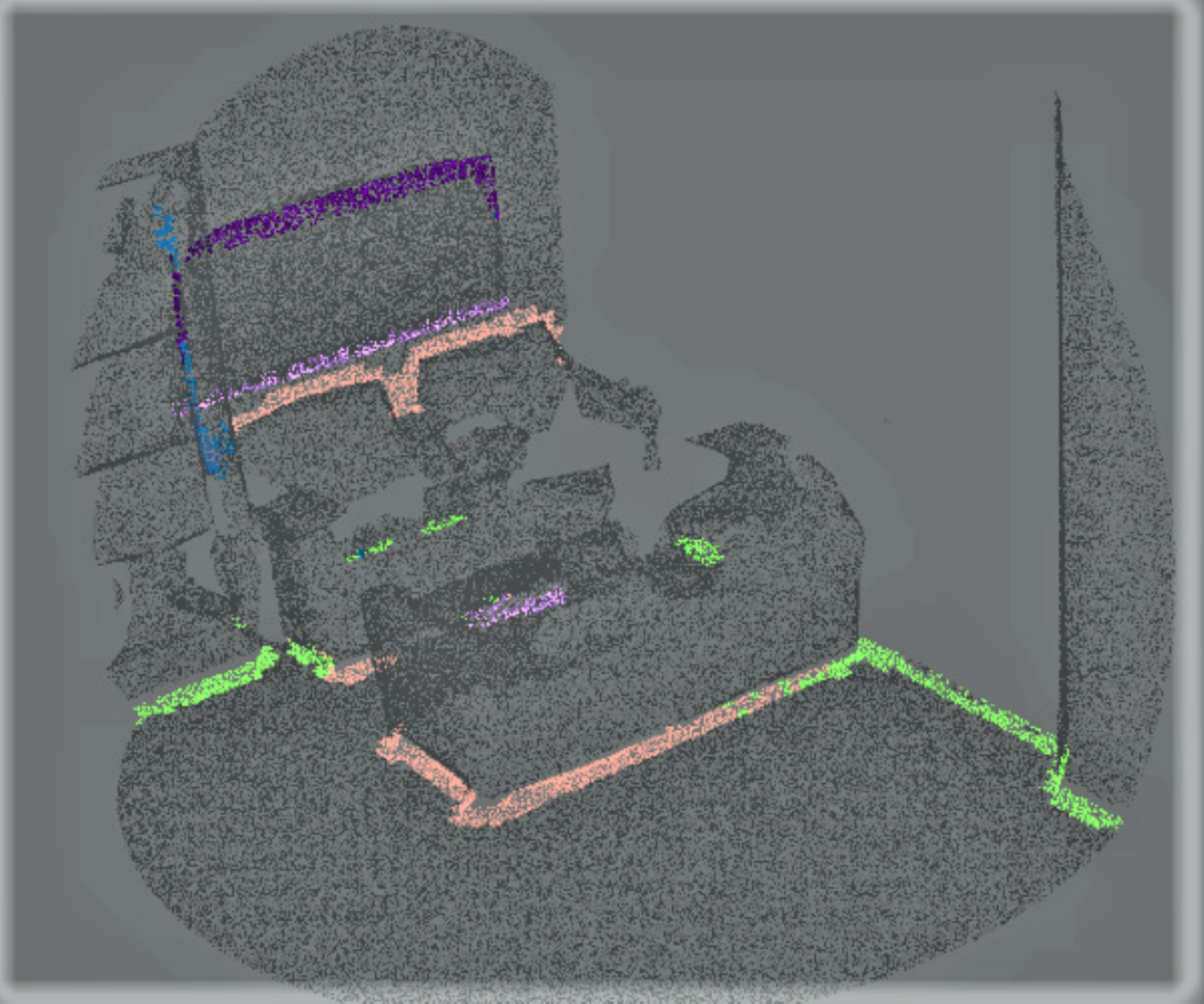}
\end{subfigure}

\begin{subfigure}{.193\textwidth}
  \centering
  \includegraphics[width=\textwidth, height=0.76\textwidth]{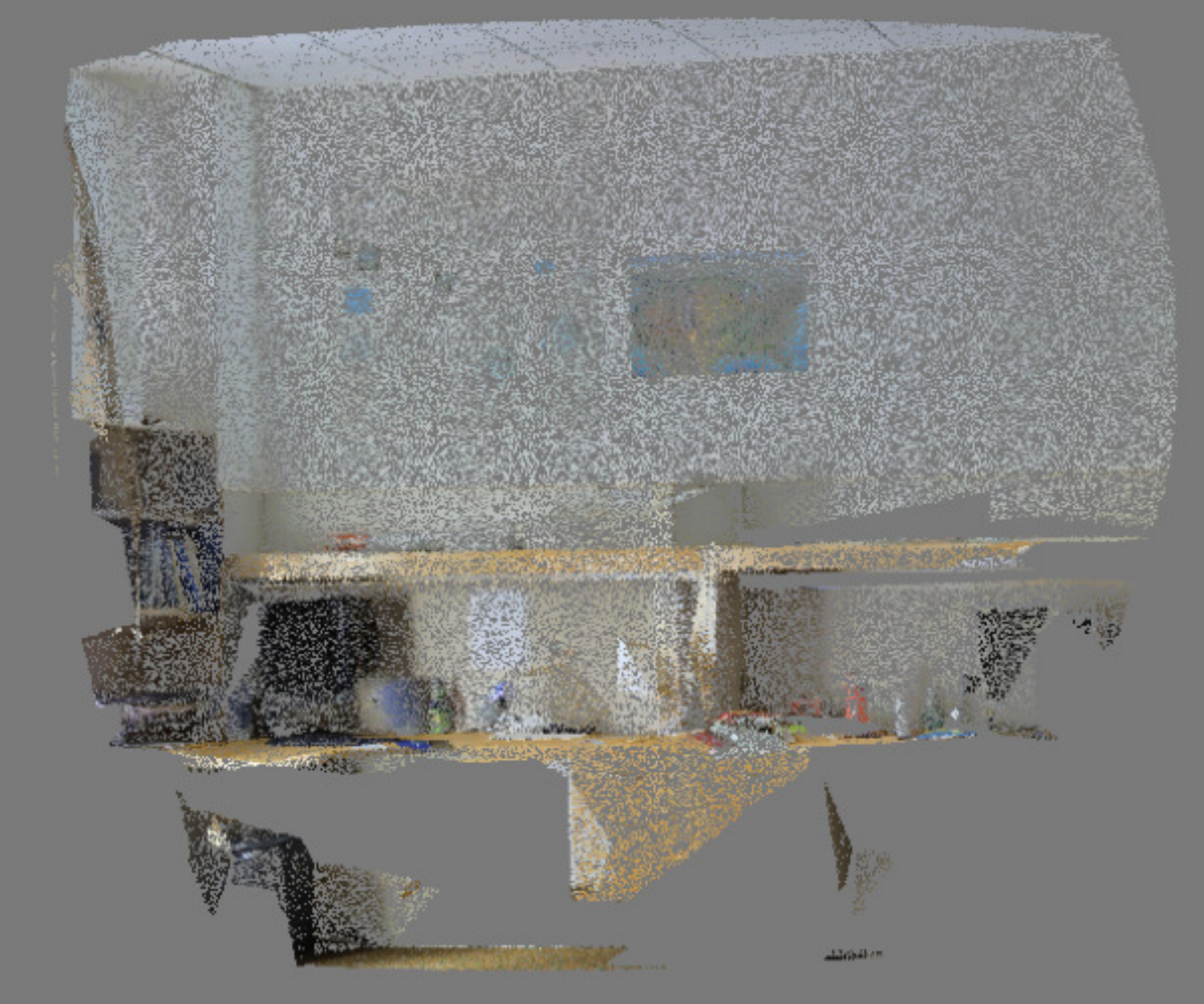}
  \caption{Point Cloud}
\end{subfigure} \hfil
\begin{subfigure}{.193\textwidth}
  \centering
  \includegraphics[width=\textwidth, height=0.76\textwidth]{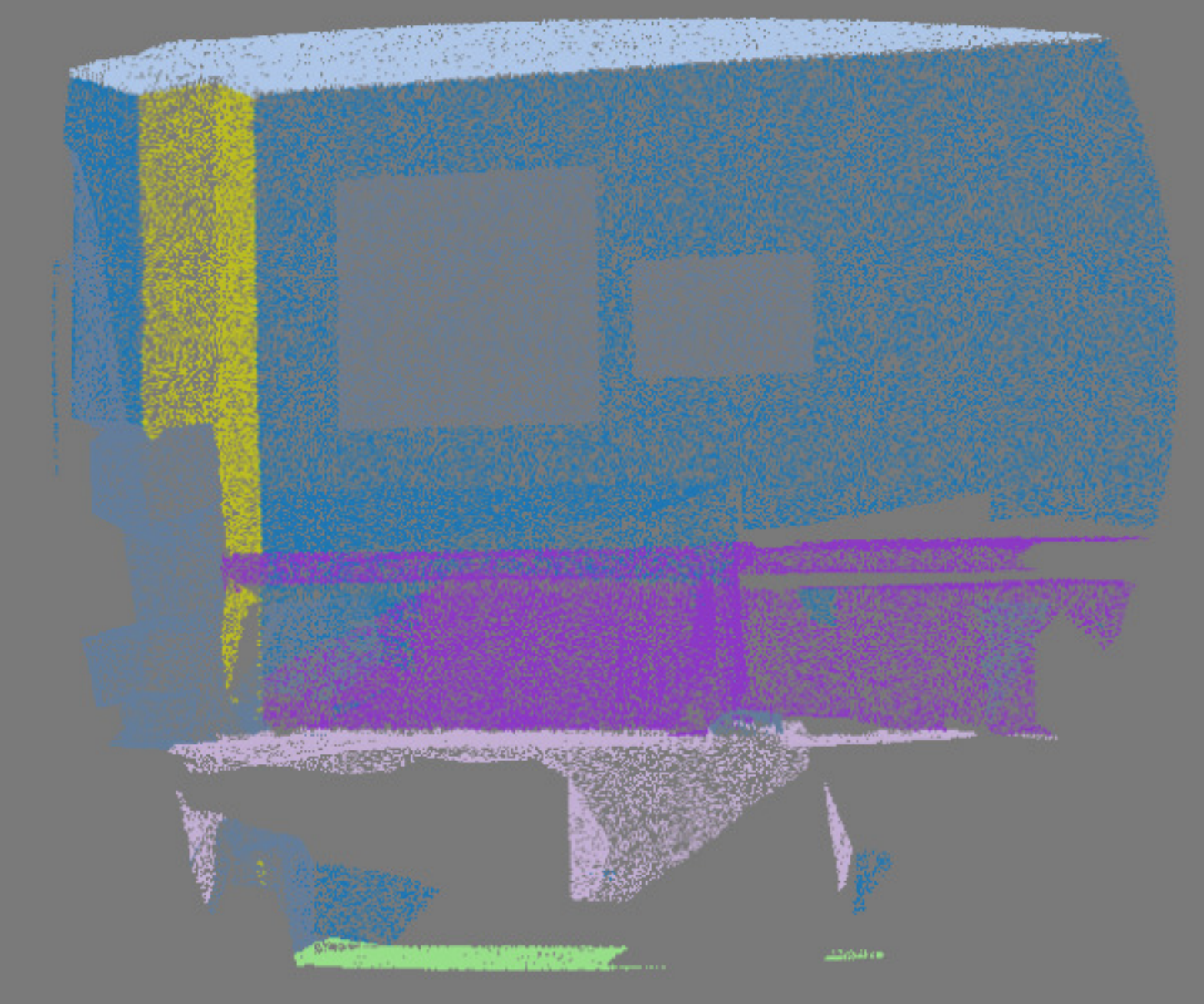}
  \caption{GT-{\SemSegPoint} Mask}
\end{subfigure} \hfil
\begin{subfigure}{.193\textwidth}
  \centering
  \includegraphics[width=\textwidth, height=0.76\textwidth]{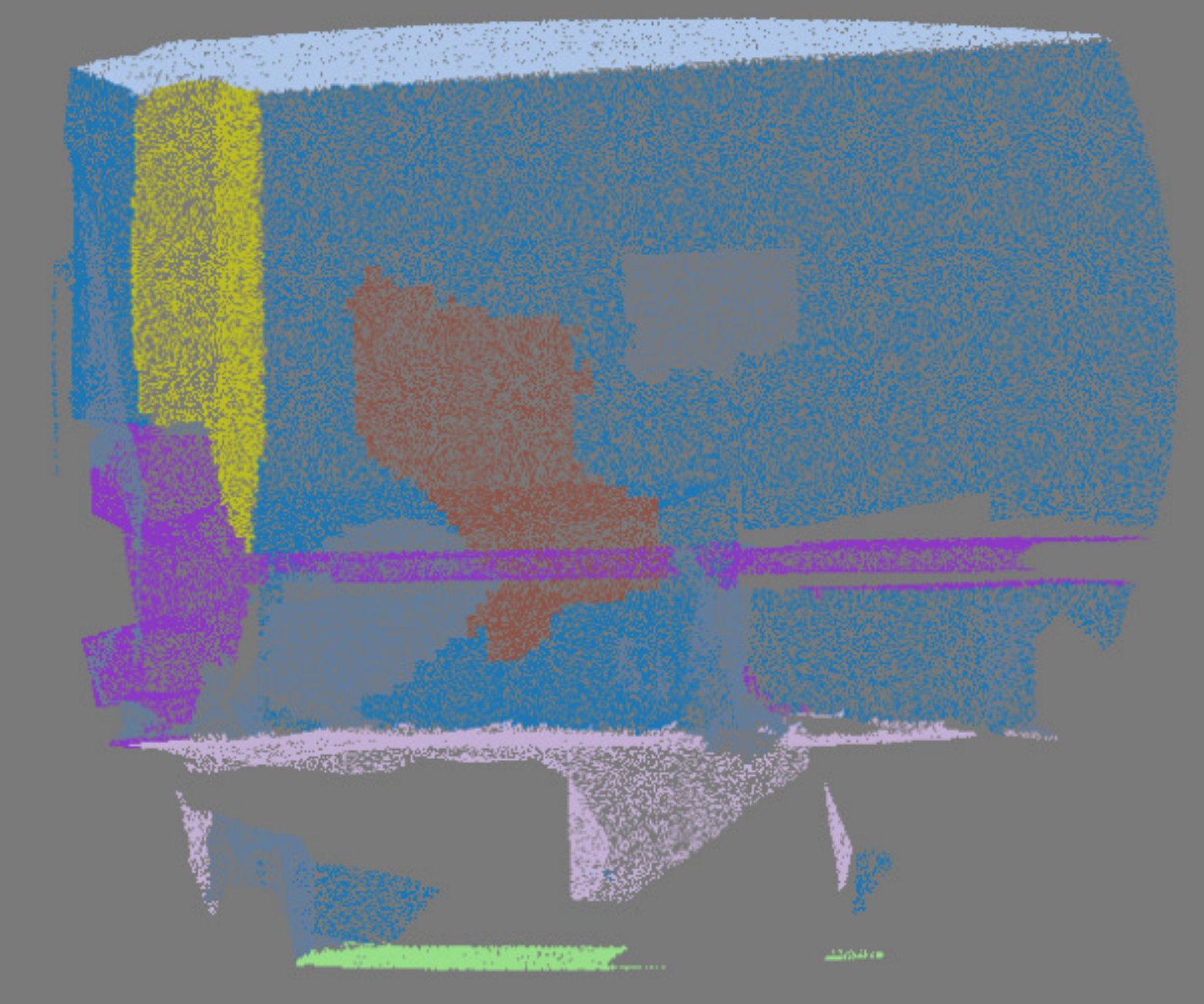}
  \caption{Pred-{\SemSegPoint} Mask}
\end{subfigure} \hfil
\begin{subfigure}{.193\textwidth}
  \centering
  \includegraphics[width=\textwidth, height=0.76\textwidth]{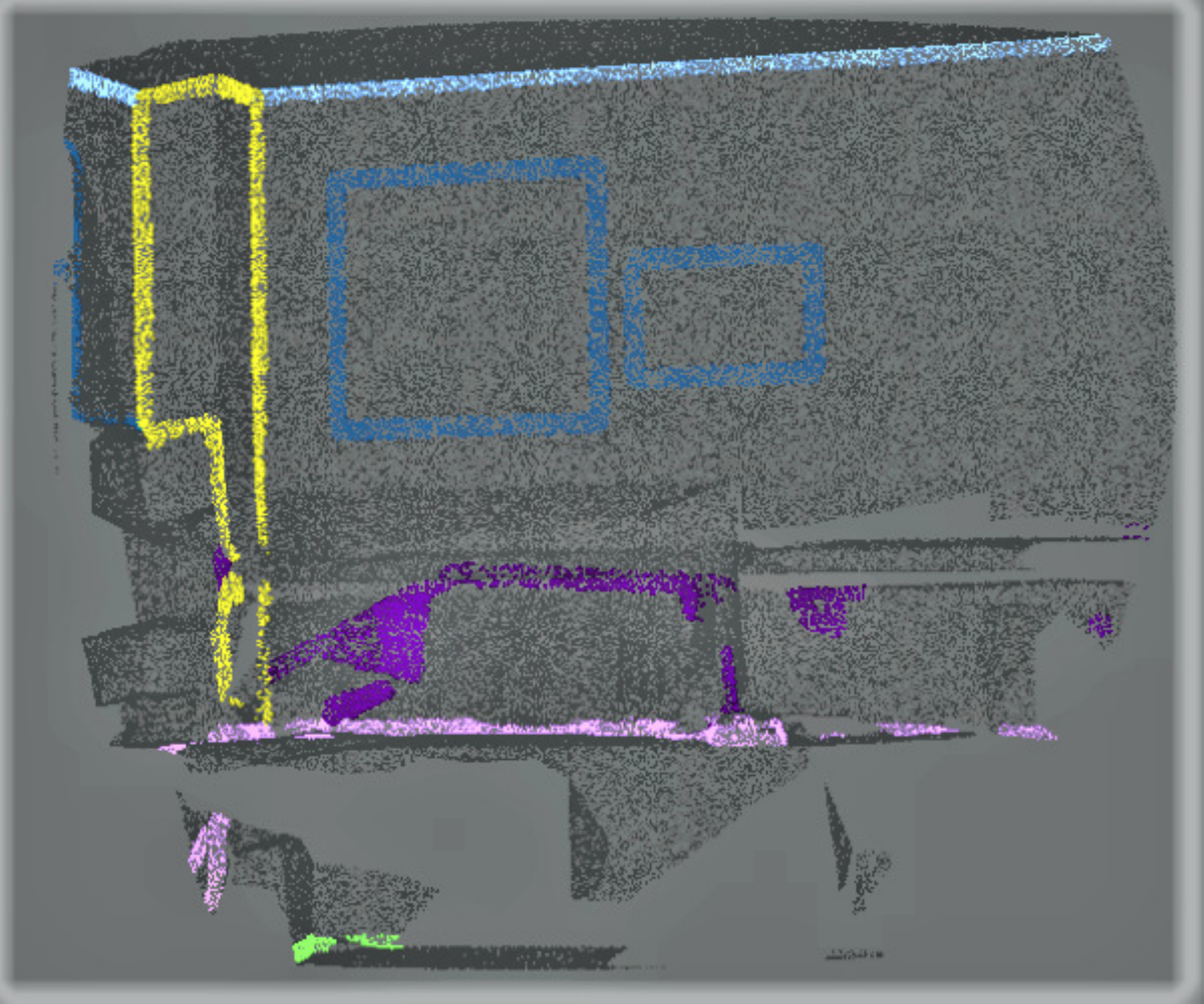}
  \caption{GT-{\SemEdgePoint} Map}
\end{subfigure} \hfil
\begin{subfigure}{.193\textwidth}
  \centering
  \includegraphics[width=\textwidth, height=0.76\textwidth]{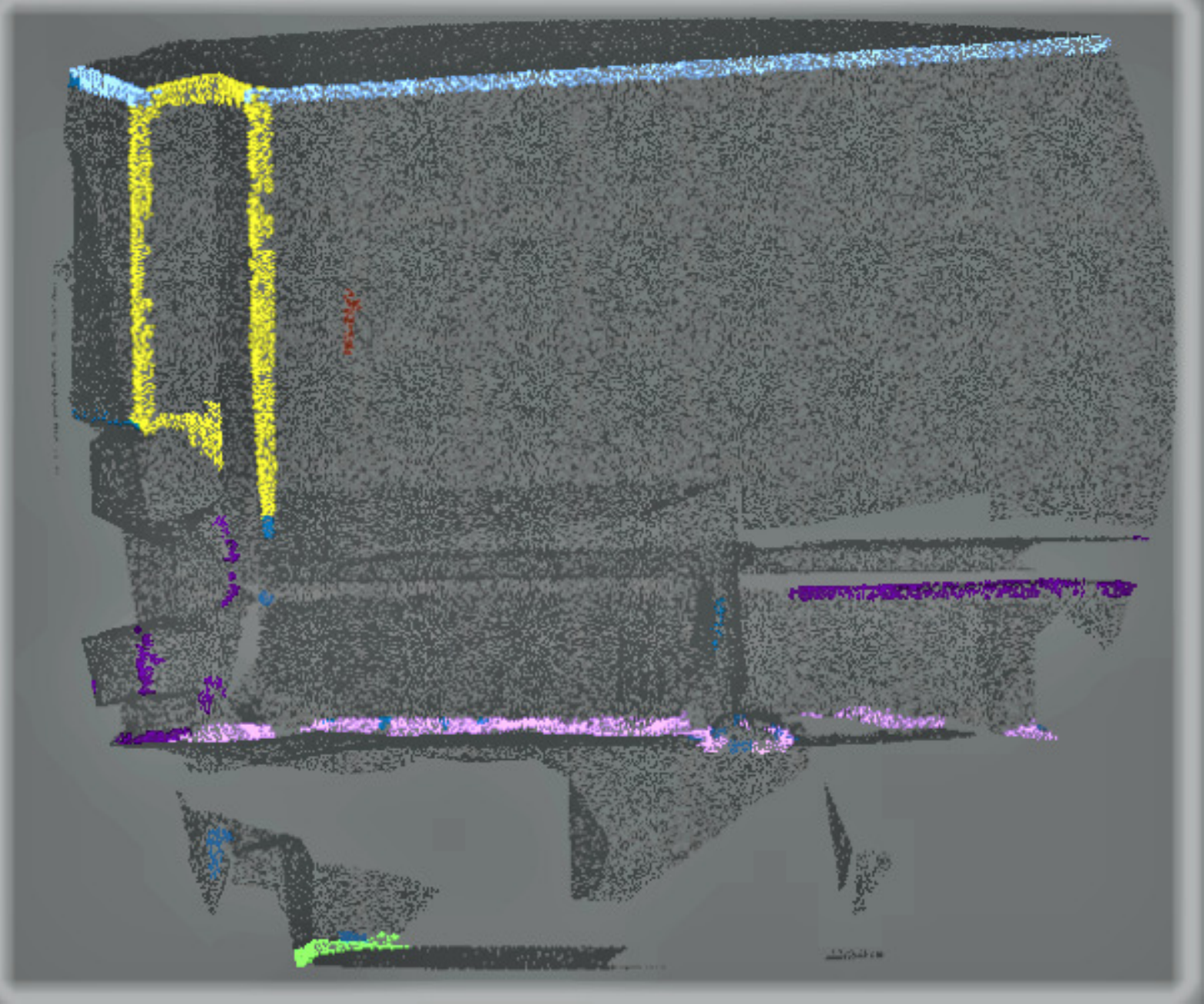}
  \caption{Pred-{\SemEdgePoint} Map}
\end{subfigure}

\caption{Qualitative results on S3DIS Area-5. For better visualization, we thickened all the semantic edges.
}
\label{fig:qualitative}
\end{figure}

\subsection{Results on S3DIS \& ScanNet Datasets} \label{S3DIS}

\begin{table}
\scriptsize
\begin{center}
\caption{{mIoU scores (\%) {of semantic segmentation task.} 
}}
\label{table:ss}
\begin{tabular}{  | m{4cm} | C{2cm}| C{2cm} |}
\hline
Method & S3DIS & ScanNet\\
\hline




TangentConv \cite{tatarchenko2018tangent} & 52.6 & 43.8\\

RNN Fusion \cite{ye20183d} & 53.4 & -\\

SPGraph \cite{landrieu2018large} & 58.0 & -\\

FCPN \cite{rethage2018fully} & - & 44.7\\

PointCNN \cite{li2018pointcnn} & 57.3 & 45.8\\

ParamConv \cite{wang2018deep} & 58.3 & -\\

PanopticFusion \cite{narita2019panopticfusion} & - & 52.9\\

TextureNet \cite{huang2019texturenet} & - & 56.6\\

SPH3D-GCN \cite{lei2019spherical} & 59.5 & 61.0\\

HPEIN \cite{jiang2019hierarchical} & 61.9 & 61.8\\

MCCNN \cite{hermosilla2018monte} & - & 63.3\\

MVPNet \cite{jaritz2019multi} & 62.4 & 64.1\\

PointConv \cite{wu2019pointconv} & - & 66.6\\

KPConv rigid \cite{thomas2019kpconv} & 65.4 & 68.6\\

KPConv deform \cite{thomas2019kpconv} & 67.1 & 68.4\\

SparseConvNet \cite{graham20183d} & - & 72.5\\

MinkowskiNet \cite{Choy_2019} & 65.4 & \textbf{73.6}\\
\hline
JSENet (ours) & \textbf{67.7} & 69.9\\
\hline
\end{tabular}
\end{center}
\end{table}

To compare JSENet with the state-of-the-arts, for {\SemSeg}, we choose the latest methods {\cite{li2018pointcnn,jaritz2019multi,qi2017pointnet,jiang2019hierarchical,Choy_2019,tchapmi2017segcloud,tatarchenko2018tangent,ye20183d,landrieu2018large,rethage2018fully,wang2018deep,narita2019panopticfusion,huang2019texturenet,lei2019spherical,hermosilla2018monte,su2018splatnet,zhang2018efficient,wu2019pointconv,graham20183d}} with reported results on {the} S3DIS or the ScanNet datasets as {our} baselines.
For {\SemEdgeD}, since we cannot find any existing {solutions} in 3D, we extend CASENet \cite{yu2017casenet}, which is the state-of-the-art method {for 2D SED}
to 3D {for comparison.}
Besides, we build another baseline network using the same structure as the one presented in KPConv with a changed output layer.

For the S3DIS dataset, we use the same \emph{train-test} splitting setting as in Section \ref{Ablation study}.
For the ScanNet dataset, since it is built for online benchmarking, the GT semantic labels for its test set are not available. Thus, for {the} {\SemSeg} task, we train our network using the 1513 training scenes and report the testing result{s} on the 100 test scenes following the common setting. To evaluate the task of {\SemEdgeD}, as explained in Section \ref{Datasets}, we obtain semantic edge labels from the training set and divide the 1513 training scenes into a new training set with 1201 scenes and a new test set with 312 scenes following the {\emph{train-val}} splitting file offered by ScanNet. All the compared networks for {\SemEdgeD} are trained on the new training set and tested on the new test set. Qualitative results are shown in Fig. \ref{fig:qualitative}. Complexity comparison and more qualitative results can be found in \textbf{supplementary}.

\smallskip \noindent \textbf{Results of {\SemSeg} task.}
The results for the {\SemSeg} task are reported in Table \ref{table:ss}. The detailed IoU scores of each class for the S3DIS dataset can be found in the supplementary. The details for the ScanNet dataset can be found on the ScanNet benchmarking website\footnote{\url{http://kaldir.vc.in.tum.de/scannet_benchmark/semantic_label_3d}}.
For the S3DIS dataset, JSENet {achieves a $67.7\%$ mIoU score and} outperforms all the baseline methods.
For the ScanNet dataset, JSENet ranks third with a $69.9\%$ mIoU score. The first two (i.e., SparseConvNet and MinkowskiNet) are two voxel-based methods that require intense computation power. {Compared to other point-based methods, JSENet achieves the best performance and consistently outperforms the results of KPConv.}

\begin{table}
\begin{center}
\caption{MF (ODS) scores (\%) {of semantic edge detection} on S3DIS Area-5.}
\label{table:ed_s3dis}
\resizebox{\textwidth}{!}{
    \begin{tabular}{  | l | c | c c c c c c c c c c c c c|}
    \hline
    Method & mean & ceil. & floor & wall & beam & col. & wind. & door & chair & table & book. & sofa & board & clut.\\
    \hline
    CASENet \cite{yu2017casenet} & 27.1 & \textbf{46.5} & \textbf{49.0} & 33.3 & 0.2 & 21.9 & 12.6 & 22.6 & 36.9 & 33.6 & 21.8 & 25.1 & 22.6 & 26.1\\
    \hline
    KPConv \cite{thomas2019kpconv} & 29.4 & 43.7 & 41.8 & 36.4 & 0.2 & 23.6 & \textbf{13.4} & 29.7 & \textbf{39.8} & \textbf{37.3} & 26.6 & 29.4 & 29.2 & 31.3\\
    \hline
    JSENet (ours) & \textbf{31.0} & 44.5 & 43.2 & \textbf{38.8} & \textbf{0.2} & \textbf{24.1} & 13.2 & \textbf{36.7} & 37.7 & 36.3 & \textbf{29.1} & \textbf{34.0} & \textbf{33.3} & \textbf{32.4}\\
    \hline
    \end{tabular}}
\end{center}
\end{table}

\smallskip \noindent \textbf{Results of {\SemEdgeD} task.}
We present the results for the {\SemEdgeD} task in Tables \ref{table:ed_s3dis} and \ref{table:ed_scannet}. It can be seen that our method outperforms the two baseline
methods on both datasets. We find that the extended CASENet architecture performs worse than KPConv with the changed output layer. This result supports our previous argument (Section \ref{2D SED}) that 3D semantic edges have stronger relevance between different classes (since they are physical boundaries of objects and the 2D semantic edges are occlusion boundaries of projected objects), {and} thus {the} network designs that enforce limitations between interactions of different classes would harm the edge detection performance.

\begin{table}
\begin{center}
\caption{MF (ODS) scores (\%) of semantic edge detection on ScanNet val set.}
\label{table:ed_scannet}
\resizebox{\textwidth}{!}{
    \begin{tabular}{  | l | c | c c c c c c c c c c c c c c c c c c c c|}
    \hline
    Method & mean & bath & bed & bksf & cab & chair & cntr & curt & desk & door & floor & othr & pic & ref & show & sink & sofa & tab & toil & wall & wind\\
    \hline
    CASENet \cite{yu2017casenet} & 32.3 & 38.2 & 55.9 & 29.9 & 36.0 & 36.0 & 36.8 & 28.1 & 28.5 & 19.1 & 26.6 & 25.1 & 32.2 & 28.6 & 26.6 & 23.9 & 19.0 & 45.7 & 27.1 & 54.6 & 27.4\\
    \hline
    KPConv \cite{thomas2019kpconv} & 34.8 & 40.5 & \textbf{55.9} & 33.6 & \textbf{38.5} & 39.3 & 38.0 & 32.9 & 31.2 & 22.0 & 25.1 & 28.5 & 36.2 & 30.8 & 30.9 & 22.7 & 22.8 & 46.5 & 33.3 & \textbf{55.2} & 32.3\\
    \hline
    JSENet (ours) & \textbf{37.3} & \textbf{43.8} & 55.8 & \textbf{35.9} & 38.2 & \textbf{41.0} & \textbf{40.8} & \textbf{34.5} & \textbf{35.9} & \textbf{25.5} & \textbf{28.7} & \textbf{29.5} & \textbf{37.3} & \textbf{36.2} & \textbf{31.7} & \textbf{28.1} & \textbf{28.3} & \textbf{48.5} & \textbf{35.6} & 53.2 & \textbf{37.8} \\
    \hline
    \end{tabular}}
\end{center}
\end{table}

%% file: conclusion.tex
In this paper, we propose{d} a joint semantic segmentation and semantic edge detection network (JSENet), which is a new two-stream fully-convolutional architecture with a lightweight joint refinement module that explicitly wires region information and edge information for output refinement. 
{We constructed a two-branch structure with simple feature fusion sub-modules and novel edge map generation sub-modules for the joint refinement module and designed the dual semantic edge loss that encourages the network to produce sharper predictions around object boundaries.}
Our experiments show that JSENet is an effective architecture that produces {\SemSegPoint} masks and {\SemEdgePoint} maps of high qualities. 
JSENet achieves the state-of-the-art results on the challenging S3DIS and ScanNet dataset{s}, significantly improving over strong baselines. For future works, one straightforward direction is to explore the potential of joint improvement of these two tasks in the 2D field. Moreover, we notice that our  {\SemEdgeD} method is easily affected by the noises in the GT labels introduced by human annotation. We believe that future works with special treatments on the misaligned GT semantic edges will further improve the performance of the {\SemEdgeD} task.

\noindent\textbf{Acknowledgements.} This work is supported by Hong Kong RGC GRF 16206819, 16203518 and Centre for Applied Computing and Interactive Media (ACIM) of School of Creative Media, City University of Hong Kong.


%% file: supplementary.tex
\pdfoutput=1
\renewcommand\thesection{\Alph{section}}

\pagestyle{headings}
\mainmatter
\def\ECCVSubNumber{3411}  

\title{\emph{Supplementary Material} for \\
JSENet: Joint Semantic Segmentation and Edge Detection Network for 3D Point Clouds} 

\titlerunning{JSENet Supplementary} 
\authorrunning{Z. HU et al.} 
\author{}
\institute{}

\maketitle

\begin{abstract}
This supplementary document is organized as follows:

\begin{itemize}
  \item Section \ref{dataset} explains in more detail about the dataset selection and preparation.
  \item Section \ref{complexity} compares the model sizes and speeds of our network with others.
  \item Section \ref{qualitative} provides some qualitative comparison examples and more visualization results on the ScanNet \cite{dai2017scannet} dataset.
  \item Section \ref{class_results} enumerates detailed semantic segmentation results with class scores.
\end{itemize}
\end{abstract}

\section{Dataset selection and preparation.}\label{dataset}
In the main paper, we conduct all the experiments on two indoor-scene datasets: S3DIS \cite{armeni_cvpr16} and ScanNet \cite{dai2017scannet}. The main reason for choosing no outdoor-scene dataset is that we find semantic edges are not well defined in existing outdoor-scene datasets. As shown in Fig. 1, compared to the indoor-scene datasets, existing outdoor-scene datasets suffer more from incompletion. Objects in an outdoor-scene are often not densely connected due to the missing parts in the point cloud. Therefore, it is hard to define meaningful semantic edges on these point clouds for our evaluation.

\begin{figure}[htp]

\centering
\includegraphics[width=.4\textwidth]{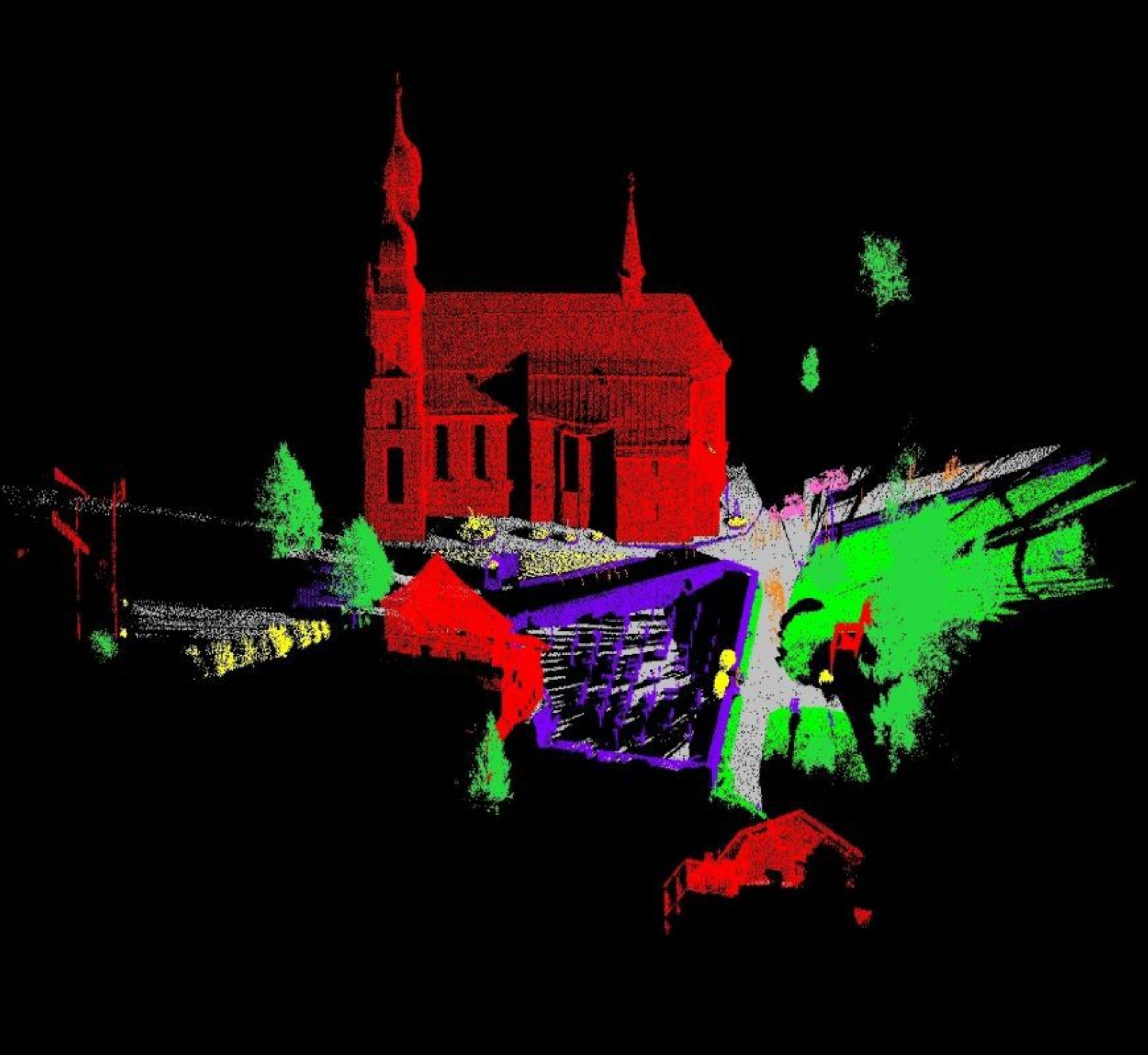}\hfil
\includegraphics[width=.4\textwidth]{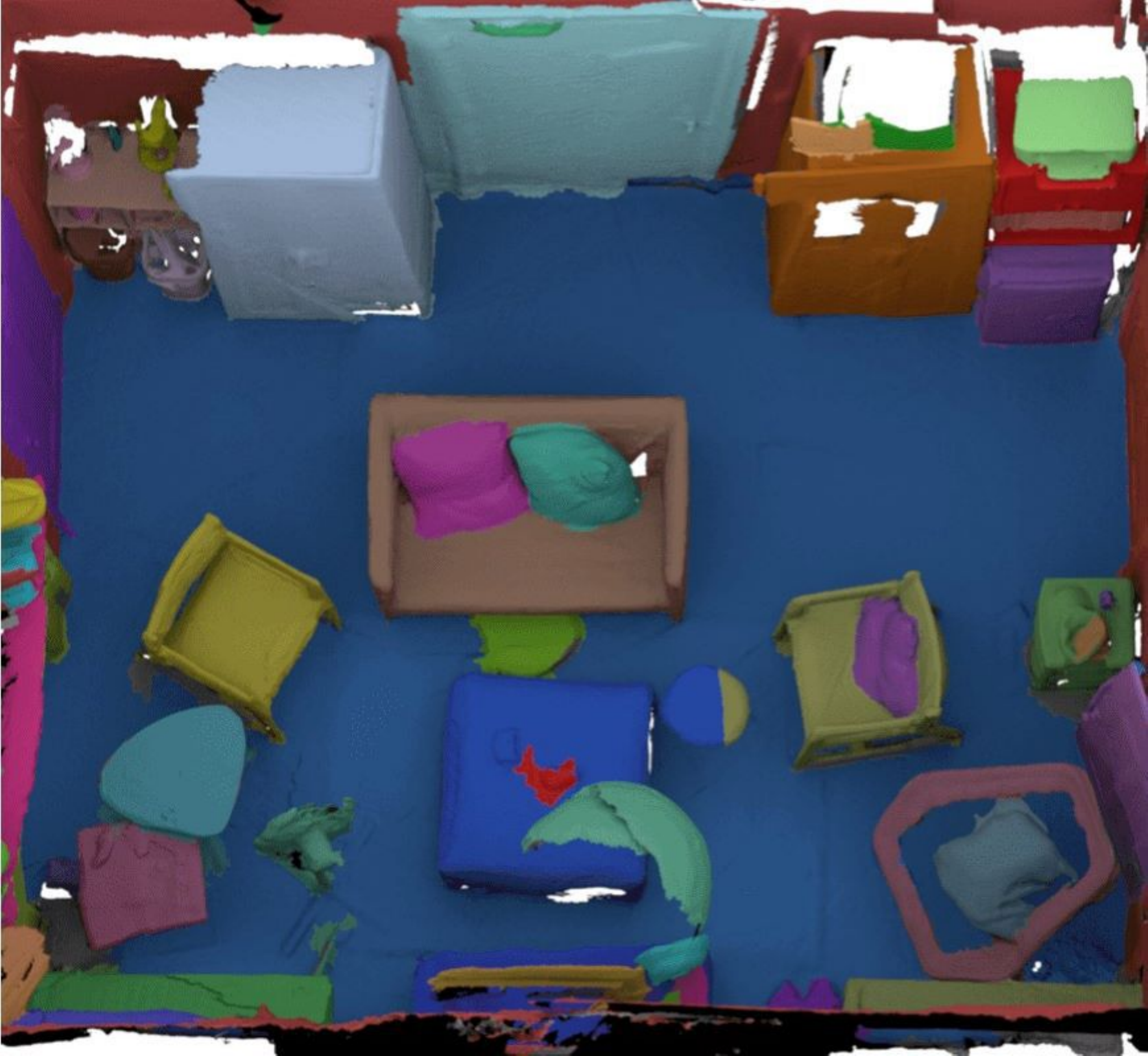}

\caption{(Left) An outdoor scene from Semantic3D \cite{hackel2017isprs}; (Right) An indoor scene from ScanNet \cite{dai2017scannet} }

\end{figure}

We generate 3D semantic edges following the idea from 2D works \cite{yu2017casenet,prasad2006learning} with slight differences. In 2D, thin semantic edges of one or two pixels width are generated. In contrast, we generate thick semantic edges in 3D since points in a point cloud are much sparser than pixels in an image. Moreover, boundaries between an object and the background are considered as semantic edges in 2D images. However, these boundaries are meaningless in the 3D case. Thus, in 3D, we only consider semantic edges between different objects. In general, all semantic edge points will have two or more than two class labels. Since there are unconsidered classes in the ScanNet dataset, semantic edges between a considered class and an unconsidered class might have only one class label.

\section{Complexity of the network, in comparison with other works.}\label{complexity}

In this section, we present the comparison on the complexity of our network against state-of-the-art methods. All the experiments have been conducted on a PC with 8 Intel(R) i7-7700 CPUs and a single GeForce GTX 1080Ti GPU.

\subsubsection{Training.} We train KPConv and JSENet on the ScanNet dataset. Using the setting presented in their paper, KPConv takes about 0.7s for one training iteration and converges in 160K iterations, taking about 31h in total. Using the setting presented in our paper, in the first step, JSENet takes about 0.9s for one training iteration and converges in 170K iterations. In the second step, JSENet takes about 0.6s for one training iteration and converges in 40K iterations. The whole training takes about 49h.

\begin{table}
\small
\begin{center}
\caption{Comparison on runtime complexity of JSENet against state-of-the-art methods.
}
\label{table:complexity}
\resizebox{0.7\textwidth}{!}{
    \begin{tabular}{  | l | c | c |}
    \hline
    Method & Average time (s) & Parameters (m)\\
    \hline
    KPConv \cite{thomas2019kpconv} & 0.044 & 14.1 \\
    \hline
    PointConv \cite{wu2019pointconv} & 0.307 & 21.7\\
    \hline
    MinkowskiNet \cite{Choy_2019} & 0.185 & 37.9\\
    \hline
    JSENet & 0.097 & 16.2\\
    \hline
    \end{tabular}}
\end{center}
\end{table}
\vspace{-5mm}

\subsubsection{Inference.}

We compare KPConv, PointConv, MinkowskiNet, and JSENet for their runtime complexity given the same sets of points extracted from the ScanNet dataset (13000 points each). Results are shown in Table. \ref{table:complexity}. It can be seen that for both the inference time and the parameter size, JSENet is largely comparable to KPConv and is both more efficient and compact than PointConv (another recent point-based method) and MinkowskiNet (the SOTA voxel-based method).

\section{Qualitative Visualization.}\label{qualitative}
\begin{figure}
\captionsetup[subfigure]{labelformat=empty}
\begin{subfigure}{.193\textwidth}
  \centering
  \includegraphics[width=\textwidth]{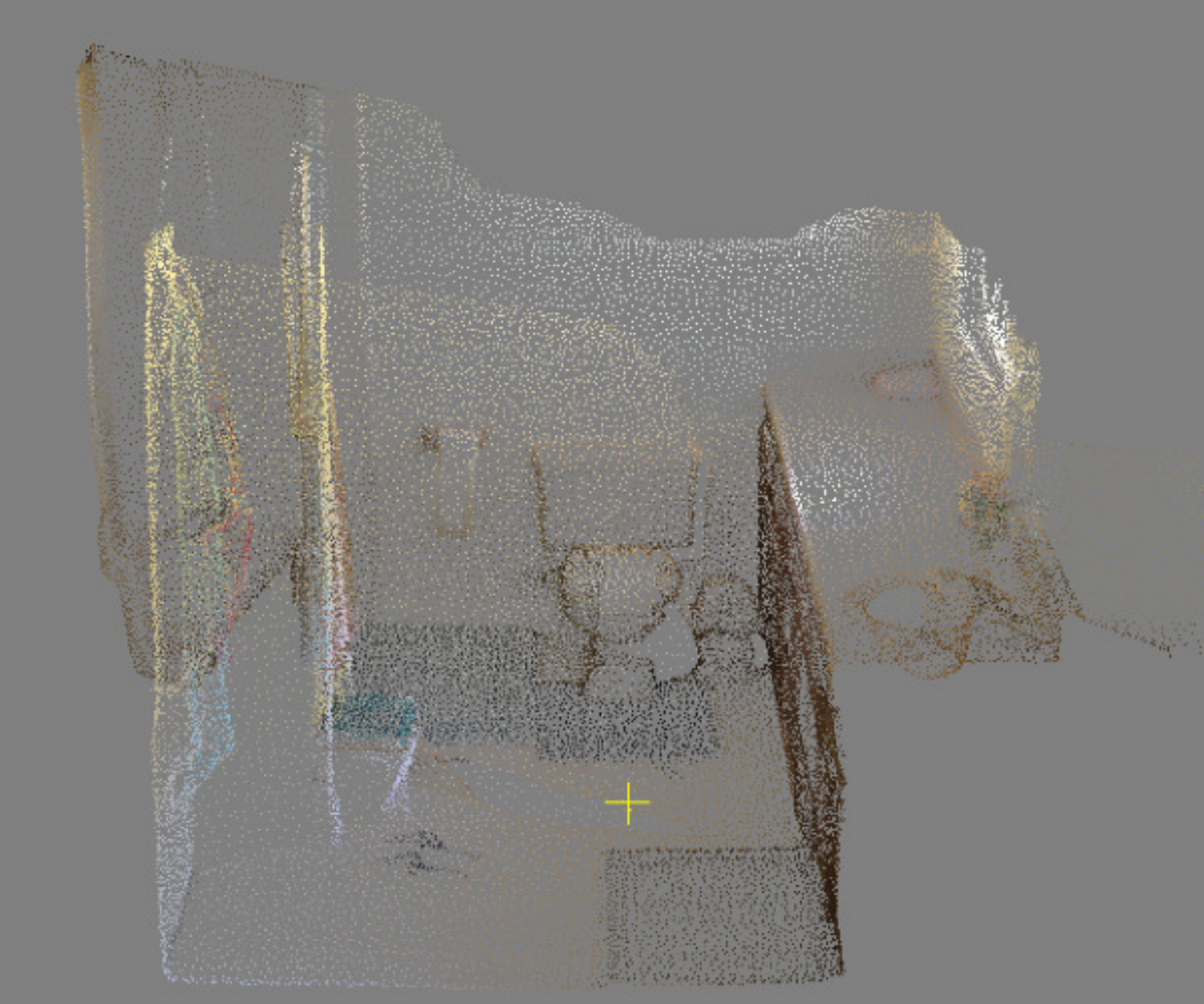}
\end{subfigure} \hfil
\begin{subfigure}{.193\textwidth}
  \centering
  \includegraphics[width=\textwidth]{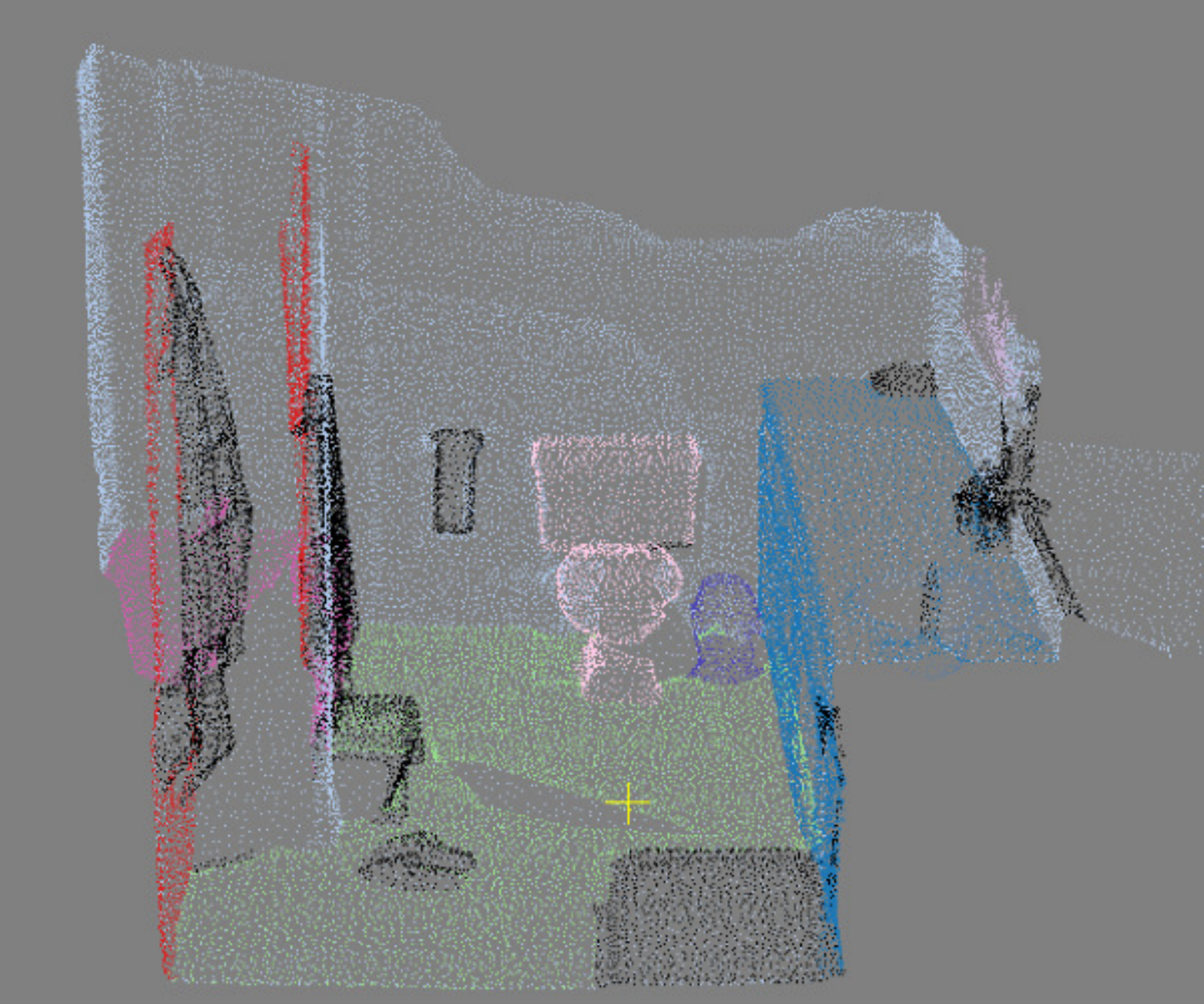}
\end{subfigure} \hfil
\begin{subfigure}{.193\textwidth}
  \centering
  \includegraphics[width=\textwidth]{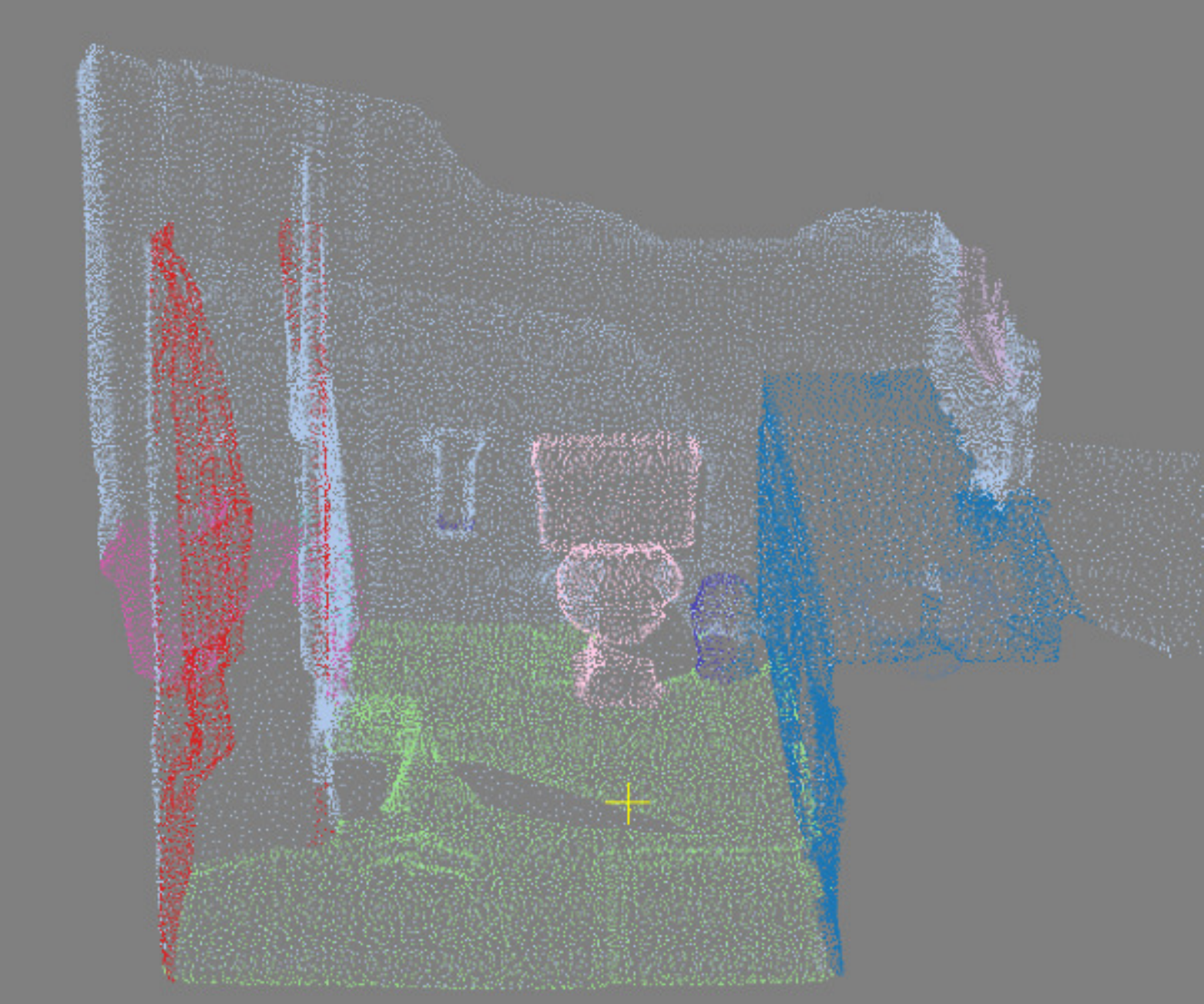}
\end{subfigure} \hfil
\begin{subfigure}{.193\textwidth}
  \centering
  \includegraphics[width=\textwidth]{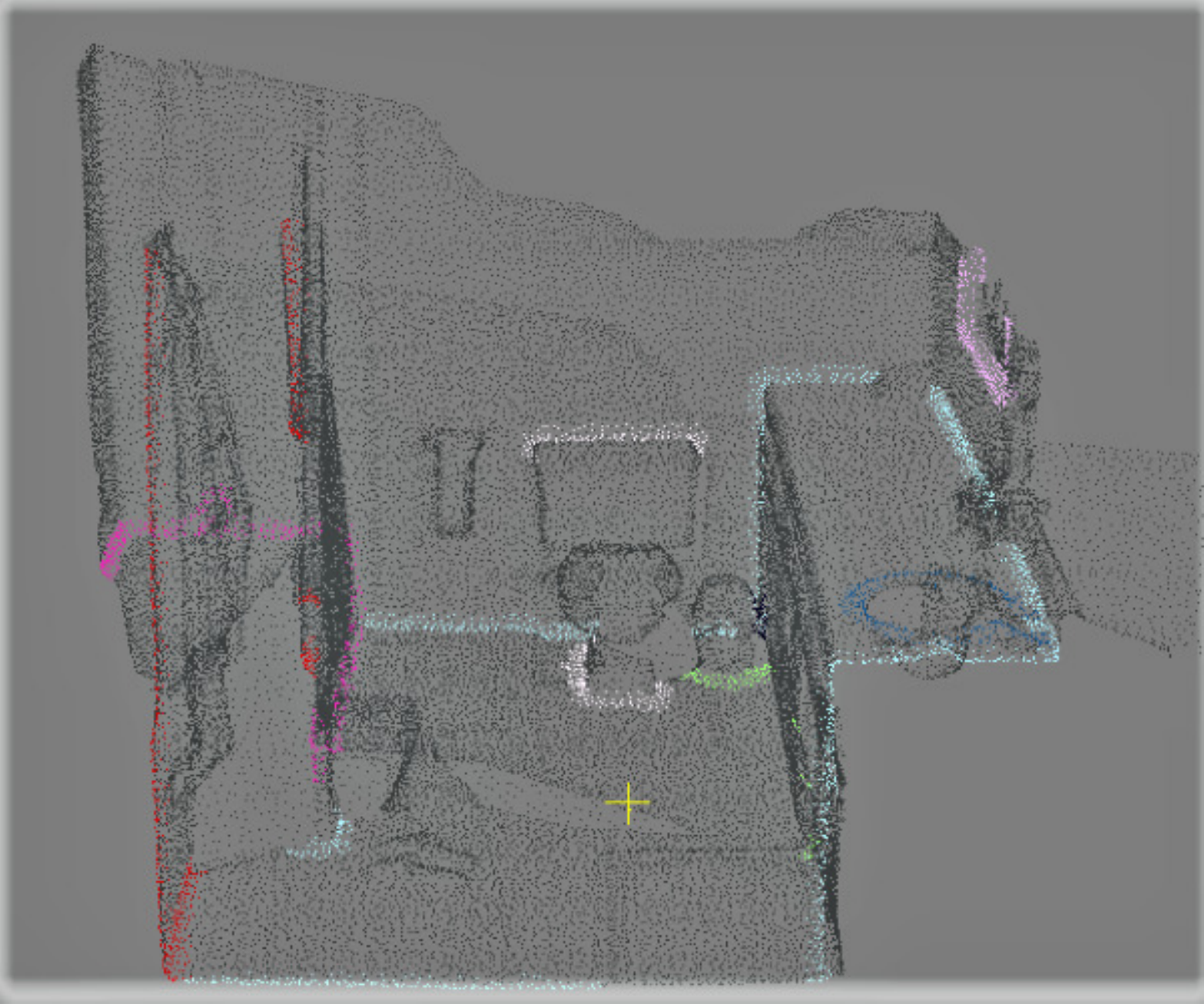}
\end{subfigure} \hfil
\begin{subfigure}{.193\textwidth}
  \centering
  \includegraphics[width=\textwidth]{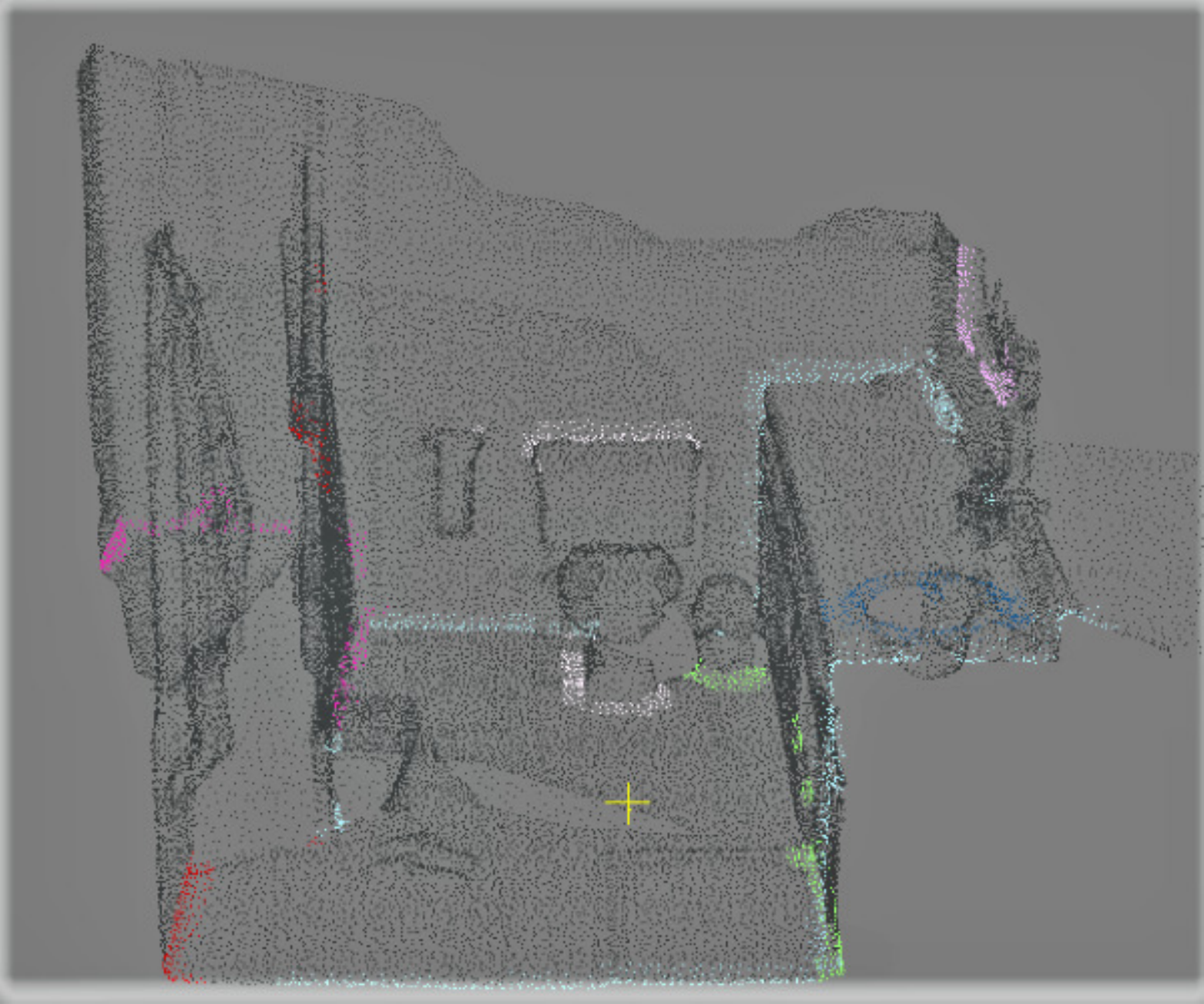}
\end{subfigure}

\begin{subfigure}{.193\textwidth}
  \centering
  \includegraphics[width=\textwidth]{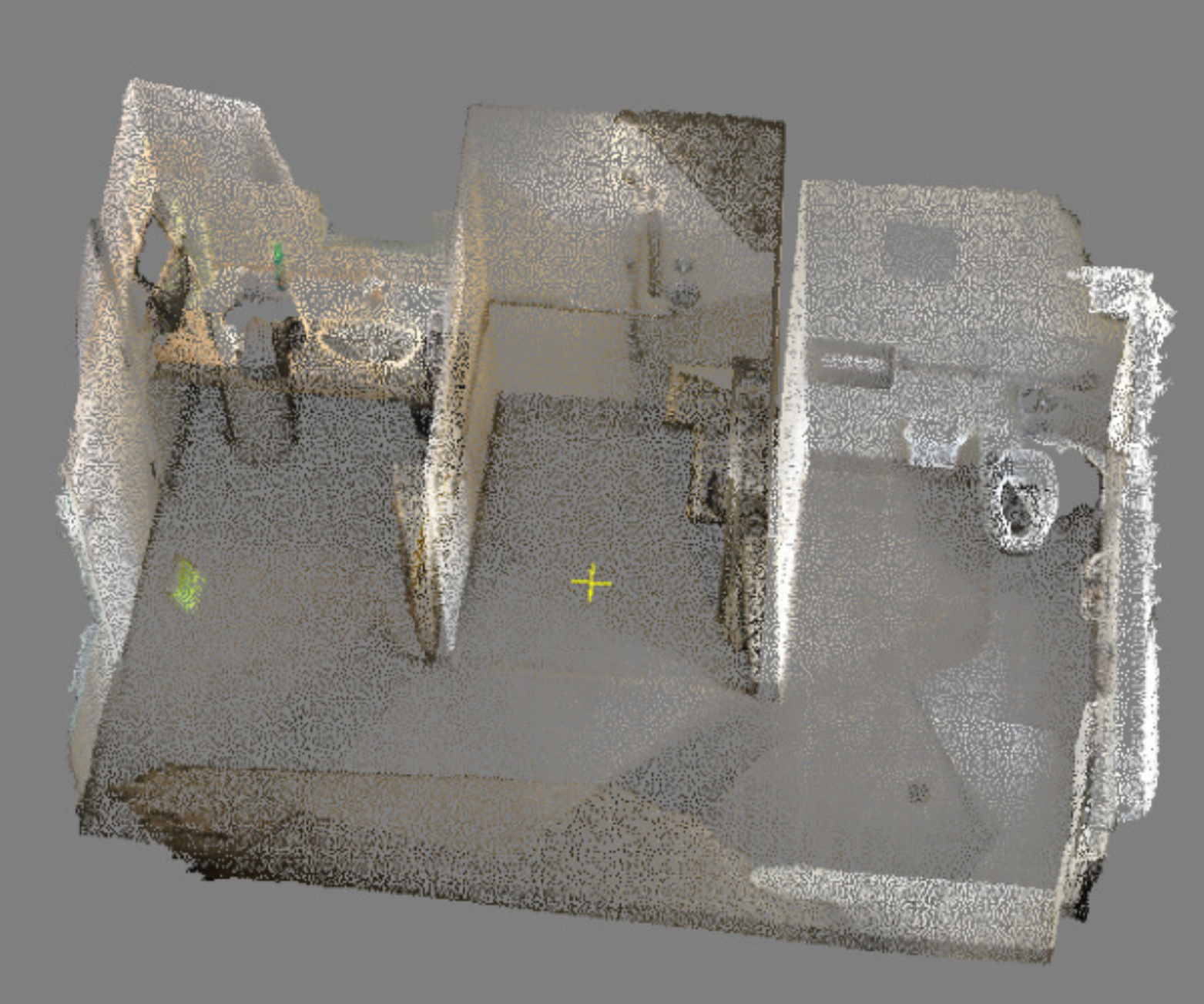}
\end{subfigure} \hfil
\begin{subfigure}{.193\textwidth}
  \centering
  \includegraphics[width=\textwidth]{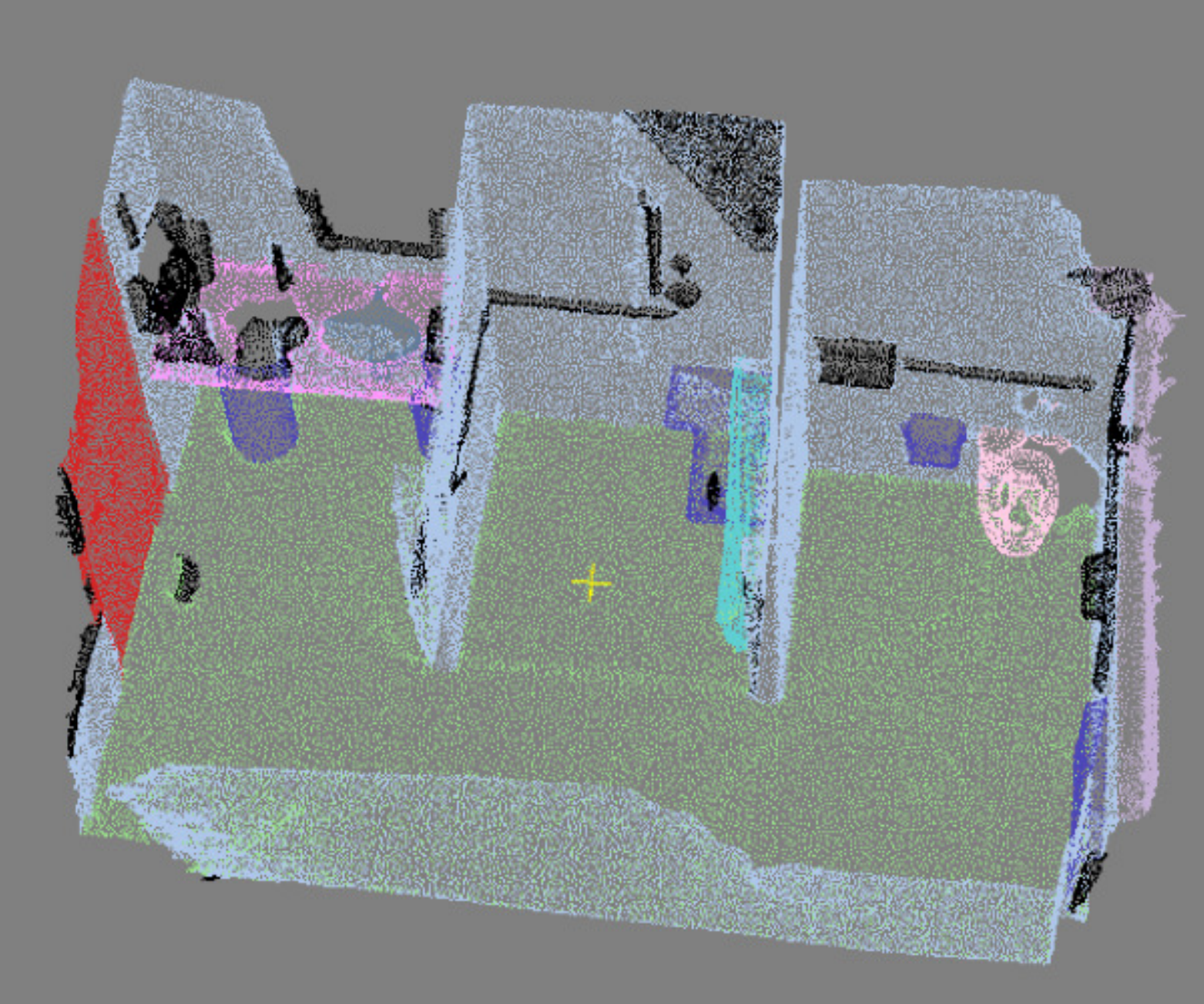}
\end{subfigure} \hfil
\begin{subfigure}{.193\textwidth}
  \centering
  \includegraphics[width=\textwidth]{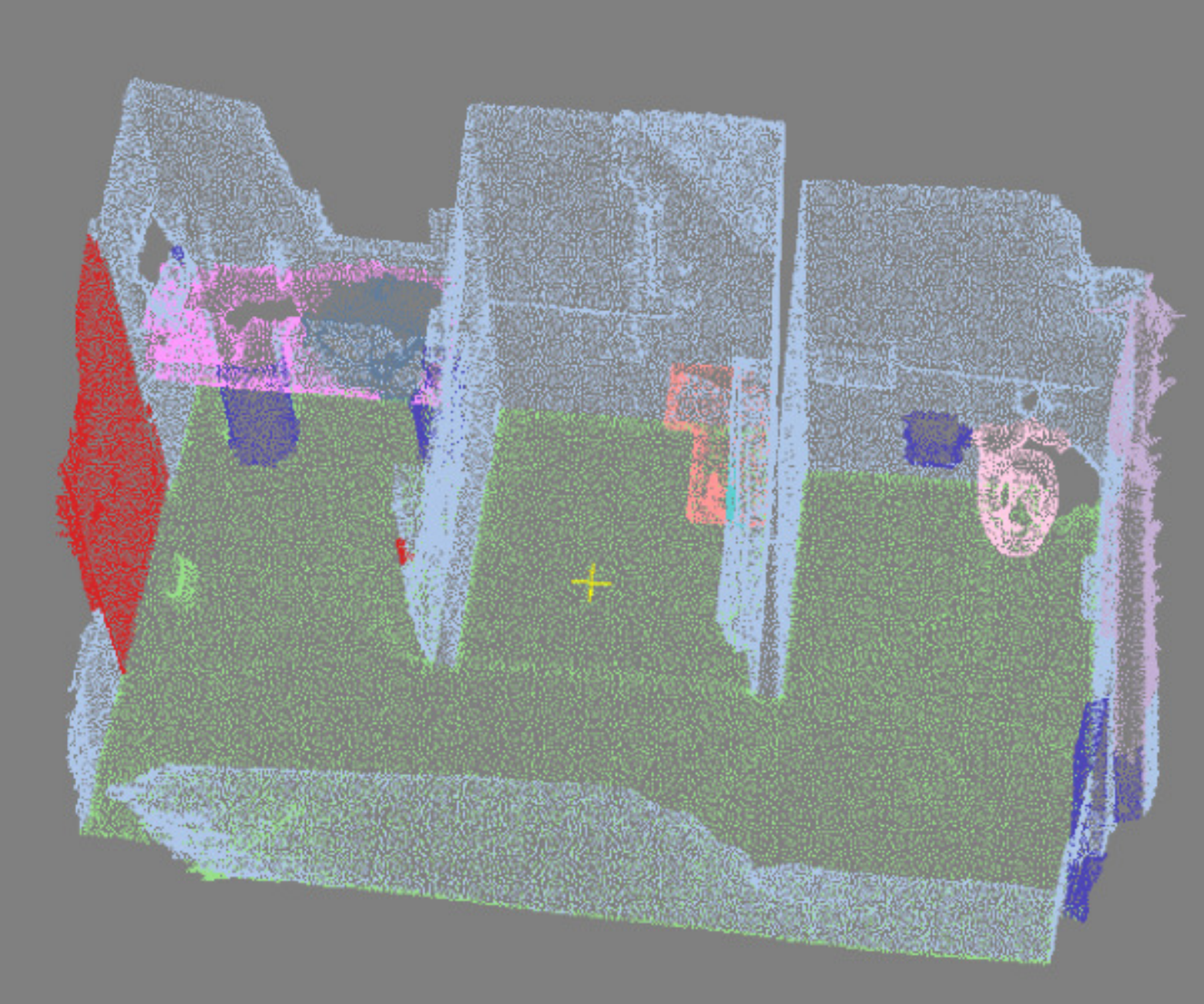}
\end{subfigure} \hfil
\begin{subfigure}{.193\textwidth}
  \centering
  \includegraphics[width=\textwidth]{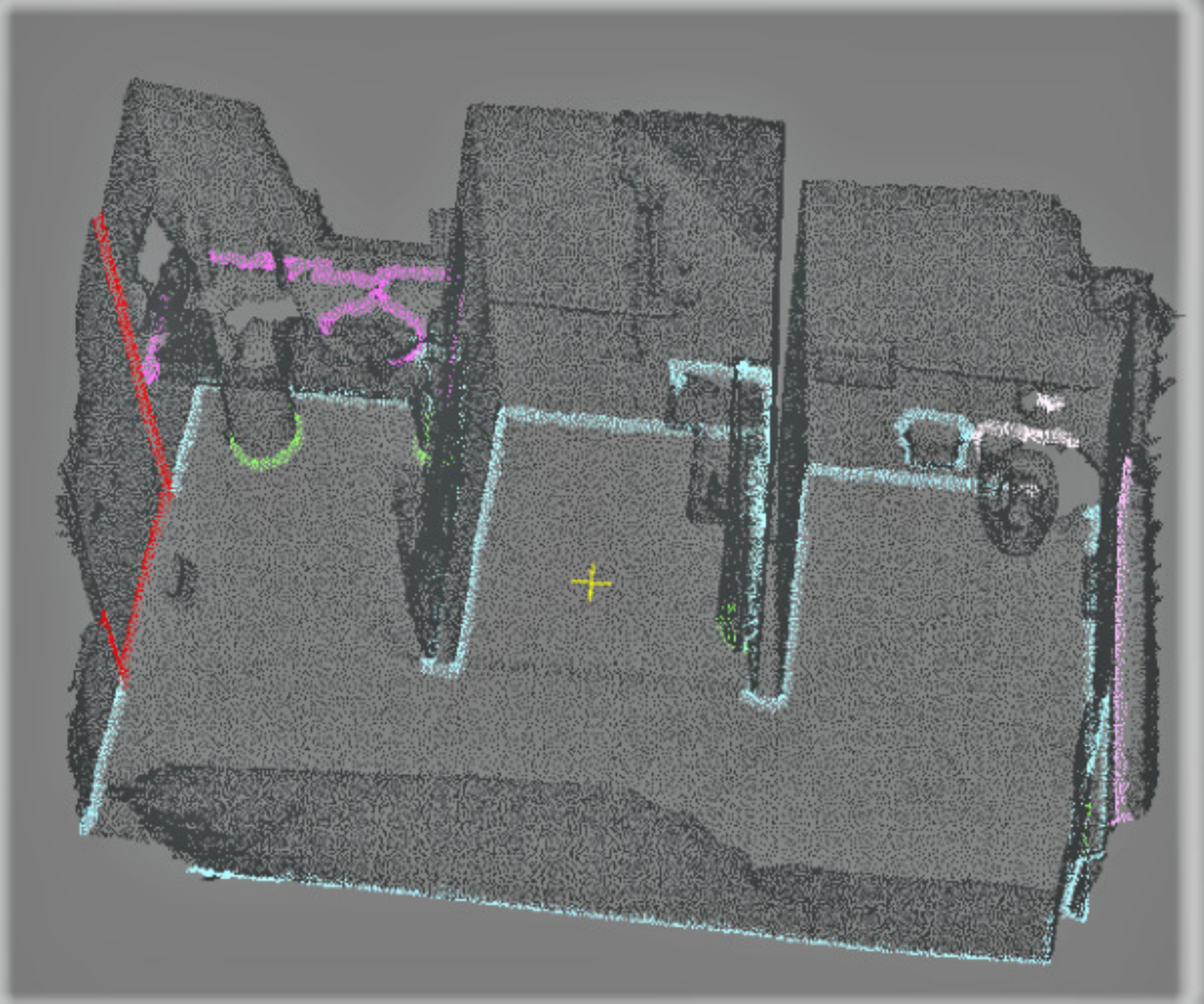}
\end{subfigure} \hfil
\begin{subfigure}{.193\textwidth}
  \centering
  \includegraphics[width=\textwidth]{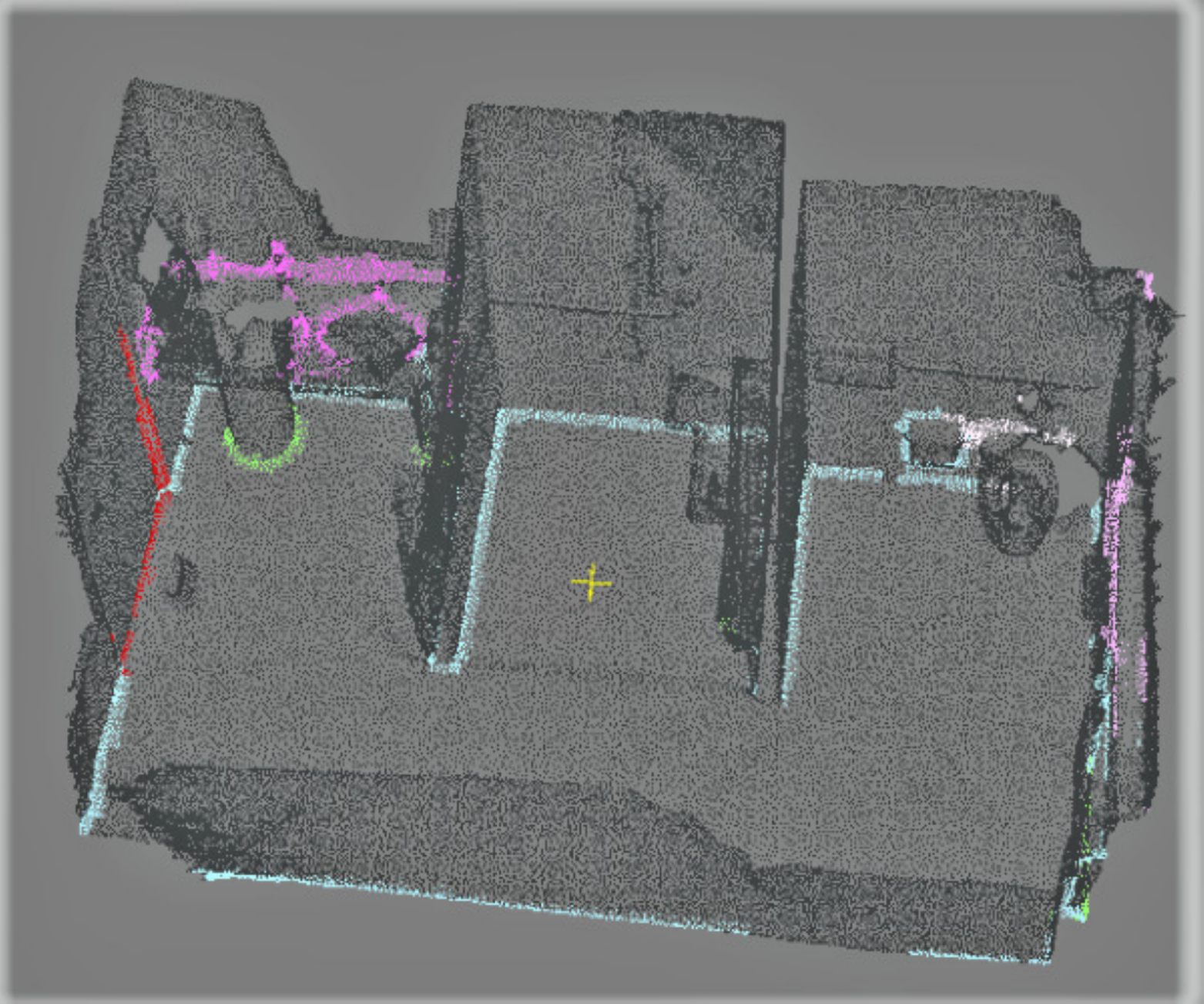}
\end{subfigure}

\begin{subfigure}{.193\textwidth}
  \centering
  \includegraphics[width=\textwidth]{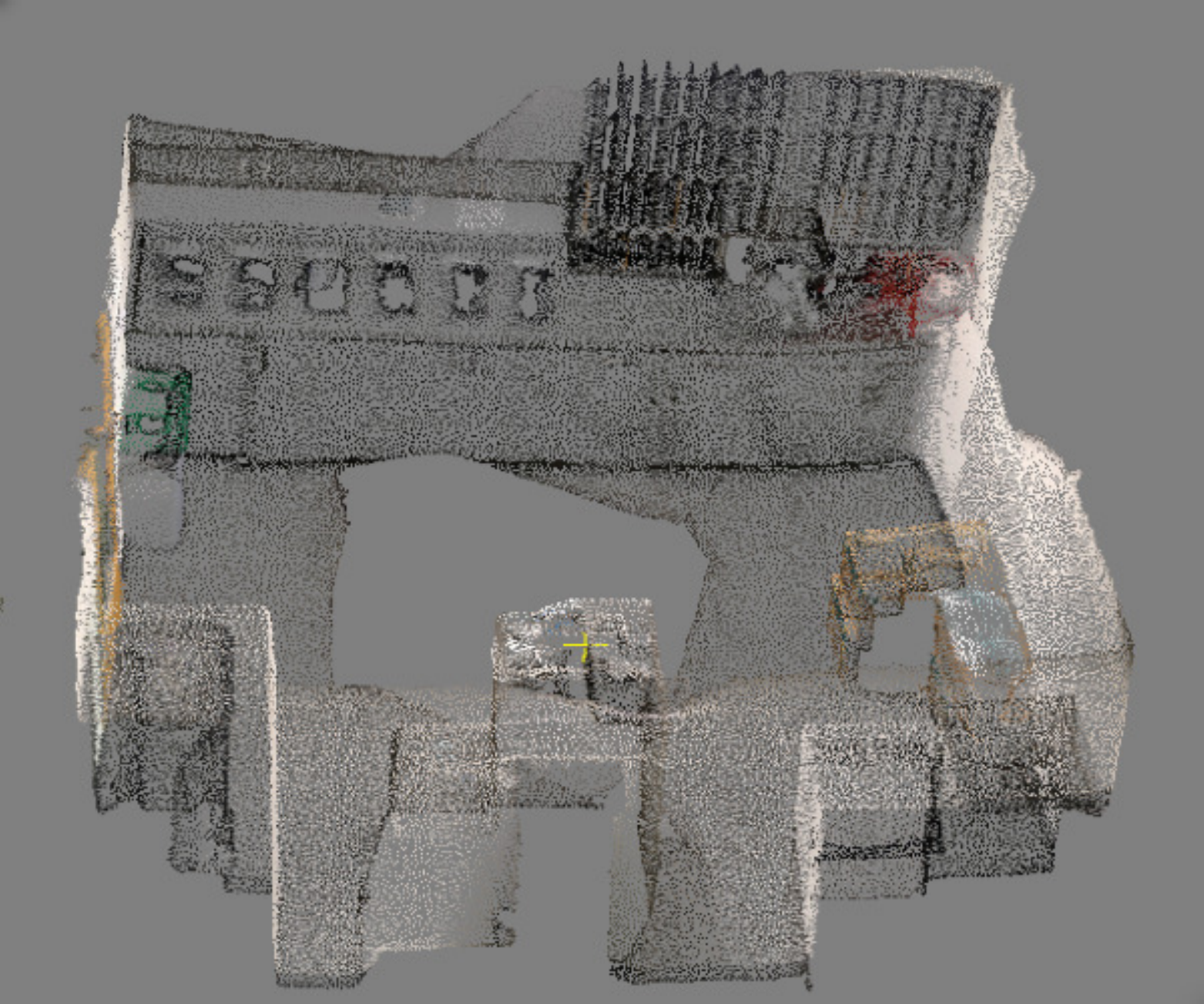}
\end{subfigure} \hfil
\begin{subfigure}{.193\textwidth}
  \centering
  \includegraphics[width=\textwidth]{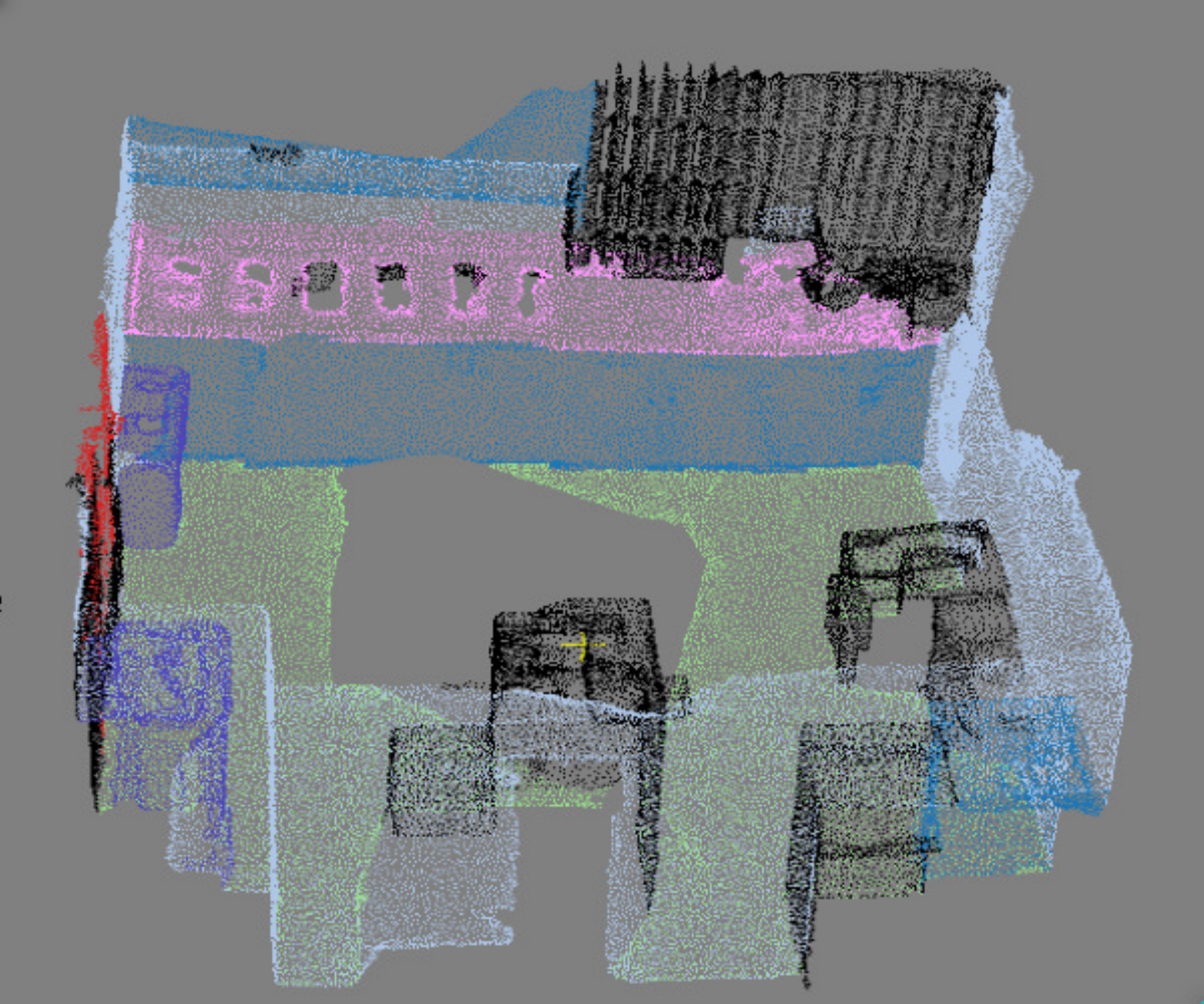}
\end{subfigure} \hfil
\begin{subfigure}{.193\textwidth}
  \centering
  \includegraphics[width=\textwidth]{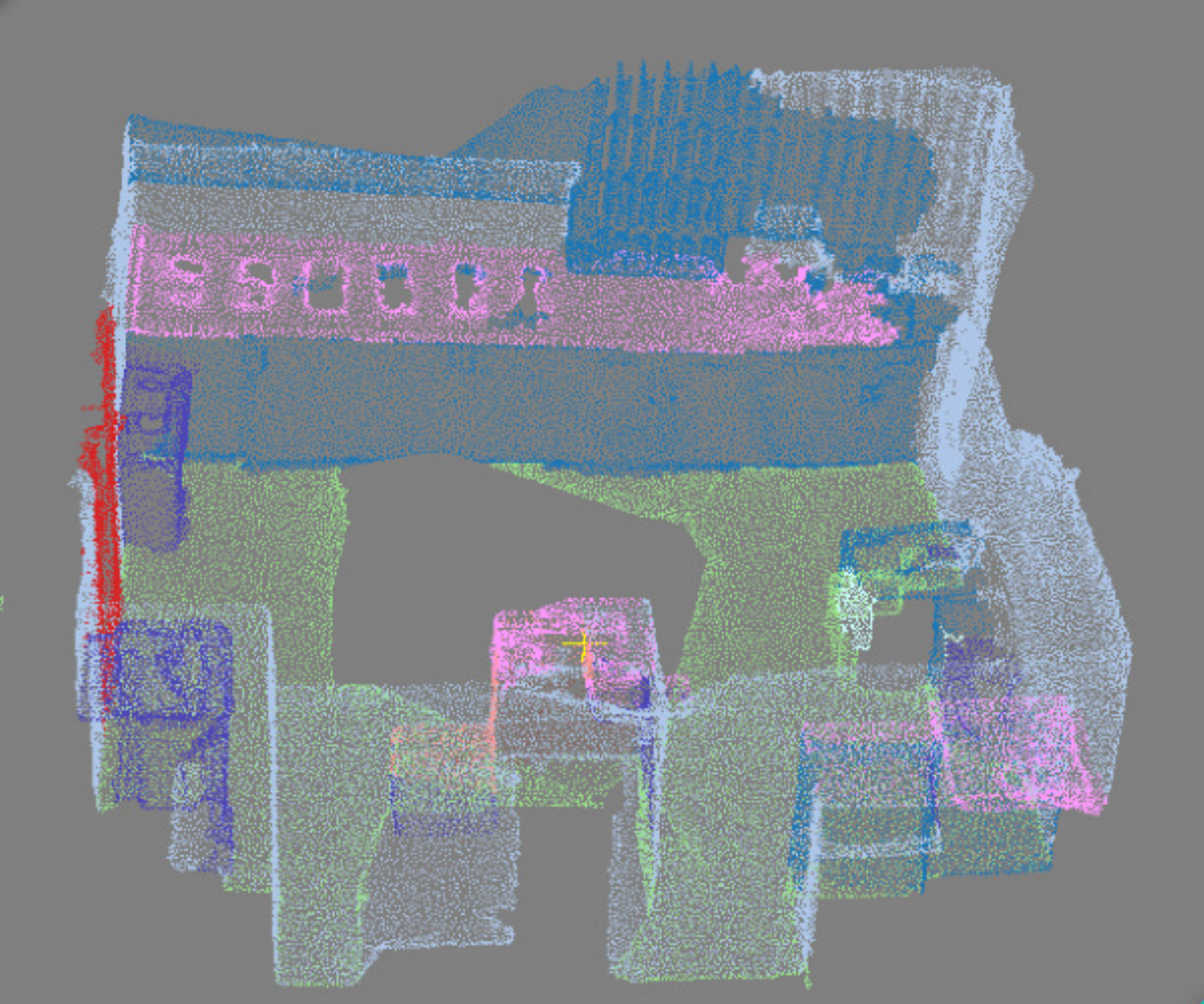}
\end{subfigure} \hfil
\begin{subfigure}{.193\textwidth}
  \centering
  \includegraphics[width=\textwidth]{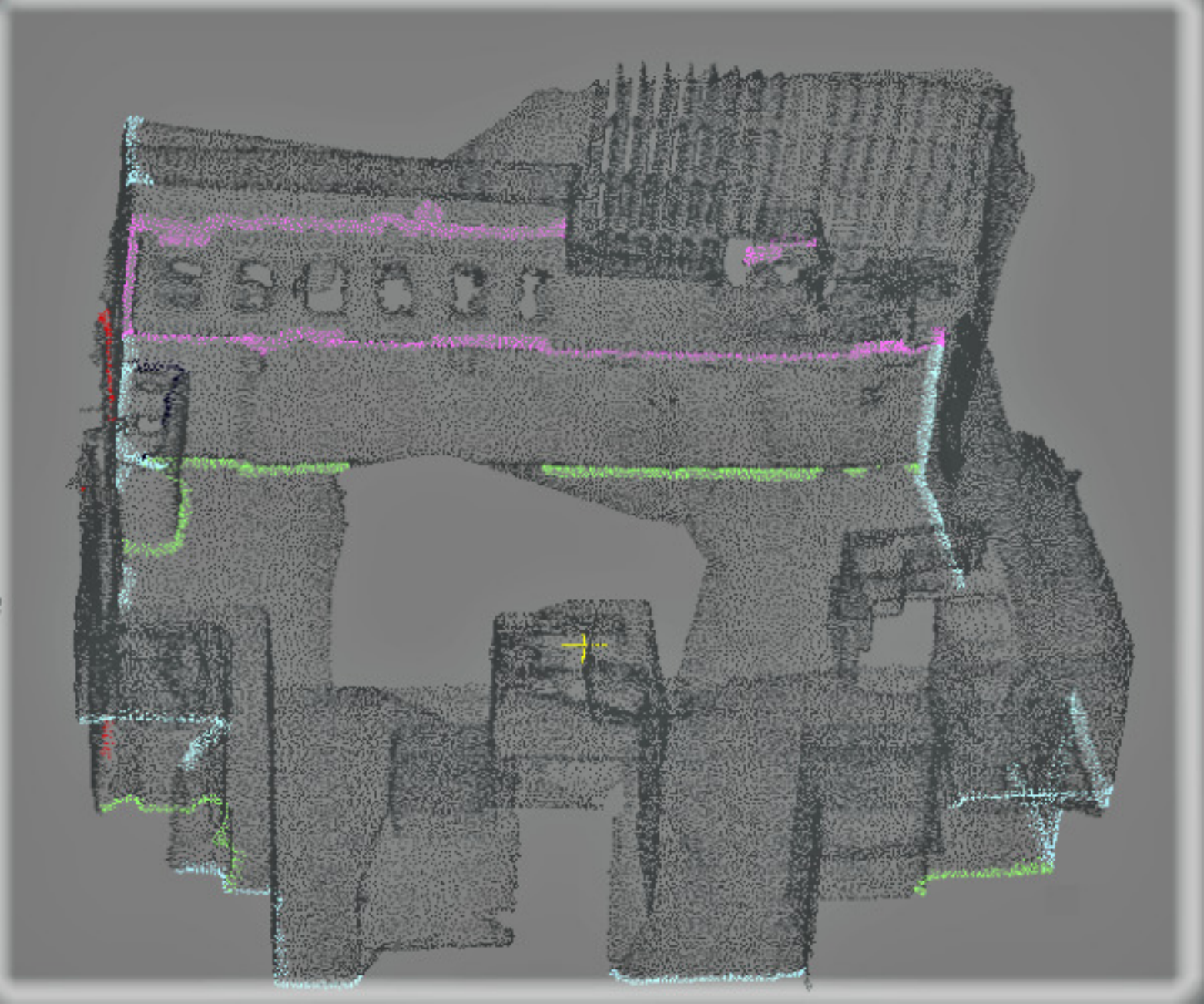}
\end{subfigure} \hfil
\begin{subfigure}{.193\textwidth}
  \centering
  \includegraphics[width=\textwidth]{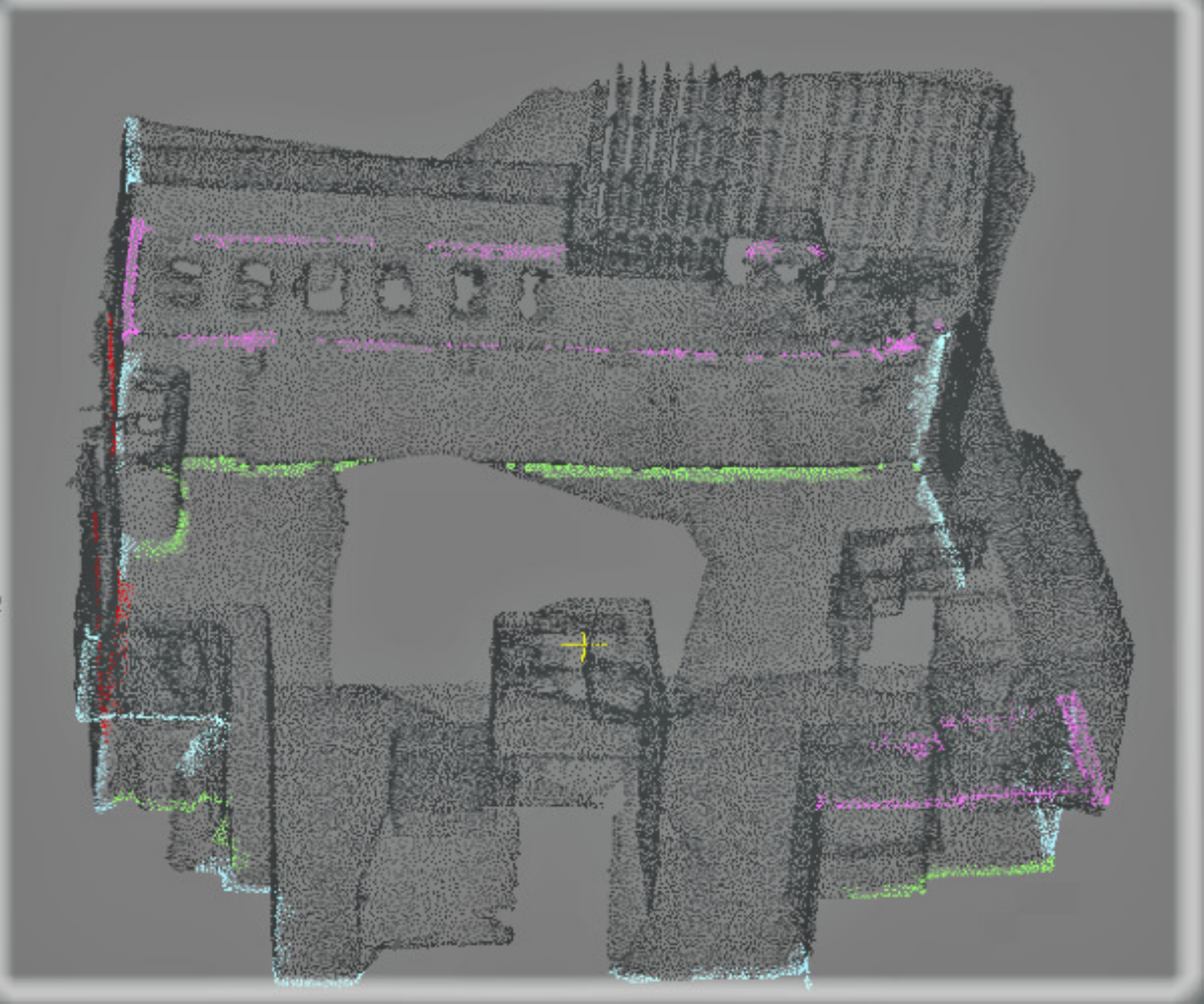}
\end{subfigure}

\begin{subfigure}{.193\textwidth}
  \centering
  \includegraphics[width=\textwidth]{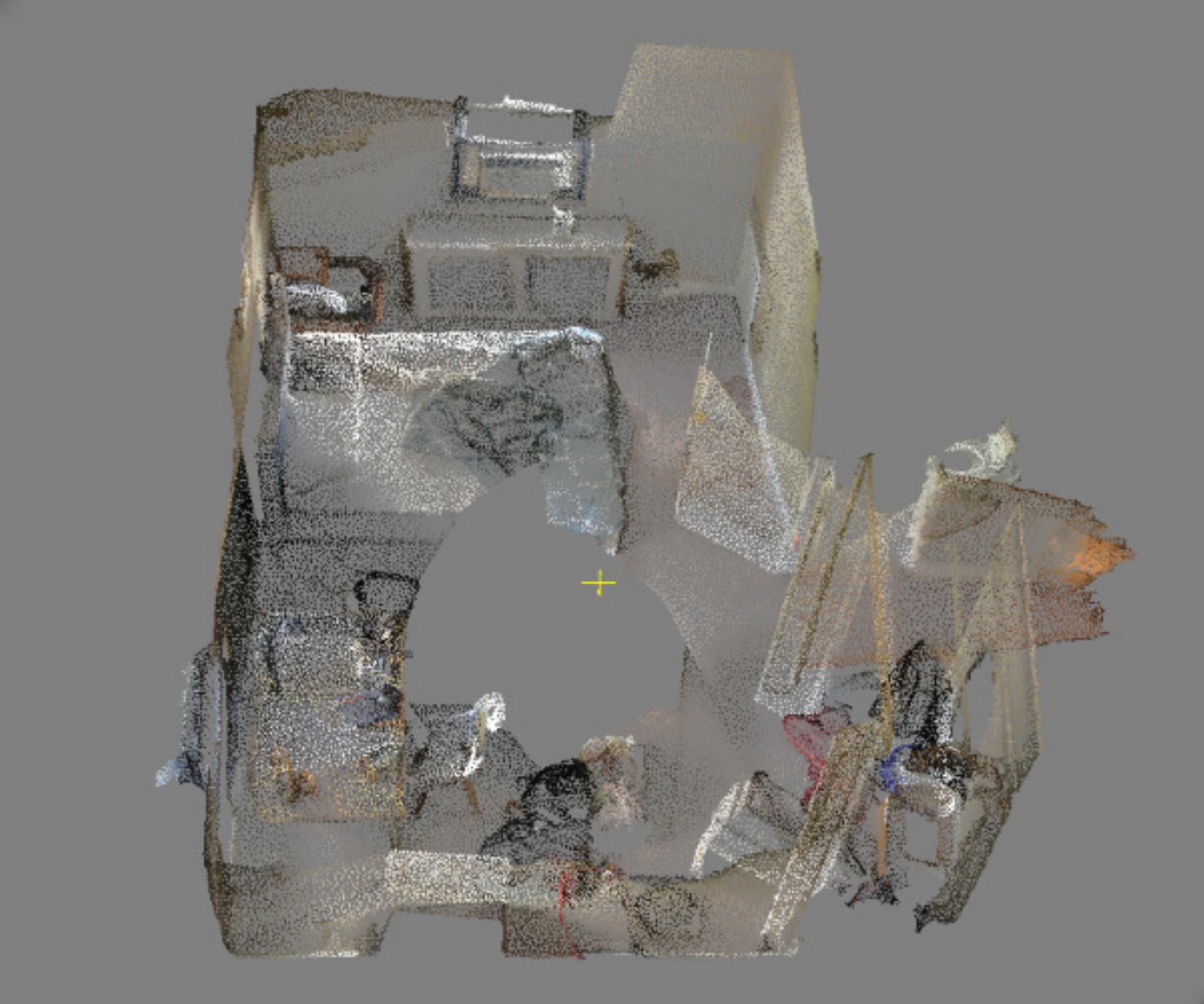}
\end{subfigure} \hfil
\begin{subfigure}{.193\textwidth}
  \centering
  \includegraphics[width=\textwidth]{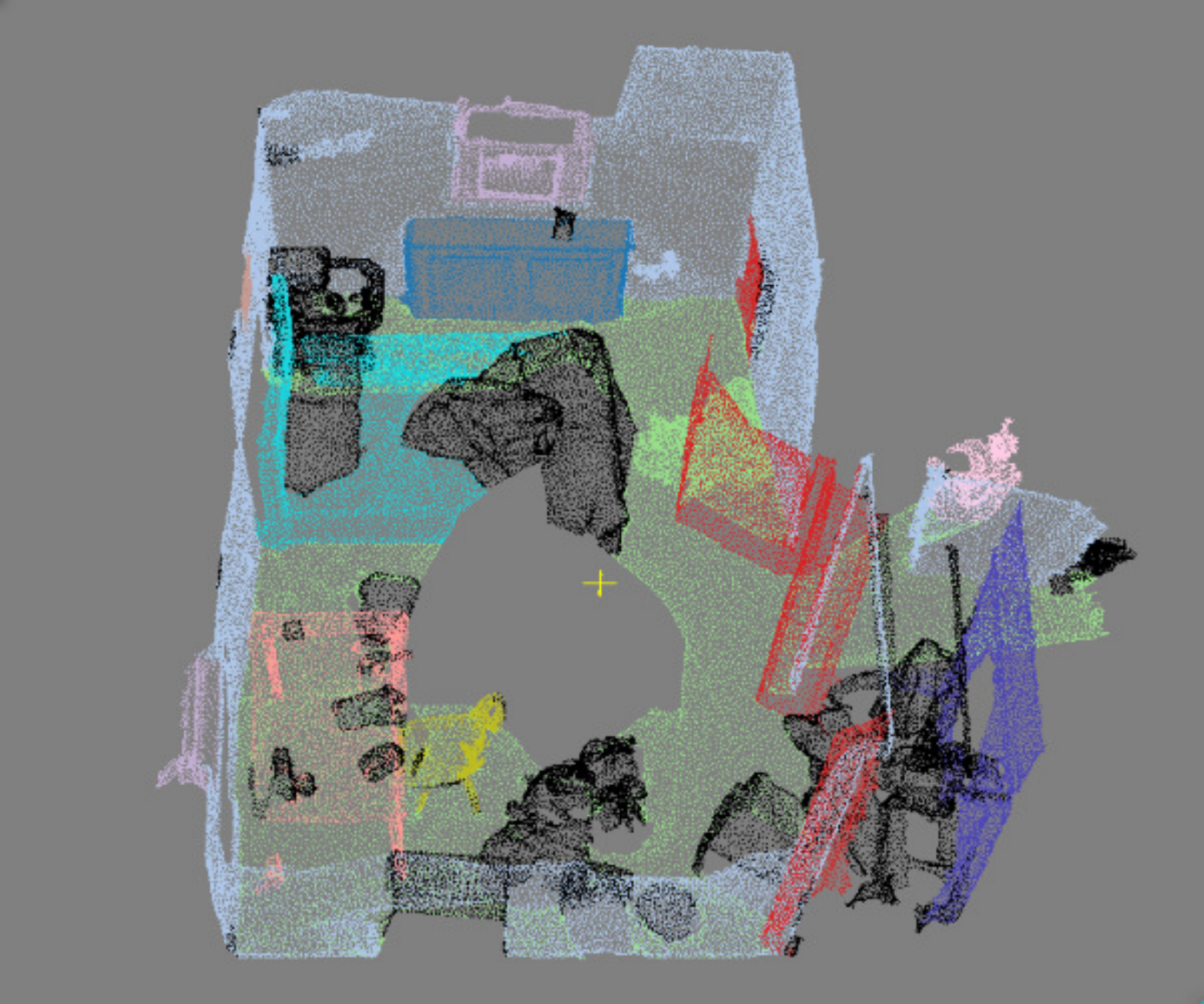}
\end{subfigure} \hfil
\begin{subfigure}{.193\textwidth}
  \centering
  \includegraphics[width=\textwidth]{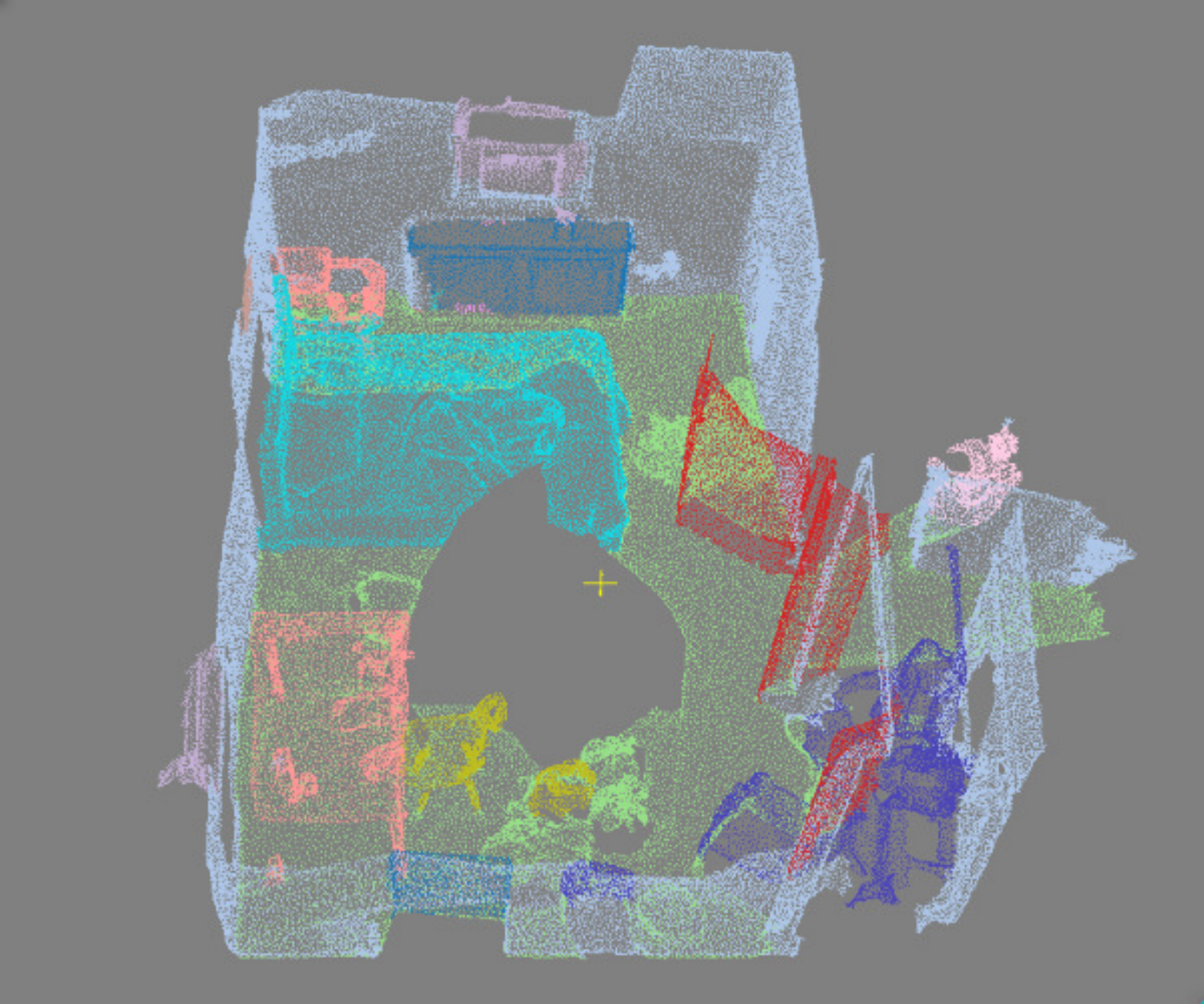}
\end{subfigure} \hfil
\begin{subfigure}{.193\textwidth}
  \centering
  \includegraphics[width=\textwidth]{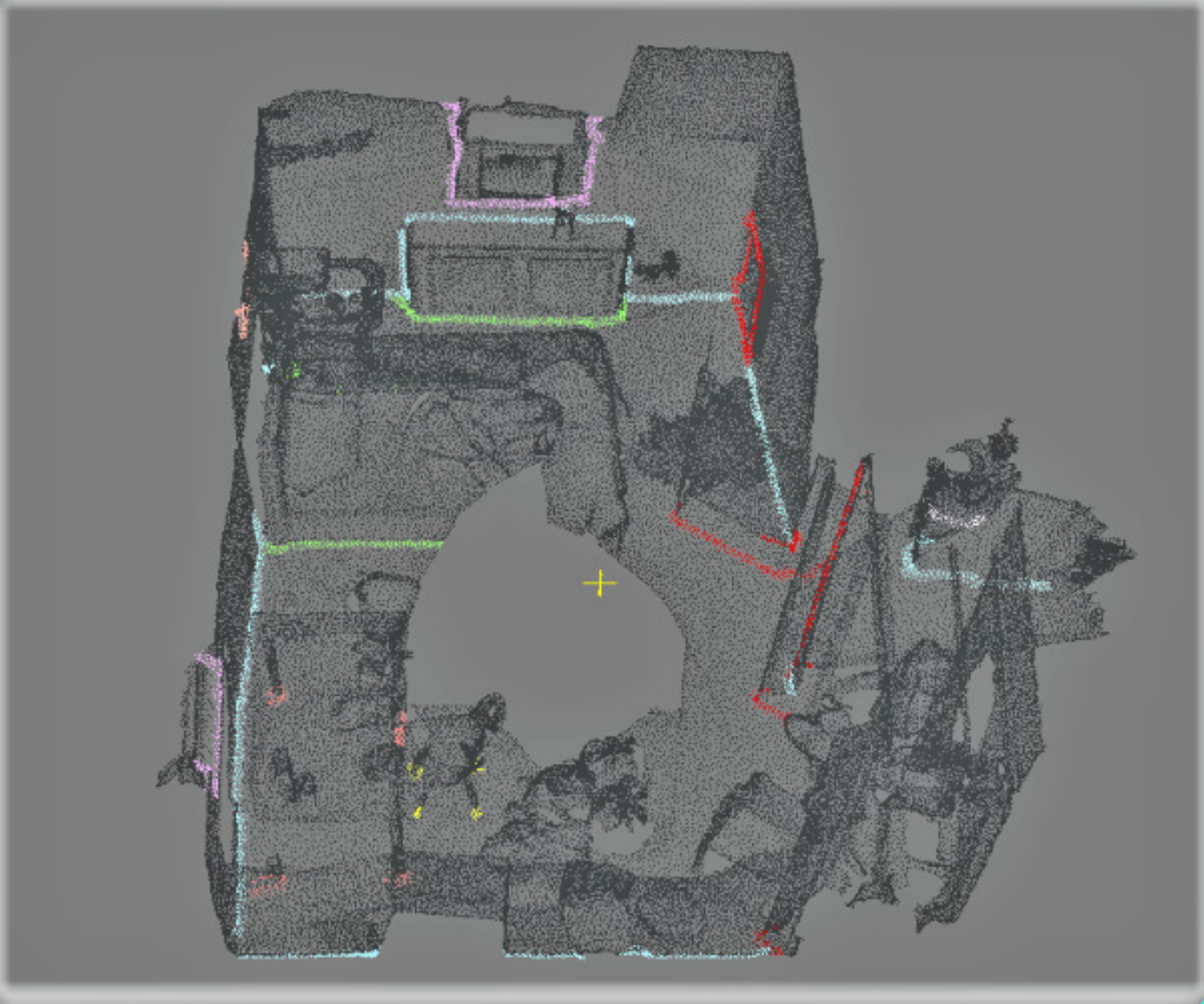}
\end{subfigure} \hfil
\begin{subfigure}{.193\textwidth}
  \centering
  \includegraphics[width=\textwidth]{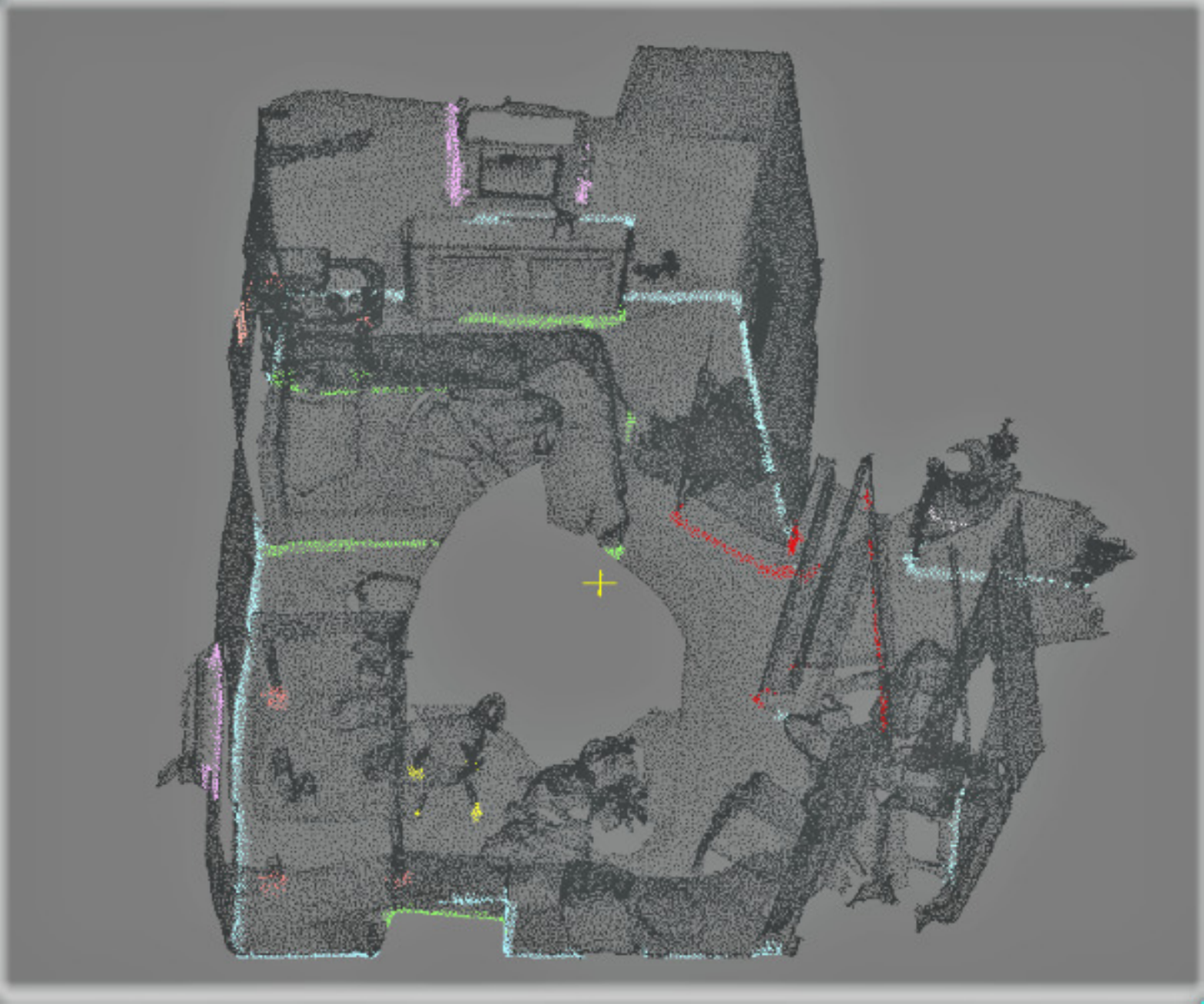}
\end{subfigure}

\begin{subfigure}{.193\textwidth}
  \centering
  \includegraphics[width=\textwidth]{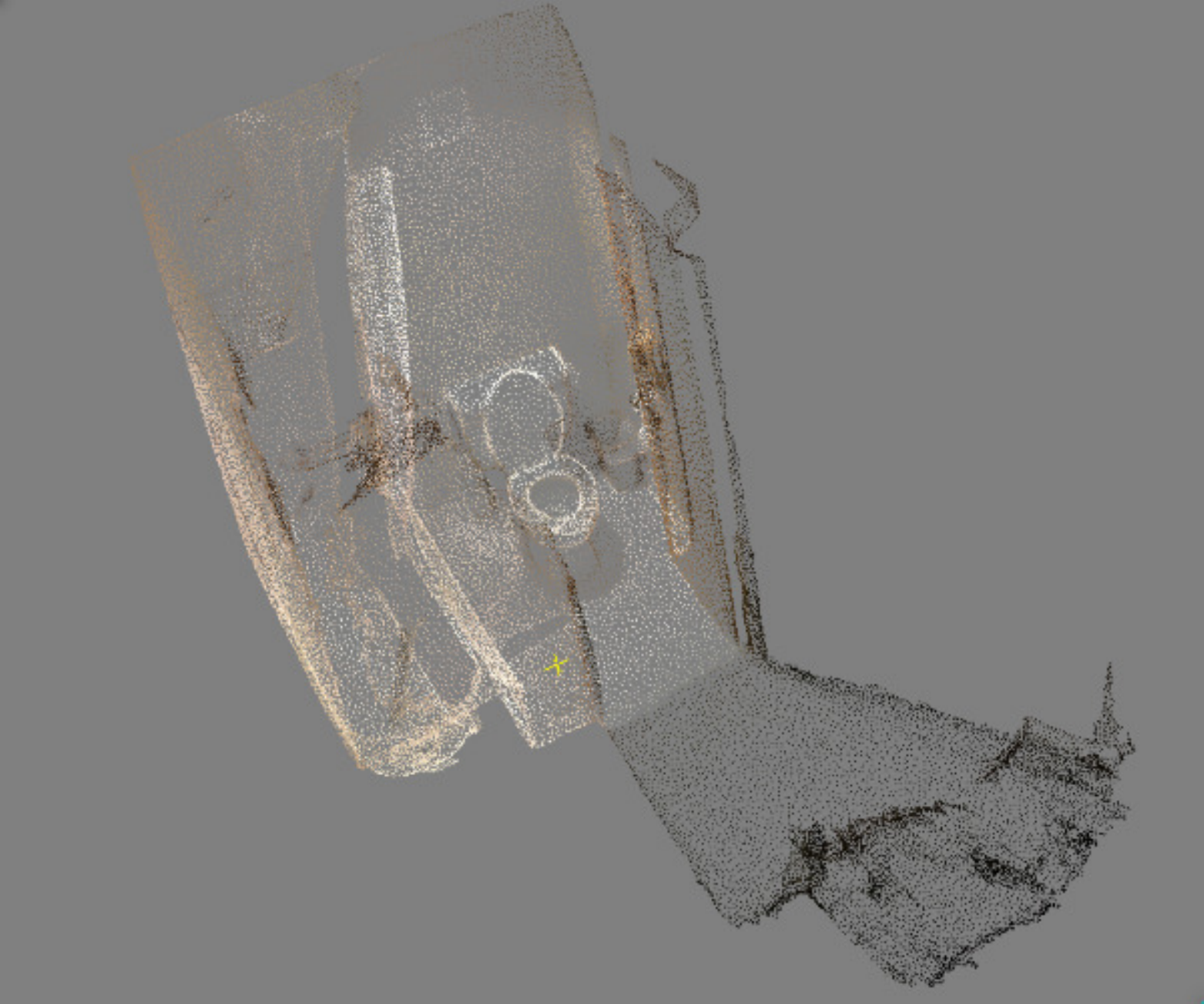}
\end{subfigure} \hfil
\begin{subfigure}{.193\textwidth}
  \centering
  \includegraphics[width=\textwidth]{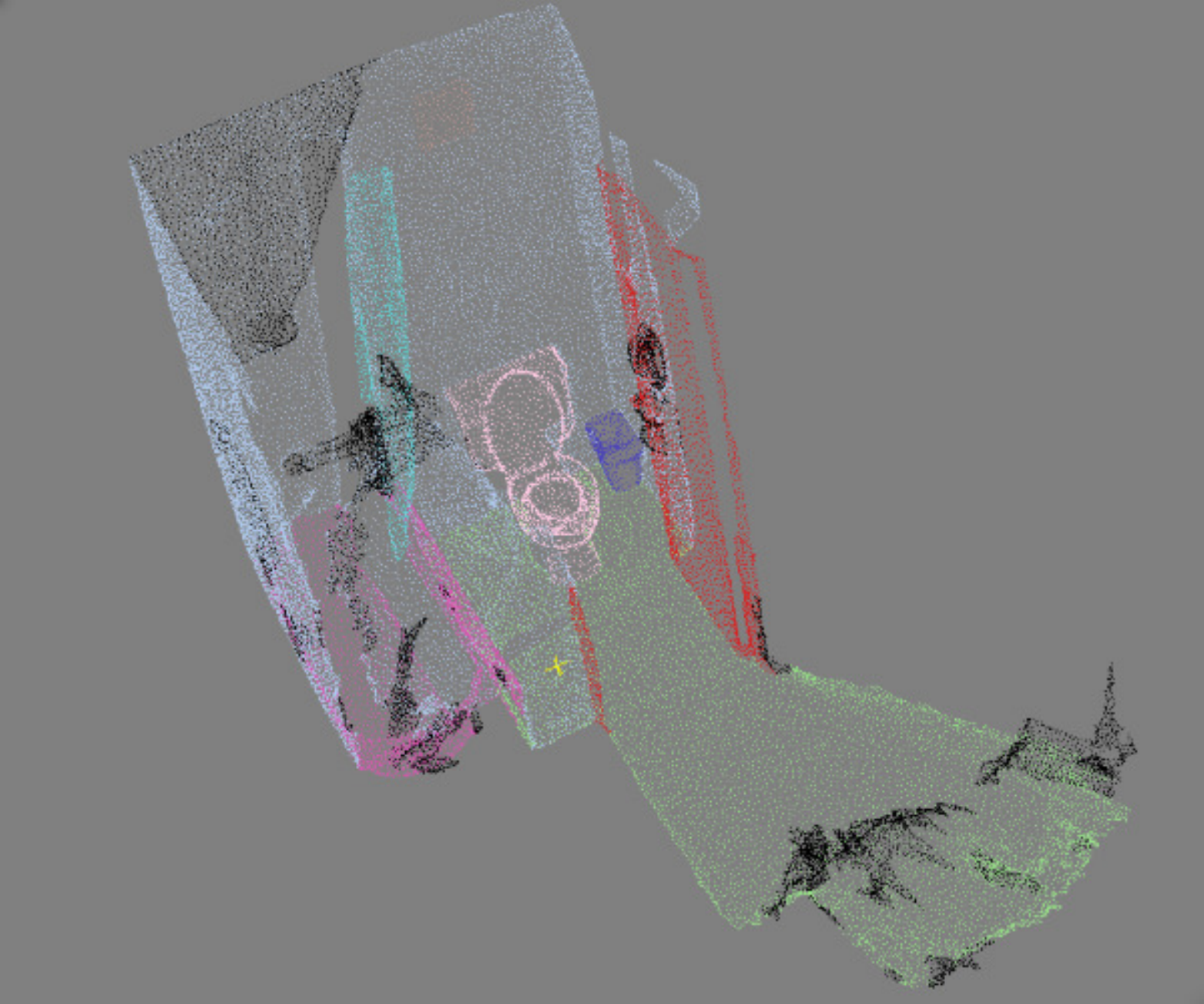}
\end{subfigure} \hfil
\begin{subfigure}{.193\textwidth}
  \centering
  \includegraphics[width=\textwidth]{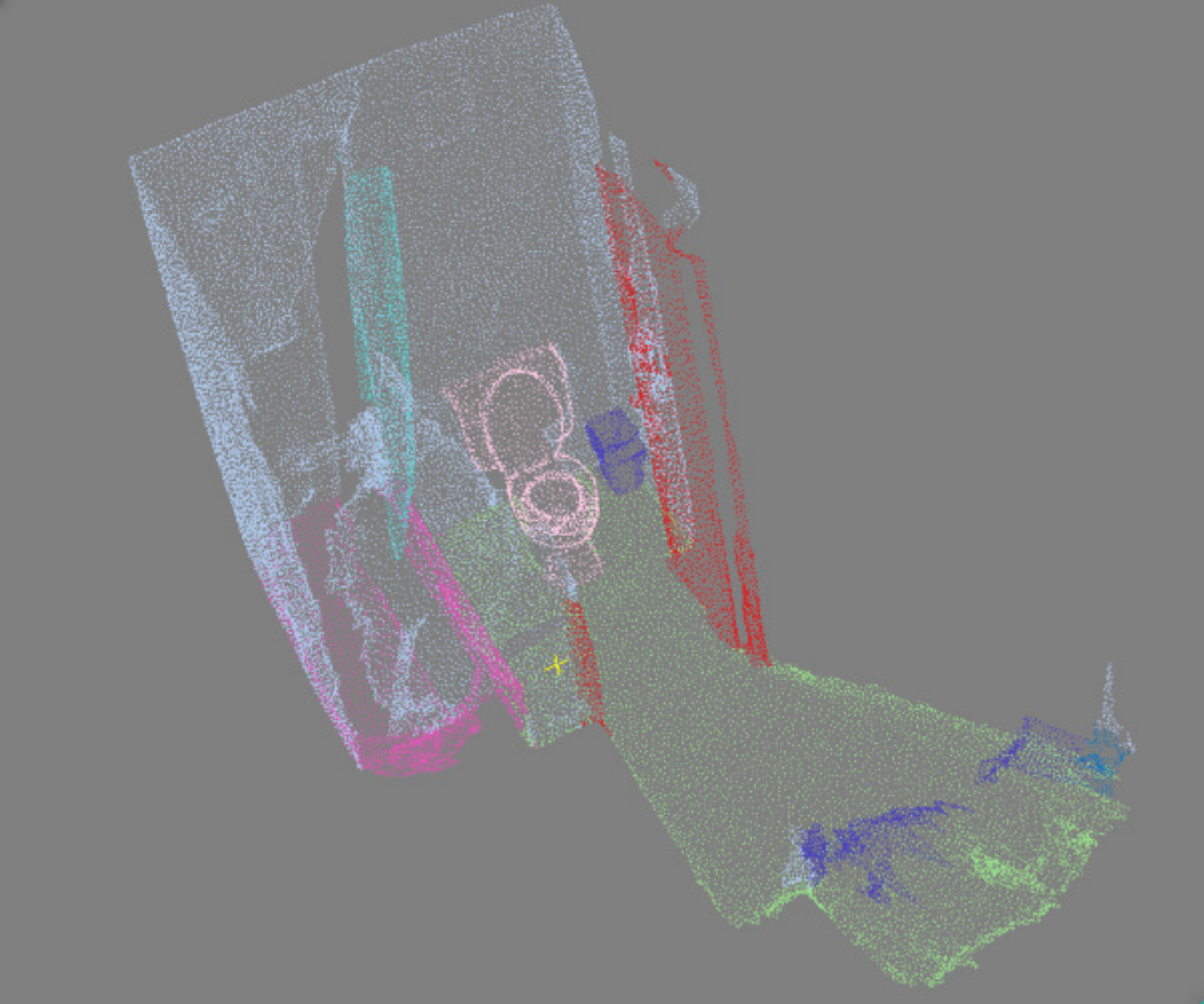}
\end{subfigure} \hfil
\begin{subfigure}{.193\textwidth}
  \centering
  \includegraphics[width=\textwidth]{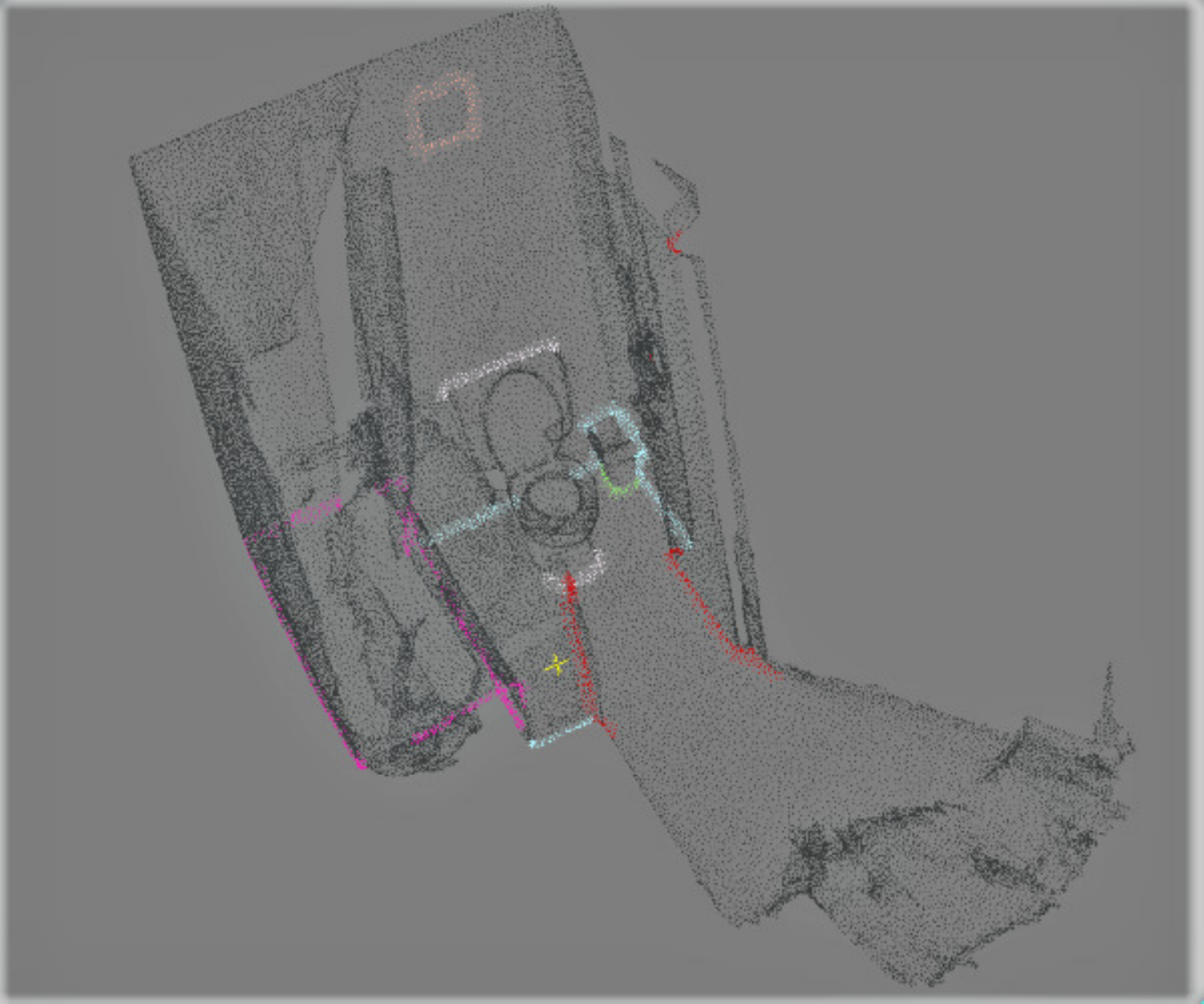}
\end{subfigure} \hfil
\begin{subfigure}{.193\textwidth}
  \centering
  \includegraphics[width=\textwidth]{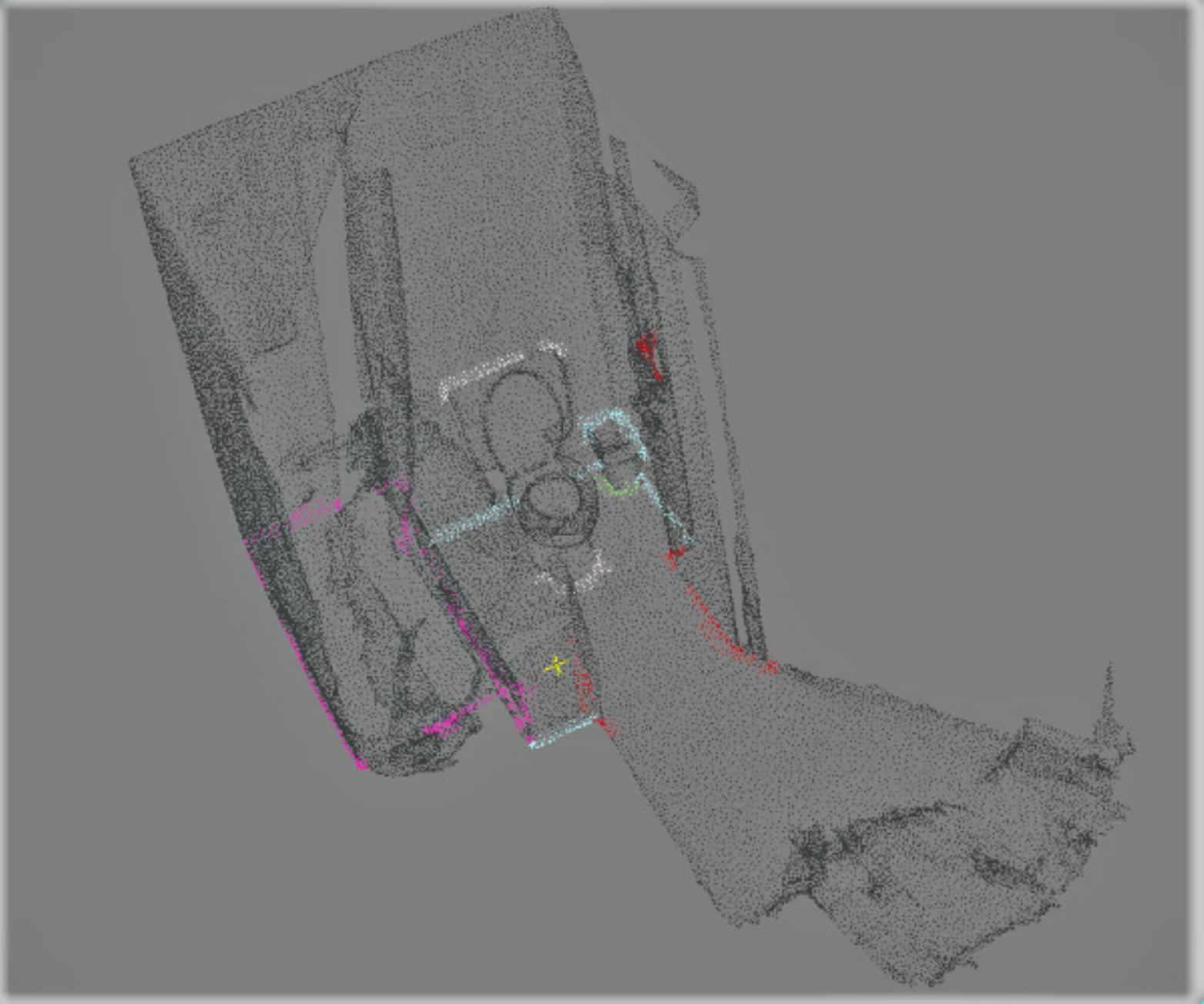}
\end{subfigure}

\begin{subfigure}{.193\textwidth}
  \centering
  \includegraphics[width=\textwidth]{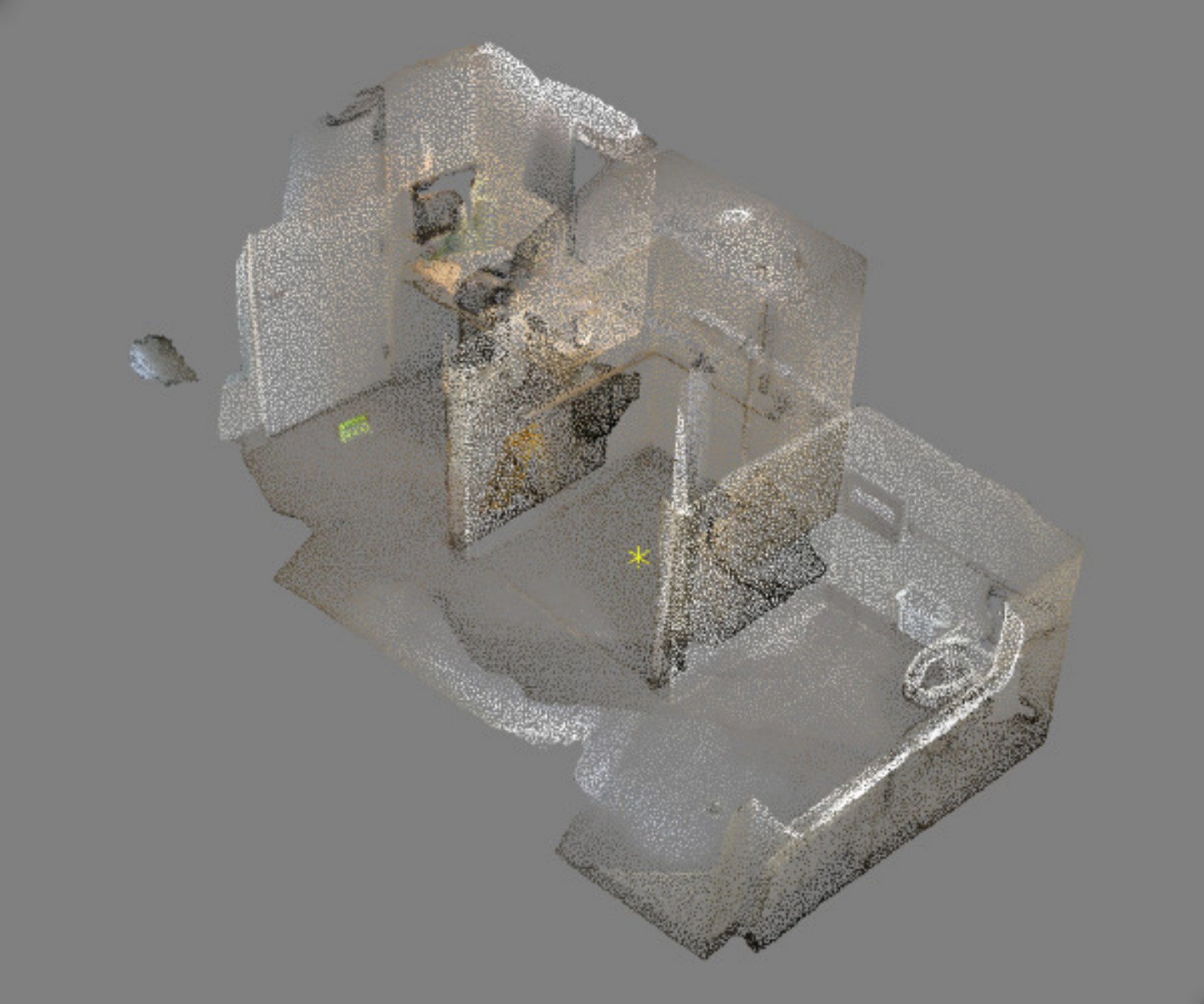}
  \caption{Point Cloud}
\end{subfigure} \hfil
\begin{subfigure}{.193\textwidth}
  \centering
  \includegraphics[width=\textwidth]{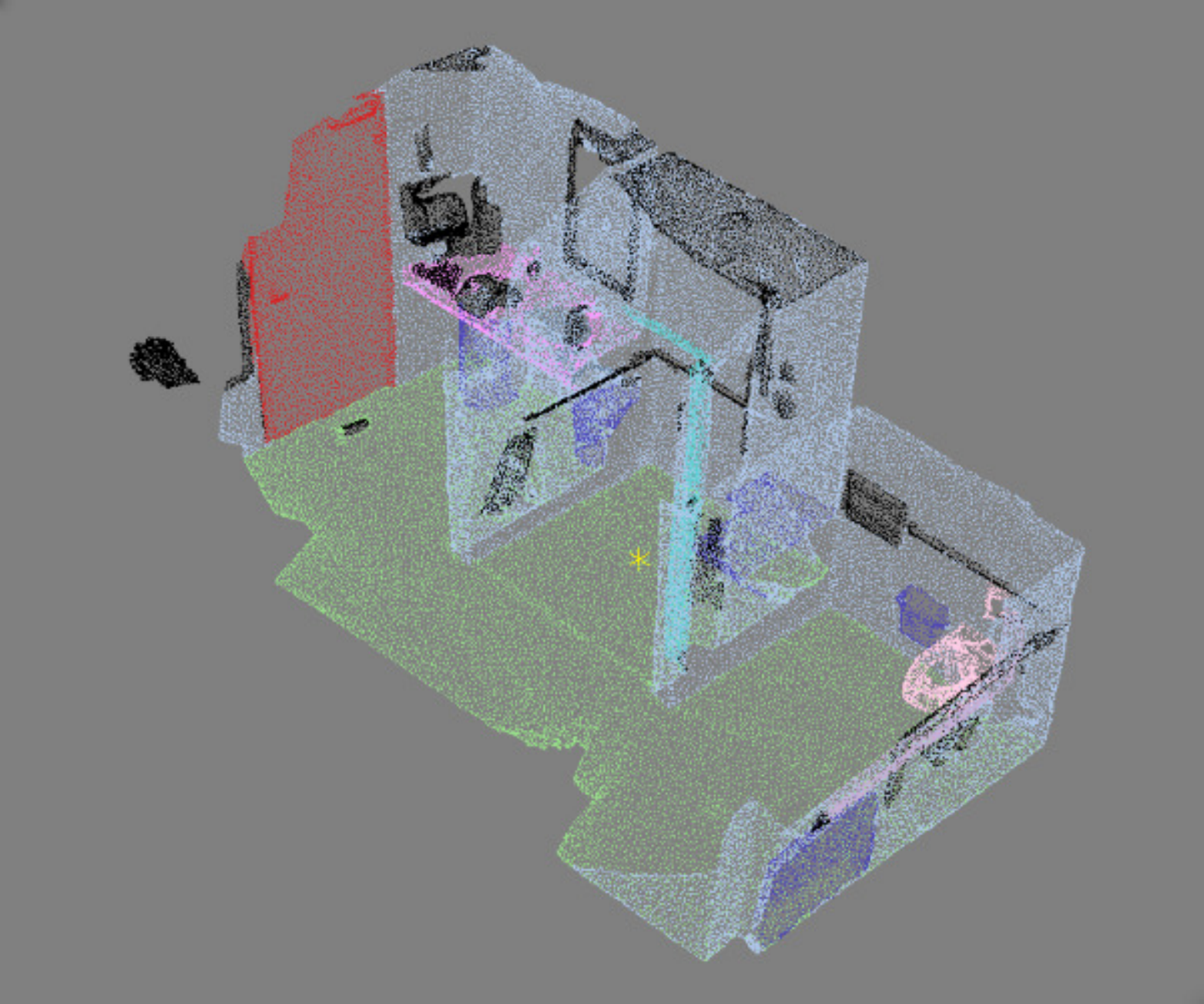}
  \caption{GT-SSP Mask}
\end{subfigure} \hfil
\begin{subfigure}{.193\textwidth}
  \centering
  \includegraphics[width=\textwidth]{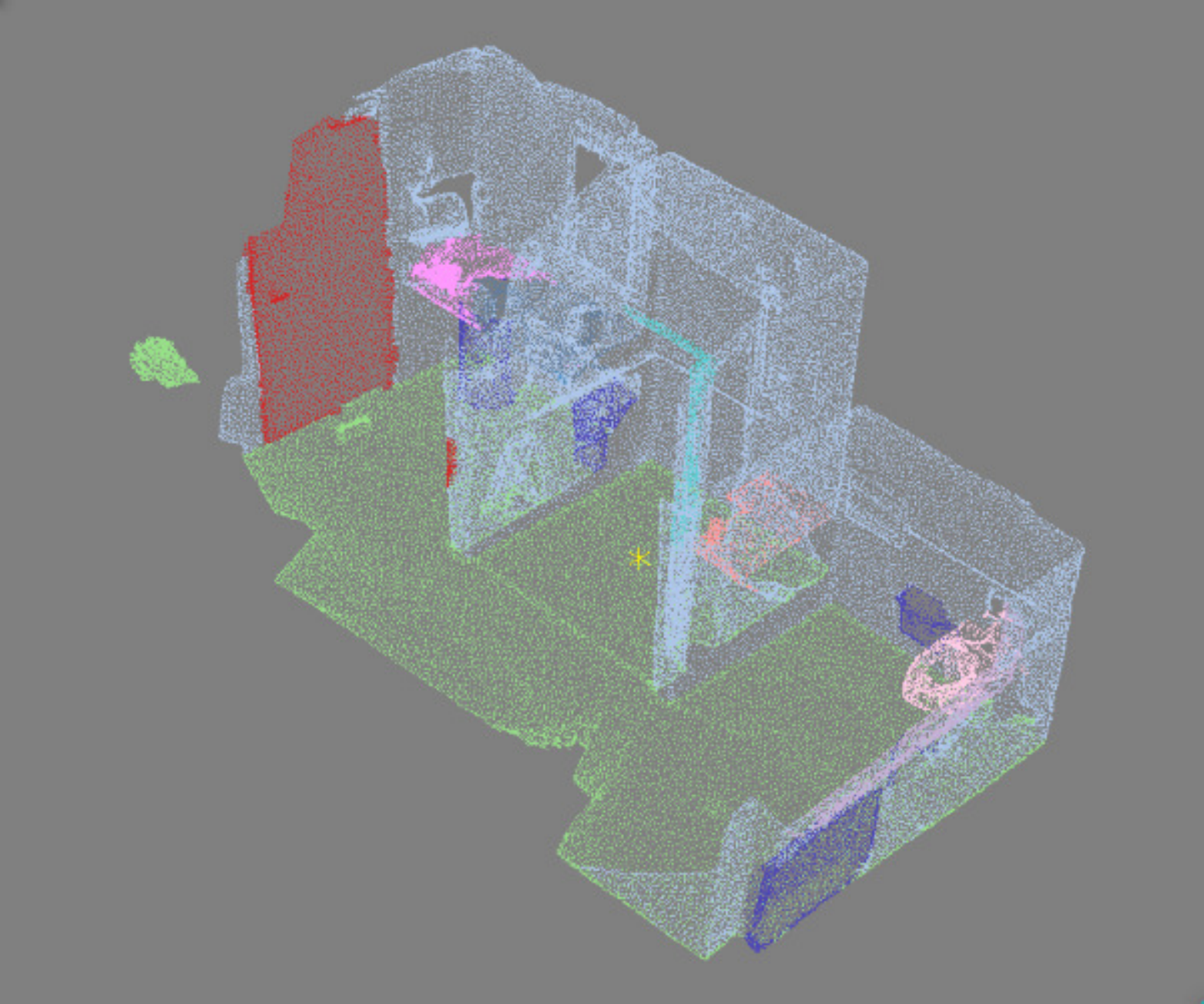}
  \caption{Pred-SSP Mask}
\end{subfigure} \hfil
\begin{subfigure}{.193\textwidth}
  \centering
  \includegraphics[width=\textwidth]{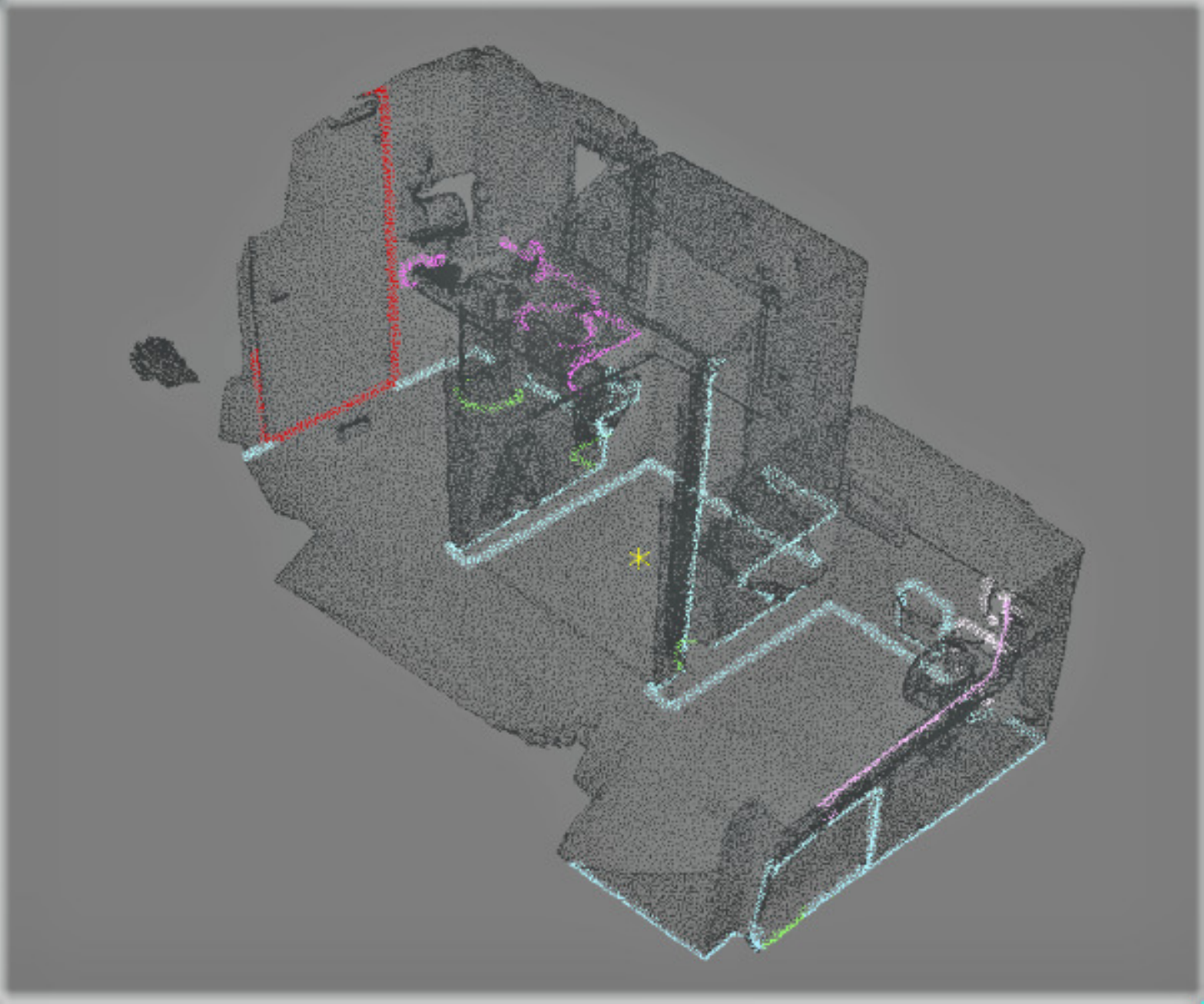}
  \caption{GT-SEP Map}
\end{subfigure} \hfil
\begin{subfigure}{.193\textwidth}
  \centering
  \includegraphics[width=\textwidth]{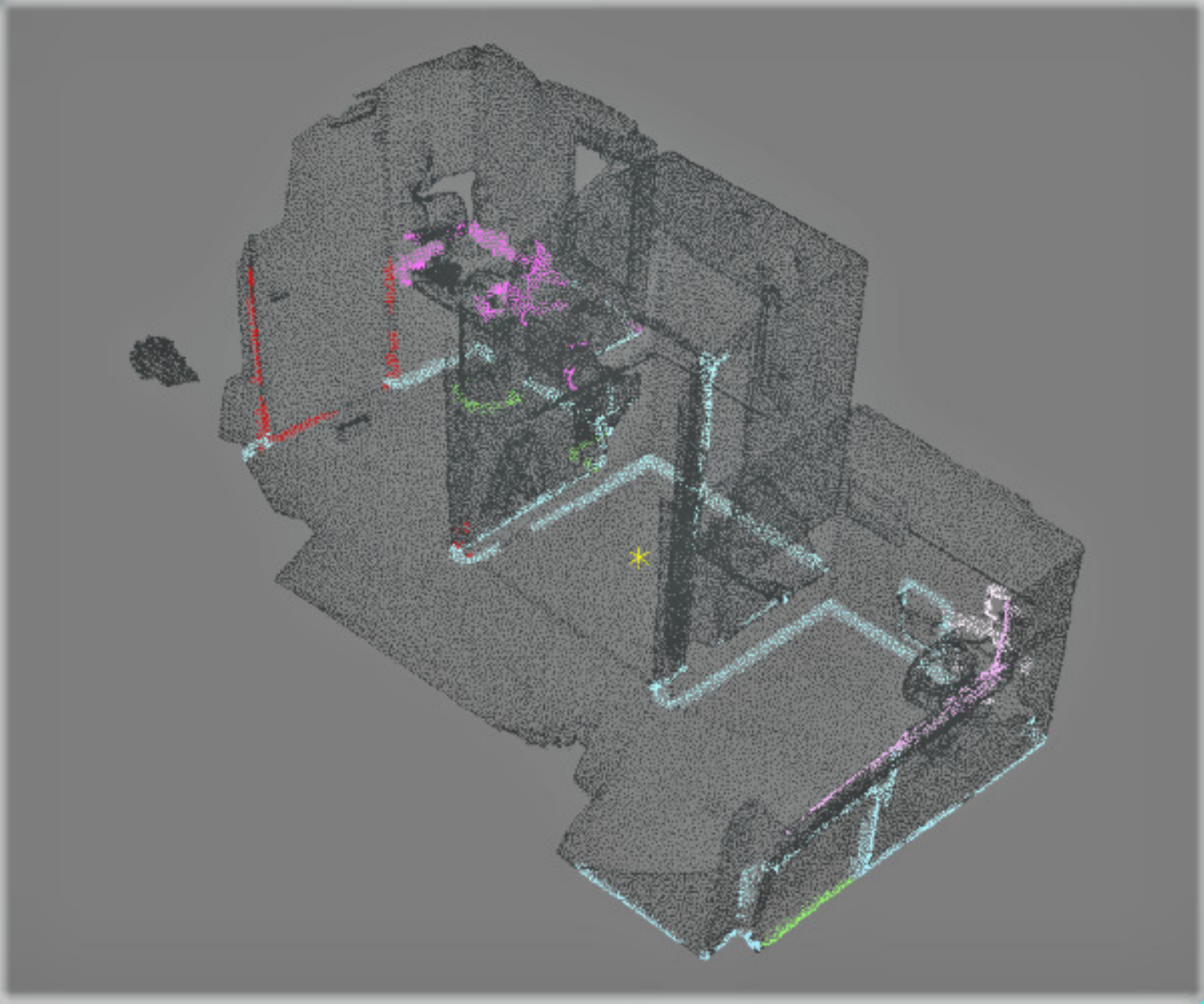}
  \caption{Pred-SEP Map}
\end{subfigure}

\caption{Qualitative results on ScanNet val set.}
\label{fig:qualitative_scannet}
\end{figure}

\begin{figure}[htp]

\centering
\includegraphics[width=\textwidth, height=0.68\textwidth]{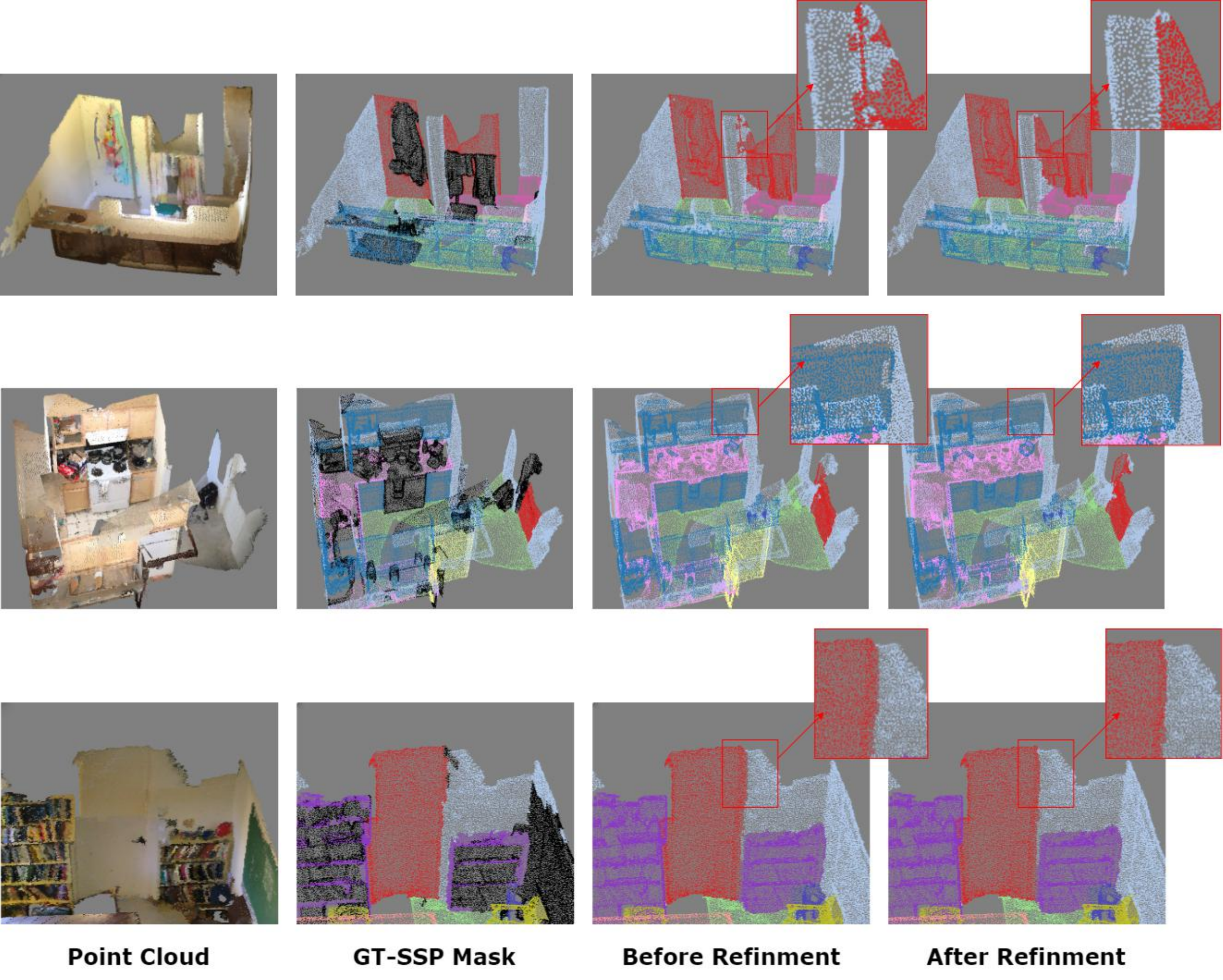}

\caption{ Some visualization comparison examples for semantic segmentation before and after joint refinement (best viewed in color).
}
\label{fig:seg_compare}

\end{figure}

\begin{figure}[htp]

\centering
\includegraphics[width=\textwidth]{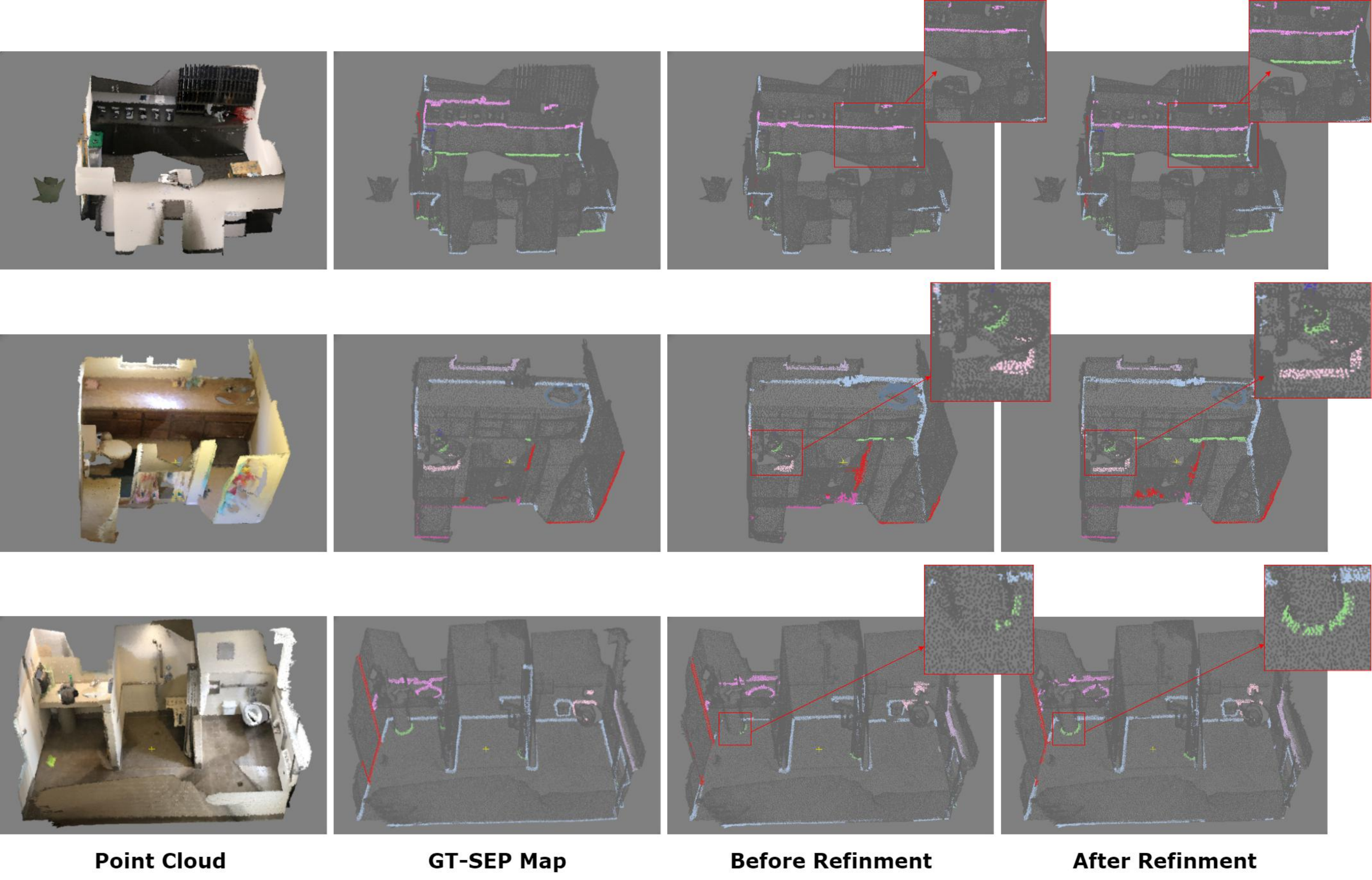}

\caption{ Some visualization comparison examples for semantic edge detection before and after joint refinement (best viewed in color). For better visualization, we thickened all the semantic edges.
}
\label{fig:edge_compare}

\end{figure}

In this section, we present more visualization results. More visualization results of our method on the ScanNet dataset are shown in Fig. \ref{fig:qualitative_scannet}. Qualitative comparison on the effects of joint refinement are shown in Fig. \ref{fig:seg_compare} and Fig. \ref{fig:edge_compare}. Black points in the GT SSP masks are unlabeled points or points of unconsidered classes. All semantic edges are thickened for visualization.

\clearpage

\section{Detailed semantic segmentation results.}\label{class_results}
In this section, we provide more details on our semantic segmentation experiments, for benchmarking purpose with future works. Detailed class scores for the S3DIS dataset and the ScanNet dataset are presented in Table \ref{table:ss_s3dis} and Table \ref{table:ss_scannet}, respectively.

\begin{table}
\begin{center}
\caption{
Detailed mIoU scores (\%) {of semantic segmentation} on S3DIS Area-5.}
\label{table:ss_s3dis}
\resizebox{\textwidth}{!}{
    \begin{tabular}{  | l | c | c c c c c c c c c c c c c|}
    \hline
    Method & mIoU & ceil. & floor & wall & beam & col. & wind. & door & chair & table & book. & sofa & board & clut.\\
    \hline
    Pointnet \cite{qi2017pointnet} & 41.1 & 88.8 & 97.3 & 69.8 & 0.1 & 3.9 & 46.3 & 10.8 & 52.6 & 58.9 & 40.3 & 5.9 & 26.4 & 33.2\\ 
    SegCloud \cite{tchapmi2017segcloud} & 48.9 & 90.1 & 96.1 & 69.9 & 0.0 & 18.4 & 38.4 & 23.1 & 75.9 & 70.4 & 58.4 & 40.9 & 13.0 & 41.6\\
    Eff 3D Conv \cite{zhang2018efficient} & 51.8 & 79.8 & 93.9 & 69.0 & 0.2 & 28.3 & 38.5 & 48.3 & 71.1 & 73.6 & 48.7 & 59.2 & 29.3 & 33.1\\
    TangentConv \cite{tatarchenko2018tangent} & 52.6 & 90.5 & 97.7 & 74.0 & 0.0 & 20.7 & 39.0 & 31.3 & 69.4 & 77.5 & 38.5 & 57.3 & 48.8 & 39.8\\
    RNN Fusion \cite{ye20183d} & 53.4 & \textbf{95.2} & 98.6 & 77.4 & \textbf{0.8} & 9.8 & 52.7 & 27.9 & 78.3 & 76.8 & 27.4 & 58.6 & 39.1 & 51.0\\
    PointCNN \cite{li2018pointcnn} & 57.3 & 92.3 & 98.2 & 79.4 & 0.0 & 17.6 & 22.8 & 62.1 & 74.4 & 80.6 & 31.7 & 66.7 & 62.1 & 56.7\\
    SPGraph \cite{landrieu2018large} & 58.0 & 89.4 & 96.9 & 78.1 & 0.0 & \textbf{42.8} & 48.9 & 61.6 & 84.7 & 75.4 & 69.8 & 52.6 & 2.1 & 52.2\\
    ParamConv \cite{wang2018deep} & 58.3 & 92.3 & 96.2 & 75.9 & 0.3 & 6.0 & \textbf{69.5} & 63.5 & 66.9 & 65.6 & 47.3 & 68.9 & 59.1 & 46.2\\
    SPH3D-GCN \cite{lei2019spherical} & 59.5 & 93.3 & 97.1 & 81.1 & 0.0 & 33.2 & 45.8 & 43.8 & 79.7 & 86.9 & 33.2 & 71.5 & 54.1 & 53.7\\
    HPEIN \cite{jiang2019hierarchical} & 61.9 & 91.5 & 98.2 & 81.4 & 0.0 & 23.3 & 65.3 & 40.0 & 75.5 & \textbf{87.7} & 58.5 & 67.8 & 65.6 & 49.4\\
    MinkowskiNet \cite{Choy_2019} & 65.4 & 91.8 & \textbf{98.7} & \textbf{86.2} & 0.0 & 34.1 & 48.9 & 62.4 & 89.8 & 81.6 & 74.9 & 47.2 & \textbf{74.4} & 58.6\\
    KPConv rigid \cite{thomas2019kpconv} & 65.4 & 92.6 & 97.3 & 81.4 & 0.0 & 16.5 & 54.5 & 69.5 & \textbf{90.1} & 80.2 & 74.6 & 66.4 & 63.7 & 58.1\\
    \hline
    JSENet (ours) & \textbf{67.7} & 93.8 & 97.0 & 83.0 & 0.0 & 23.2 & 61.3 & \textbf{71.6} & 89.9 & 79.8 & \textbf{75.6} & \textbf{72.3} & 72.7 & \textbf{60.4}\\
    \hline
    \end{tabular}}
\end{center}
\end{table}

\begin{table}
\begin{center}
\caption{
Detailed mIoU scores (\%) {of semantic segmentation} on ScanNet test set.}
\label{table:ss_scannet}
\resizebox{\textwidth}{!}{
    \begin{tabular}{  | l | c | c c c c c c c c c c c c c c c c c c c c|}
    \hline
    Method & mIoU & bath & bed & bksf & cab & chair & cntr & curt & desk & door & floor & othr & pic & ref & show & sink & sofa & tab & toil & wall & wind\\
    \hline
    ScanNet \cite{dai2017scannet} & 30.6 & 20.3 & 36.6 & 50.1 & 31.1 & 52.4 & 21.1 & 0.2 & 34.2 & 18.9 & 78.6 & 14.5 & 10.2 & 24.5 & 15.2 & 31.8 & 34.8 & 30.0 & 46.0 & 43.7 & 18.2\\
    PointNet++ \cite{qi2017pointnet++} & 33.9 & 58.4 & 47.8 & 45.8 & 25.6 & 36.0 & 25.0 & 24.7 & 27.8 & 26.1 & 67.7 & 18.3 & 11.7 & 21.2 & 14.5 & 36.4 & 34.6 & 23.2 & 54.8 & 52.3 & 25.2\\
    SPLATNet \cite{Su_2018} & 39.3 & 47.2 & 51.1 & 60.6 & 31.1 & 65.6 & 24.5 & 40.5 & 32.8 & 19.7 & 92.7 & 22.7 & 0
    0 & 0.1 & 24.9 & 27.1 & 51.0 & 38.3 & 59.3 & 69.9 & 26.7\\
    TangentConv \cite{tatarchenko2018tangent} & 43.8 & 43.7 & 64.6 & 47.4 & 36.9 & 64.5 & 35.3 & 25.8 & 28.2 & 27.9 & 91.8 & 29.8 & 14.7 & 28.3 & 29.4 & 48.7 & 56.2 & 42.7 & 61.9 & 63.3 & 35.2\\
    PointCNN \cite{li2018pointcnn} & 45.8 & 57.7 & 61.1 & 35.6 & 32.1 & 71.5 & 29.9 & 37.6 & 32.8 & 31.9 & 94.4 & 28.5 & 16.4 & 21.6 & 22.9 & 48.4 & 54.5 & 45.6 & 75.5 & 70.9 & 47.5\\
    PanopticFusion \cite{narita2019panopticfusion} & 52.9 & 49.1 & 68.8 & 60.4 & 38.6 & 63.2 & 22.5 & 70.5 & 43.4 & 29.3 & 81.5 & 34.8 & 24.1 & 49.9 & 66.9 & 50.7 & 64.9 & 44.2 & 79.6 & 60.2 & 56.1\\
    TextureNet \cite{huang2019texturenet} & 56.6 & 67.2 & 66.4 & 67.1 & 49.4 & 71.9 & 44.5 & 67.8 & 41.1 & 39.6 & 93.5 & 35.6 & 22.5 & 41.2 & 53.5 & 56.5 & 63.6 & 46.4 & 79.4 & 68.0 & 56.8\\
    SPH3D-GCN \cite{lei2019spherical} & 61.0 & 85.8 & 77.2 & 48.9 & 53.2 & 79.2 & 40.4 & 64.3 & 57.0 & 50.7 & 93.5 & 41.4 & 4.6 & 51.0 & 70.2 & 60.2 & 70.5 & 54.9 & 85.9 & 77.3 & 53.4\\
    HPEIN \cite{jiang2019hierarchical} & 61.8 & 72.9 & 66.8 & 64.7 & 59.7 & 76.6 & 41.4 & 68.0 & 52.0 & 52.5 & 94.6 & 43.2 & 21.5 & 49.3 & 59.9 & 63.8 & 61.7 & 57.0 & 89.7 & 80.6 & 60.5\\
    KP-FCNN \cite{thomas2019kpconv} & 68.4 & 84.7 & 75.8 & 78.4 & 64.7 & 81.4 & 47.3 & 77.2 & 60.5 & 59.4 & 93.5 & 45.0 & 18.1 & 58.7 & 80.5 & 69.0 & 78.5 & 61.4 & 88.2 & 81.9 & 63.2\\
    SparseCOnvNet \cite{graham20183d} & 72.5 & 64.7 & 82.1 & 84.6 & 72.1 & 86.9 & 53.3 & 75.4 & 60.3 & 61.4 & 95.5 & 57.2 & 32.5 & 71.0 & 87.0 & 72.4 & 82.3 & 62.8 & 93.4 & 86.5 & 68.3\\
    MinkowskiNet \cite{Choy_2019} & 73.6 & 85.9 & 81.8 & 83.2 & 70.9 & 84.0 & 52.1 & 85.3 & 66.0 & 64.3 & 95.1 & 54.4 & 28.6 & 73.1 & 89.3 & 67.5 & 77.2 & 68.3 & 87.4 & 85.2 & 72.7\\
    \hline
    JSENet (ours) & 69.9 & 88.1 & 76.2 & 82.1 & 66.7 & 80.0 & 52.2 & 79.2 & 61.3 & 60.7 & 93.5 & 49.2 & 20.5 & 57.6 & 85.3 & 69.1 & 75.8 & 65.2 & 87.2 & 82.8 & 64.9\\
    \hline
    \end{tabular}}
\end{center}
\end{table}

\clearpage

%% file: eccv2020submission.bbl
\begin{thebibliography}{10}

\bibitem{long2015fully}
Long, J., Shelhamer, E., Darrell, T.:
\newblock Fully convolutional networks for semantic segmentation.
\newblock In: Proceedings of the IEEE conference on computer vision and pattern
  recognition. (2015)  3431--3440

\bibitem{takikawa2019gated}
Takikawa, T., Acuna, D., Jampani, V., Fidler, S.:
\newblock Gated-scnn: Gated shape cnns for semantic segmentation.
\newblock In: Proceedings of the IEEE International Conference on Computer
  Vision. (2019)  5229--5238

\bibitem{li2018pointcnn}
Li, Y., Bu, R., Sun, M., Wu, W., Di, X., Chen, B.:
\newblock Pointcnn: Convolution on x-transformed points.
\newblock In: Advances in neural information processing systems. (2018)
  820--830

\bibitem{jaritz2019multi}
Jaritz, M., Gu, J., Su, H.:
\newblock Multi-view pointnet for 3d scene understanding.
\newblock In: Proceedings of the IEEE International Conference on Computer
  Vision Workshops. (2019)  0--0

\bibitem{qi2017pointnet}
Qi, C.R., Su, H., Mo, K., Guibas, L.J.:
\newblock Pointnet: Deep learning on point sets for 3d classification and
  segmentation.
\newblock In: Proceedings of the IEEE conference on computer vision and pattern
  recognition. (2017)  652--660

\bibitem{qi2017pointnet++}
Qi, C.R., Yi, L., Su, H., Guibas, L.J.:
\newblock Pointnet++: Deep hierarchical feature learning on point sets in a
  metric space.
\newblock In: Advances in neural information processing systems. (2017)
  5099--5108

\bibitem{dai2017scannet}
Dai, A., Chang, A.X., Savva, M., Halber, M., Funkhouser, T., Nie{\ss}ner, M.:
\newblock Scannet: Richly-annotated 3d reconstructions of indoor scenes.
\newblock In: Proc. Computer Vision and Pattern Recognition (CVPR), IEEE.
  (2017)

\bibitem{yu2017casenet}
Yu, Z., Feng, C., Liu, M.Y., Ramalingam, S.:
\newblock Casenet: Deep category-aware semantic edge detection.
\newblock In: Proceedings of the IEEE Conference on Computer Vision and Pattern
  Recognition. (2017)  5964--5973

\bibitem{liu2018semantic}
Liu, Y., Cheng, M.M., Fan, D.P., Zhang, L., Bian, J., Tao, D.:
\newblock Semantic edge detection with diverse deep supervision (2018)

\bibitem{yu2018simultaneous}
Yu, Z., Liu, W., Zou, Y., Feng, C., Ramalingam, S., Vijaya~Kumar, B., Kautz,
  J.:
\newblock Simultaneous edge alignment and learning.
\newblock In: Proceedings of the European Conference on Computer Vision (ECCV).
  (2018)  388--404

\bibitem{acuna2019devil}
Acuna, D., Kar, A., Fidler, S.:
\newblock Devil is in the edges: Learning semantic boundaries from noisy
  annotations.
\newblock In: Proceedings of the IEEE Conference on Computer Vision and Pattern
  Recognition. (2019)  11075--11083

\bibitem{armeni_cvpr16}
Armeni, I., Sener, O., Zamir, A.R., Jiang, H., Brilakis, I., Fischer, M.,
  Savarese, S.:
\newblock 3d semantic parsing of large-scale indoor spaces.
\newblock In: Proceedings of the IEEE International Conference on Computer
  Vision and Pattern Recognition. (2016)

\bibitem{zhao2017pyramid}
Zhao, H., Shi, J., Qi, X., Wang, X., Jia, J.:
\newblock Pyramid scene parsing network.
\newblock In: Proceedings of the IEEE conference on computer vision and pattern
  recognition. (2017)  2881--2890

\bibitem{lin2017refinenet}
Lin, G., Milan, A., Shen, C., Reid, I.:
\newblock Refinenet: Multi-path refinement networks for high-resolution
  semantic segmentation.
\newblock In: Proceedings of the IEEE conference on computer vision and pattern
  recognition. (2017)  1925--1934

\bibitem{wu2019stacked}
Wu, Z., Su, L., Huang, Q.:
\newblock Stacked cross refinement network for edge-aware salient object
  detection.
\newblock In: Proceedings of the IEEE International Conference on Computer
  Vision. (2019)  7264--7273

\bibitem{yu2018learning}
Yu, C., Wang, J., Peng, C., Gao, C., Yu, G., Sang, N.:
\newblock Learning a discriminative feature network for semantic segmentation.
\newblock In: Proceedings of the IEEE conference on computer vision and pattern
  recognition. (2018)  1857--1866

\bibitem{cheng2017fusionnet}
Cheng, D., Meng, G., Xiang, S., Pan, C.:
\newblock Fusionnet: Edge aware deep convolutional networks for semantic
  segmentation of remote sensing harbor images.
\newblock IEEE Journal of Selected Topics in Applied Earth Observations and
  Remote Sensing \textbf{10}(12) (2017)  5769--5783

\bibitem{bertasius2016semantic}
Bertasius, G., Shi, J., Torresani, L.:
\newblock Semantic segmentation with boundary neural fields.
\newblock In: Proceedings of the IEEE conference on computer vision and pattern
  recognition. (2016)  3602--3610

\bibitem{peng2017large}
Peng, C., Zhang, X., Yu, G., Luo, G., Sun, J.:
\newblock Large kernel matters--improve semantic segmentation by global
  convolutional network.
\newblock In: Proceedings of the IEEE conference on computer vision and pattern
  recognition. (2017)  4353--4361

\bibitem{su2019selectivity}
Su, J., Li, J., Zhang, Y., Xia, C., Tian, Y.:
\newblock Selectivity or invariance: Boundary-aware salient object detection.
\newblock In: Proceedings of the IEEE International Conference on Computer
  Vision. (2019)  3799--3808

\bibitem{thomas2019kpconv}
Thomas, H., Qi, C.R., Deschaud, J.E., Marcotegui, B., Goulette, F., Guibas,
  L.J.:
\newblock Kpconv: Flexible and deformable convolution for point clouds.
\newblock In: Proceedings of the IEEE International Conference on Computer
  Vision. (2019)  6411--6420

\bibitem{boulch2017unstructured}
Boulch, A., Le~Saux, B., Audebert, N.:
\newblock Unstructured point cloud semantic labeling using deep segmentation
  networks.
\newblock 3DOR \textbf{2} (2017) ~7

\bibitem{lawin2017deep}
Lawin, F.J., Danelljan, M., Tosteberg, P., Bhat, G., Khan, F.S., Felsberg, M.:
\newblock Deep projective 3d semantic segmentation.
\newblock In: International Conference on Computer Analysis of Images and
  Patterns, Springer (2017)  95--107

\bibitem{roynard2018classification}
Roynard, X., Deschaud, J.E., Goulette, F.:
\newblock Classification of point cloud scenes with multiscale voxel deep
  network.
\newblock arXiv preprint arXiv:1804.03583 (2018)

\bibitem{ben20183dmfv}
Ben-Shabat, Y., Lindenbaum, M., Fischer, A.:
\newblock 3dmfv: Three-dimensional point cloud classification in real-time
  using convolutional neural networks.
\newblock IEEE Robotics and Automation Letters \textbf{3}(4) (2018)  3145--3152

\bibitem{le2018pointgrid}
Le, T., Duan, Y.:
\newblock Pointgrid: A deep network for 3d shape understanding.
\newblock In: Proceedings of the IEEE conference on computer vision and pattern
  recognition. (2018)  9204--9214

\bibitem{meng2018vvnet}
Meng, H.Y., Gao, L., Lai, Y., Manocha, D.:
\newblock Vv-net: Voxel vae net with group convolutions for point cloud
  segmentation (2018)

\bibitem{riegler2017octnet}
Riegler, G., Osman~Ulusoy, A., Geiger, A.:
\newblock Octnet: Learning deep 3d representations at high resolutions.
\newblock In: Proceedings of the IEEE Conference on Computer Vision and Pattern
  Recognition. (2017)  3577--3586

\bibitem{graham20183d}
Graham, B., Engelcke, M., van~der Maaten, L.:
\newblock 3d semantic segmentation with submanifold sparse convolutional
  networks.
\newblock In: Proceedings of the IEEE conference on computer vision and pattern
  recognition. (2018)  9224--9232

\bibitem{Choy_2019}
Choy, C., Gwak, J., Savarese, S.:
\newblock 4d spatio-temporal convnets: Minkowski convolutional neural networks.
\newblock 2019 IEEE/CVF Conference on Computer Vision and Pattern Recognition
  (CVPR) (Jun 2019)

\bibitem{li2018so}
Li, J., Chen, B.M., Hee~Lee, G.:
\newblock So-net: Self-organizing network for point cloud analysis.
\newblock In: Proceedings of the IEEE conference on computer vision and pattern
  recognition. (2018)  9397--9406

\bibitem{Huang_2018}
Huang, Q., Wang, W., Neumann, U.:
\newblock Recurrent slice networks for 3d segmentation of point clouds.
\newblock 2018 IEEE/CVF Conference on Computer Vision and Pattern Recognition
  (Jun 2018)

\bibitem{zhao2019pointweb}
Zhao, H., Jiang, L., Fu, C.W., Jia, J.:
\newblock Pointweb: Enhancing local neighborhood features for point cloud
  processing.
\newblock In: Proceedings of the IEEE Conference on Computer Vision and Pattern
  Recognition. (2019)  5565--5573

\bibitem{zhang2019shellnet}
Zhang, Z., Hua, B.S., Yeung, S.K.:
\newblock Shellnet: Efficient point cloud convolutional neural networks using
  concentric shells statistics.
\newblock In: Proceedings of the IEEE International Conference on Computer
  Vision. (2019)  1607--1616

\bibitem{wang2019dynamic}
Wang, Y., Sun, Y., Liu, Z., Sarma, S.E., Bronstein, M.M., Solomon, J.M.:
\newblock Dynamic graph cnn for learning on point clouds.
\newblock ACM Transactions on Graphics (TOG) \textbf{38}(5) (2019)  1--12

\bibitem{wang2019graph}
Wang, L., Huang, Y., Hou, Y., Zhang, S., Shan, J.:
\newblock Graph attention convolution for point cloud semantic segmentation.
\newblock In: Proceedings of the IEEE Conference on Computer Vision and Pattern
  Recognition. (2019)  10296--10305

\bibitem{jiang2019hierarchical}
Jiang, L., Zhao, H., Liu, S., Shen, X., Fu, C.W., Jia, J.:
\newblock Hierarchical point-edge interaction network for point cloud semantic
  segmentation.
\newblock In: Proceedings of the IEEE International Conference on Computer
  Vision. (2019)  10433--10441

\bibitem{liu2019dynamic}
Liu, J., Ni, B., Li, C., Yang, J., Tian, Q.:
\newblock Dynamic points agglomeration for hierarchical point sets learning.
\newblock In: Proceedings of the IEEE International Conference on Computer
  Vision. (2019)  7546--7555

\bibitem{xie2018attentional}
Xie, S., Liu, S., Chen, Z., Tu, Z.:
\newblock Attentional shapecontextnet for point cloud recognition.
\newblock In: Proceedings of the IEEE Conference on Computer Vision and Pattern
  Recognition. (2018)  4606--4615

\bibitem{Su_2018}
Su, H., Jampani, V., Sun, D., Maji, S., Kalogerakis, E., Yang, M.H., Kautz, J.:
\newblock Splatnet: Sparse lattice networks for point cloud processing.
\newblock 2018 IEEE/CVF Conference on Computer Vision and Pattern Recognition
  (Jun 2018)

\bibitem{hua2018pointwise}
Hua, B.S., Tran, M.K., Yeung, S.K.:
\newblock Pointwise convolutional neural networks.
\newblock In: Proceedings of the IEEE Conference on Computer Vision and Pattern
  Recognition. (2018)  984--993

\bibitem{wu2019pointconv}
Wu, W., Qi, Z., Fuxin, L.:
\newblock Pointconv: Deep convolutional networks on 3d point clouds.
\newblock In: Proceedings of the IEEE Conference on Computer Vision and Pattern
  Recognition. (2019)  9621--9630

\bibitem{Lei_2019}
Lei, H., Akhtar, N., Mian, A.:
\newblock Octree guided cnn with spherical kernels for 3d point clouds.
\newblock 2019 IEEE/CVF Conference on Computer Vision and Pattern Recognition
  (CVPR) (Jun 2019)

\bibitem{Komarichev_2019}
Komarichev, A., Zhong, Z., Hua, J.:
\newblock A-cnn: Annularly convolutional neural networks on point clouds.
\newblock 2019 IEEE/CVF Conference on Computer Vision and Pattern Recognition
  (CVPR) (Jun 2019)

\bibitem{Lan_2019}
Lan, S., Yu, R., Yu, G., Davis, L.S.:
\newblock Modeling local geometric structure of 3d point clouds using geo-cnn.
\newblock 2019 IEEE/CVF Conference on Computer Vision and Pattern Recognition
  (CVPR) (Jun 2019)

\bibitem{mao2019interpolated}
Mao, J., Wang, X., Li, H.:
\newblock Interpolated convolutional networks for 3d point cloud understanding.
\newblock In: Proceedings of the IEEE International Conference on Computer
  Vision. (2019)  1578--1587

\bibitem{prasad2006learning}
Prasad, M., Zisserman, A., Fitzgibbon, A., Kumar, M.P., Torr, P.H.:
\newblock Learning class-specific edges for object detection and segmentation.
\newblock In: Computer Vision, Graphics and Image Processing.
\newblock Springer (2006)  94--105

\bibitem{hariharan2011semantic}
Hariharan, B., Arbel{\'a}ez, P., Bourdev, L., Maji, S., Malik, J.:
\newblock Semantic contours from inverse detectors.
\newblock In: 2011 International Conference on Computer Vision, IEEE (2011)
  991--998

\bibitem{Bertasius_2015}
Bertasius, G., Shi, J., Torresani, L.:
\newblock High-for-low and low-for-high: Efficient boundary detection from deep
  object features and its applications to high-level vision.
\newblock 2015 IEEE International Conference on Computer Vision (ICCV) (Dec
  2015)

\bibitem{xie2015holistically}
Xie, S., Tu, Z.:
\newblock Holistically-nested edge detection.
\newblock In: Proceedings of the IEEE international conference on computer
  vision. (2015)  1395--1403

\bibitem{hu2019panoptic}
Hu, Y., Zou, Y., Feng, J.:
\newblock Panoptic edge detection (2019)

\bibitem{Ronneberger_2015}
Ronneberger, O., Fischer, P., Brox, T.:
\newblock U-net: Convolutional networks for biomedical image segmentation.
\newblock Medical Image Computing and Computer-Assisted Intervention – MICCAI
  2015 (2015)  234–241

\bibitem{tchapmi2017segcloud}
Tchapmi, L., Choy, C., Armeni, I., Gwak, J., Savarese, S.:
\newblock Segcloud: Semantic segmentation of 3d point clouds.
\newblock In: 2017 international conference on 3D vision (3DV), IEEE (2017)
  537--547

\bibitem{tatarchenko2018tangent}
Tatarchenko, M., Park, J., Koltun, V., Zhou, Q.Y.:
\newblock Tangent convolutions for dense prediction in 3d.
\newblock In: Proceedings of the IEEE Conference on Computer Vision and Pattern
  Recognition. (2018)  3887--3896

\bibitem{ye20183d}
Ye, X., Li, J., Huang, H., Du, L., Zhang, X.:
\newblock 3d recurrent neural networks with context fusion for point cloud
  semantic segmentation.
\newblock In: Proceedings of the European Conference on Computer Vision (ECCV).
  (2018)  403--417

\bibitem{landrieu2018large}
Landrieu, L., Simonovsky, M.:
\newblock Large-scale point cloud semantic segmentation with superpoint graphs.
\newblock In: Proceedings of the IEEE Conference on Computer Vision and Pattern
  Recognition. (2018)  4558--4567

\bibitem{rethage2018fully}
Rethage, D., Wald, J., Sturm, J., Navab, N., Tombari, F.:
\newblock Fully-convolutional point networks for large-scale point clouds.
\newblock In: Proceedings of the European Conference on Computer Vision (ECCV).
  (2018)  596--611

\bibitem{wang2018deep}
Wang, S., Suo, S., Ma, W.C., Pokrovsky, A., Urtasun, R.:
\newblock Deep parametric continuous convolutional neural networks.
\newblock In: Proceedings of the IEEE Conference on Computer Vision and Pattern
  Recognition. (2018)  2589--2597

\bibitem{narita2019panopticfusion}
Narita, G., Seno, T., Ishikawa, T., Kaji, Y.:
\newblock Panopticfusion: Online volumetric semantic mapping at the level of
  stuff and things.
\newblock arXiv preprint arXiv:1903.01177 (2019)

\bibitem{huang2019texturenet}
Huang, J., Zhang, H., Yi, L., Funkhouser, T., Nie{\ss}ner, M., Guibas, L.J.:
\newblock Texturenet: Consistent local parametrizations for learning from
  high-resolution signals on meshes.
\newblock In: Proceedings of the IEEE Conference on Computer Vision and Pattern
  Recognition. (2019)  4440--4449

\bibitem{lei2019spherical}
Lei, H., Akhtar, N., Mian, A.:
\newblock Spherical kernel for efficient graph convolution on 3d point clouds.
\newblock arXiv preprint arXiv:1909.09287 (2019)

\bibitem{hermosilla2018monte}
Hermosilla, P., Ritschel, T., V{\'a}zquez, P.P., Vinacua, {\`A}., Ropinski, T.:
\newblock Monte carlo convolution for learning on non-uniformly sampled point
  clouds.
\newblock ACM Transactions on Graphics (TOG) \textbf{37}(6) (2018)  1--12

\bibitem{su2018splatnet}
Su, H., Jampani, V., Sun, D., Maji, S., Kalogerakis, E., Yang, M.H., Kautz, J.:
\newblock Splatnet: Sparse lattice networks for point cloud processing.
\newblock In: Proceedings of the IEEE Conference on Computer Vision and Pattern
  Recognition. (2018)  2530--2539

\bibitem{zhang2018efficient}
Zhang, C., Luo, W., Urtasun, R.:
\newblock Efficient convolutions for real-time semantic segmentation of 3d
  point clouds.
\newblock In: 2018 International Conference on 3D Vision (3DV), IEEE (2018)
  399--408

\bibitem{hackel2017isprs}
Hackel, T., Savinov, N., Ladicky, L., Wegner, J.D., Schindler, K., Pollefeys,
  M.:
\newblock {SEMANTIC3D.NET: A new large-scale point cloud classification
  benchmark}.
\newblock In: ISPRS Annals of the Photogrammetry, Remote Sensing and Spatial
  Information Sciences. Volume IV-1-W1. (2017)  91--98

\end{thebibliography}
